\documentclass[10pt,twocolumn,letterpaper]{article}

\usepackage[pagenumbers]{cvpr} %

\usepackage[dvipsnames]{xcolor}

\usepackage{times}  %
\usepackage{helvet}  %
\usepackage{courier}  %
\usepackage[hyphens]{url}  %
\usepackage{graphicx} %
\usepackage{mathtools}
\usepackage{bm}
\usepackage{amsmath}
\usepackage{amssymb}
\usepackage{natbib}  %
\usepackage{caption} %
\usepackage{multirow}
\usepackage{dsfont}

\usepackage{algorithm}
\usepackage{colortbl}
\usepackage{pifont}

\newcommand{\xmark}{\ding{55}}
\newcommand{\cmark}{\ding{51}}
\newcommand{\bestresult}[1]{\textbf{\textcolor{red}{#1}}}
\newcommand{\secondbest}[1]{\textcolor{blue}{\underline{#1}}}

\newcommand{\venue}[1]{{$_{\text{#1}}$}}

\newcommand{\supp}[1]{{\color{blue}  #1}}
\newcommand{\decrease}[1]{\scriptsize \textcolor{red}{$\downarrow$#1} \normalsize}
\definecolor{aoenglish}{rgb}{0.0, 0.5, 0.0}
\newcommand{\increase}[1]{\scriptsize \textcolor{aoenglish}{$\uparrow$#1} \normalsize}

\usepackage{newfloat}
\usepackage{listings}
\usepackage{algorithm}
\usepackage{algpseudocode}
\usepackage{beramono} %
\definecolor{keywordcolor}{rgb}{0.13,0.29,0.53}
\definecolor{commentcolor}{rgb}{0.56,0.35,0.01}
\definecolor{variablecolor}{rgb}{0.25,0.5,0.35}
\definecolor{functioncolor}{rgb}{0.38,0.19,0.84}
\definecolor{sectionlinecolor}{rgb}{0,0,0} %

\newcommand{\var}[1]{{\color{variablecolor} \texttt{#1}}}
\newcommand{\func}[1]{{\color{functioncolor} \texttt{#1}}}
\newcommand{\comm}[1]{{\color{commentcolor} \texttt{\#} \textit{#1}}}
\algnewcommand{\Comm}[1]{\Statex \hspace{1.5em} \comm{#1}} %
\newcommand{\sectionline}{\Statex\color{sectionlinecolor}\rule{\linewidth}{0.8mm}}

\definecolor{cvprblue}{rgb}{0.21,0.49,0.74}
\usepackage[pagebackref,breaklinks,colorlinks,citecolor=cvprblue]{hyperref}

\def\paperID{*****} %
\def\confName{CVPR}
\def\confYear{2024}

\title{No More Shortcuts: Realizing the Potential of Temporal Self-Supervision}

\author{Ishan Rajendrakumar Dave\textsuperscript{\rm 1}\thanks{Majority of work done as an intern at Adobe Research, USA.},~ Simon Jenni\textsuperscript{\rm 2},~ Mubarak Shah\textsuperscript{\rm 1}\\
\textsuperscript{\rm 1}Center for Research in Computer Vision, University of Central Florida, USA\\
    \textsuperscript{\rm 2}Adobe Research, USA\\
{\tt\small ishandave@ucf.edu, jenni@adobe.com, shah@crcv.ucf.edu}\\
    Project Webpage: \url{https://daveishan.github.io/nms-webpage}
}

\begin{document}
\maketitle
\begin{abstract}
Self-supervised approaches for video have shown impressive results in video understanding tasks.
However, unlike early works that leverage temporal self-supervision, current state-of-the-art methods primarily rely on tasks from the image domain (e.g., contrastive learning) that do not explicitly promote the learning of temporal features.
We identify two factors that limit existing temporal self-supervision: 1) tasks are too simple, resulting in saturated training performance, and 2) we uncover shortcuts based on local appearance statistics that hinder the learning of high-level features.
To address these issues, we propose 1) a more challenging reformulation of temporal self-supervision as frame-level (rather than clip-level) recognition tasks and 2) an effective augmentation strategy to mitigate shortcuts. 
Our model extends a representation of single video frames, pre-trained through contrastive learning, with a transformer that we train through temporal self-supervision.  
We demonstrate experimentally that our more challenging frame-level task formulations and the removal of shortcuts drastically improve the quality of features learned through temporal self-supervision.  
The generalization capability of our self-supervised video method is evidenced by its state-of-the-art performance in a wide range of high-level semantic tasks, including video retrieval, action classification, and video attribute recognition (such as object and scene identification), as well as low-level temporal correspondence tasks like video object segmentation and pose tracking. 
Additionally, we show that the video representations learned through our method exhibit increased robustness to the input perturbations.

\end{abstract}

\section{Introduction}

\begin{figure}[t]
    \centering
    \includegraphics[width=\columnwidth]{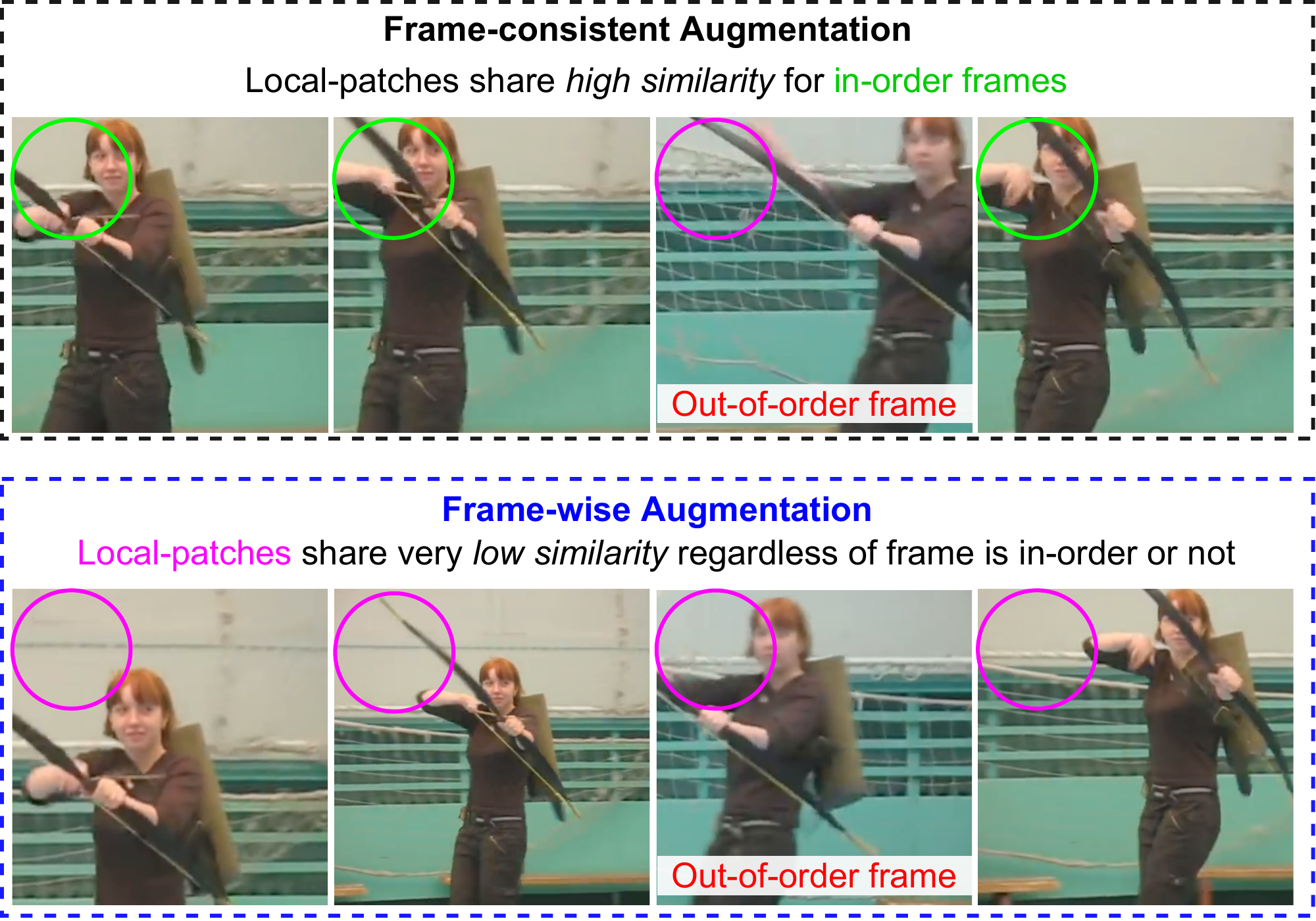}

    \caption{\textbf{Breaking Shortcuts in Temporal SSL Tasks.} 
    We illustrate an example of our pretext task where the model has to localize out-of-order frames.
    In the first row, the input sequence has identical augmentations applied to each frame, whereas the second row shows our proposed independent frame-wise augmentations. 
    We observe that the task can be solved by observing only a local image patch (highlighted circle) when identical augmentations are used.
    As a result, the task does not require learning higher-level features, \eg, object dynamics. 
    In contrast, when applying the proposed frame-wise augmentations (second row), a local observation (highlighted circle) is insufficient, and the model must consider the global scene context. %
    }
    \label{fig:shortcut}
\end{figure}

Self-supervised learning (SSL) has unlocked the potential of large amounts of unlabelled data for large-scale pre-training of image \cite{zhou2021ibot, li2021esvit, mae, mocov3, simclr} and, more recently, video representations \cite{videomae, videomoco, Feichtenhofer_2021_CVPR, tclr, stmae}.
The potential benefits of SSL methods on video are even greater than on images due to their larger dimensionality and much greater cost of comprehensive human labeling.  

Video understanding tasks, such as action recognition, depend on representations that capture both the static scene appearance (\eg, object textures, pose, and layout) and scene dynamics (\eg, the change in pose and relative dynamics of objects). 
Human activity recognition, in particular, crucially depends on accurate representations of the human pose and how it deforms over time as a person performs an action. 

However, to what extent current video SSL approaches capture both static and temporal features in videos is unclear. 
Indeed, we observe that many of the current methods rely on contrastive objectives \cite{cvrl, Feichtenhofer_2021_CVPR, videomoco, vtc}, which encourage spatial and temporal invariance and thus do not promote the learning of temporal features.
While several works demonstrated the benefits of including additional objectives that encourage the learning of temporal features, \eg, through pretext tasks \cite{simon, vcop, taco, misra2016shuffle}, it is unclear whether these methods truly capture changes in object dynamics or are rather hindered by relying on lower-level local motion features.

Such ``shortcuts" are common in self-supervised task formulations \cite{doersch2015unsupervised,jenni2018self} and prevent the learning of higher-level features.
Indeed, we observe as one of our key insights, that existing temporal learning tasks, \ie, those recognizing temporal transformations \cite{misra2016shuffle,speedNet,simon}, can be solved by relying on only local appearance statistics (see Figure~\ref{fig:shortcut} for an illustration and Table~\ref{table:shortcut} for additional empirical evidence). 
These low-level solutions hinder the learning of temporal features that capture the global object dynamics. 
Furthermore, existing temporal pretext tasks are often limited by near-perfect training performance, suggesting that the task difficulty is not sufficient.

To address these issues, we make two technical contributions.
First, to address the newly identified shortcuts in temporal self-supervision, we propose an effective frame-independent augmentation strategy. 
In contrast to best practice in (self-supervised) video representation learning with 3D-CNNs \cite{cvrl}, where temporally consistent augmentations are the norm, we demonstrate that an independent per-frame jittering is hugely beneficial for temporal pretext tasks.
As can be seen in Figure~\ref{fig:shortcut}, such frame-wise augmentations force the model to consider global features (\eg, pose) rather than local low-level shortcuts.
This increases the difficulty of the task and promotes better feature learning.

As our second contribution and to further increase the difficulty of the pretext tasks, we propose to reformulate temporal self-supervision as a frame-level time-varying recognition task instead of the typical formulation as a clip-level classification task \cite{misra2016shuffle,speedNet,simon,pace_pred,jenni2021time}.
Concretely, inspired by prior pretext tasks about the temporal ordering and playback speed of whole video clips \cite{misra2016shuffle,speedNet}, we pose the following time-varying tasks: 
1) Out-of-order Frame Localization (OFL) and 2) Time-varying Skiprate Prediction (TSP). 
In OFL, the model has to identify a subset of the frames that are out-of-order, \ie, do not match the natural temporal order of most of the frames.
For TSP, the network needs to predict the playback rate at each frame, which we artificially vary over time. 
Both these tasks require an accurate classification of \textit{each frame} in the sequence, making them more challenging than prior \textit{clip-level} formulations.
We realize these learning tasks with a video transformer architecture \cite{vtn}, wherein a frame encoder (pre-trained through image contrastive learning) is extended with a temporal transformer (trained through our temporal SSL tasks).
This model thus effectively fuses the benefits of contrastive and temporal self-supervision.

Finally, we perform a very comprehensive evaluation of the learned video representations on a large number of downstream video understanding tasks to assess their generalization ability and robustness.
While most prior video SSL studies focussed primarily on action recognition benchmarks (\eg, UCF101, HMDB51, Kinetics400, Something-SomethingV2, NTU60, Charades), we additionally evaluate our representations and compare to the prior state-of-the-art on a variety of other aspects, including holistic video understanding (HVU), temporal correspondence tasks (DAVIS and JHMDB), and gait recognition (CASIA-B). We also evaluate the robustness of the video retrieval task to input video perturbations, following ~\cite{schiappa2023large}.

\noindent \textbf{Contributions.} 
Our contributions can be summarized as follows:
1) We identify shortcuts in temporal SSL based on local patch similarity. To mitigate these shortcuts, we propose a frame-independent augmentation strategy, 
2) We propose frame-wise and time-varying reformulations of temporal pretext tasks instead of the typical clip-level formulations to increase the difficulty of the learning tasks, 
3) We validate our contributions in extensive ablation experiments and comprehensively evaluate the learned video representations' generalization ability, including action-related tasks, holistic video understanding, temporal correspondence, and robustness to input perturbations. Our results demonstrate state-of-the-art performance across numerous benchmarks.

\label{sec:intro}

\begin{figure*}[h]
    \centering
    \includegraphics[width=.86\linewidth]{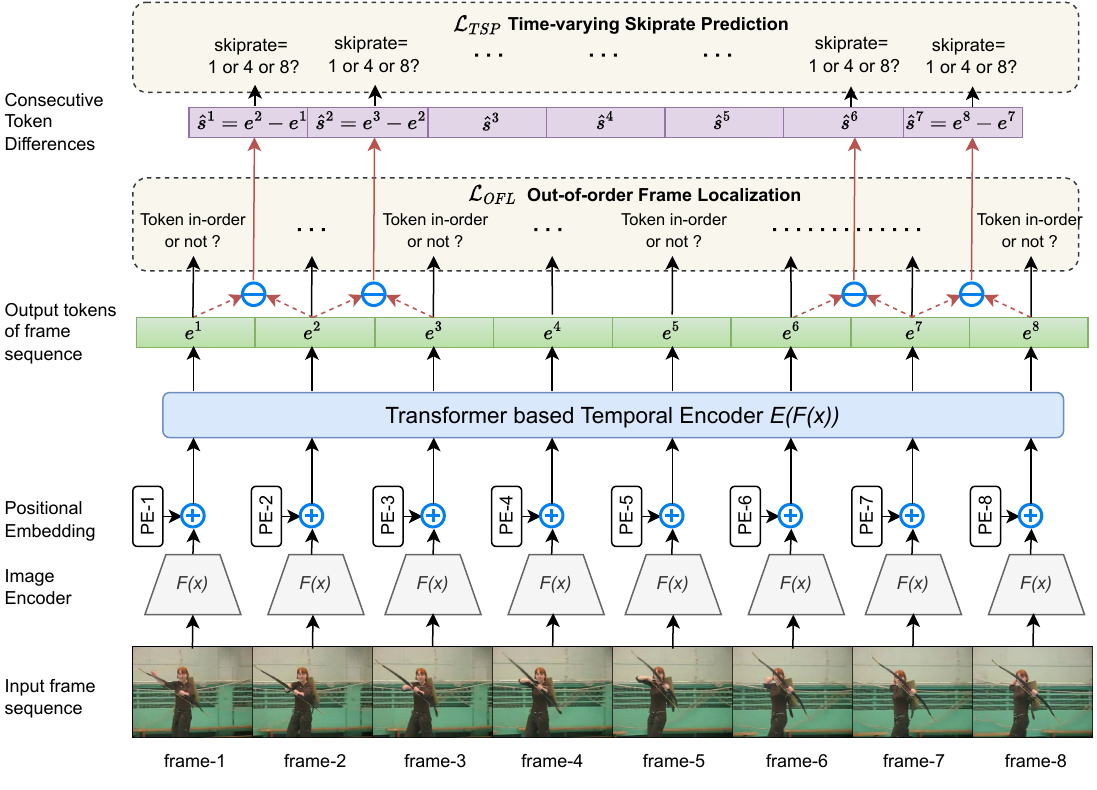}
    \caption{\textbf{Model Overview.}
    We utilize a Video Transformer Network~\cite{vtn} architecture, which extends an image encoder $F$ with a transformer $E$ that takes frame embeddings as input tokens. 
    To train such a model on unlabelled videos, we optimize a self-supervised objective consisting of
    two novel temporal pretext tasks: Out-of-order Frame Localization (OFL), and Time-Varying Skiprate Prediction (TSP). 
    Details of each objective can be found in the Method section.}
    \label{fig:framework}
\end{figure*}

\begin{figure*}[h]
    \centering
    \begin{minipage}{.5\textwidth}
    \centering
    \includegraphics[width=\textwidth]{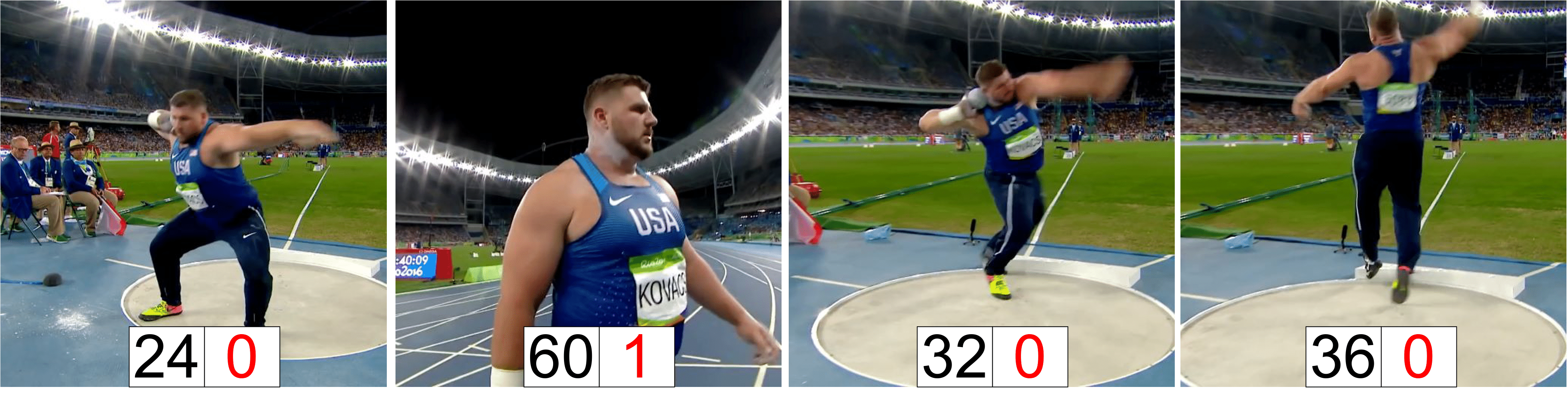}
    \caption{\textbf{Out-of-order Frame Localization (OFL).} 
    The goal is to detect which frames are out of temporal order. 
    The task is posed as a binary classification problem, where 0 indicates in-sequence frames and 1 indicates out-of-order frames. 
    Frame-IDs are shown in black, and the self-supervised targets in \textcolor{red}{red}.
    \label{fig:dto}
    }
    \end{minipage}
    \hfill
    \begin{minipage}{0.48\textwidth}
    \centering
    \includegraphics[width=\textwidth]{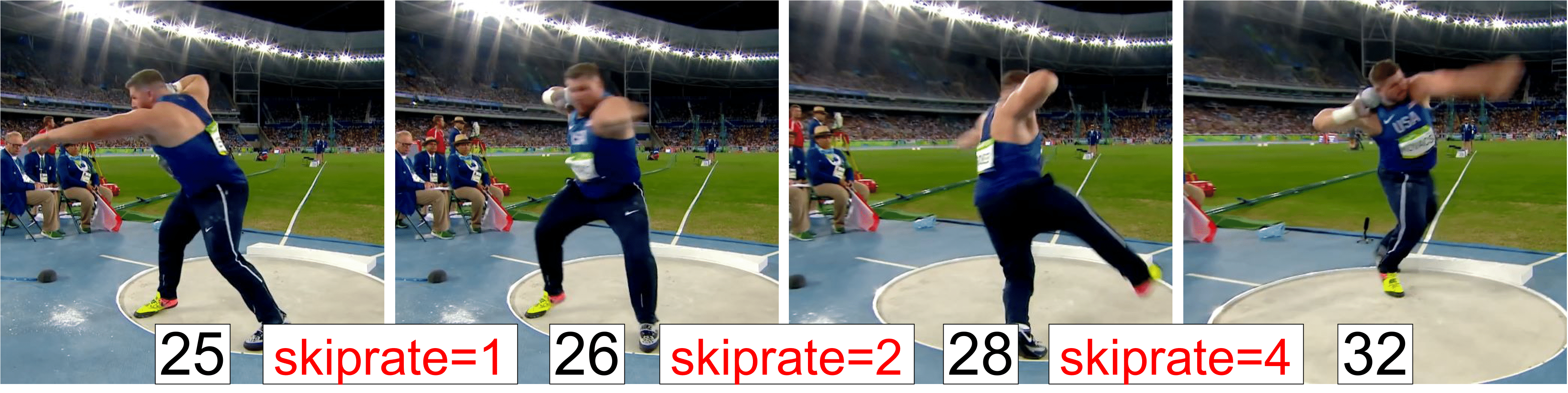}
    \caption{\textbf{Time-Varying Skiprate Prediction (TSP).} 
    The goal is to predict the skiprate between the consecutive frames. 
    We pose the task as an M-way classification problem. 
    In this example, M=3 skiprate classes are used. 
    The skiprate label is shown in \textcolor{red}{red} between consecutive frames.  
    \label{fig:dsp}
    }
    \end{minipage}
\end{figure*}

\section{Prior Work}
\label{sec:priorwork}
Prior work in self-supervised video representation learning is based on a variety of learning objectives.
These include contrastive learning \cite{cvrl,vtc,tclr, timebalance}, pretext tasks \cite{misra2016shuffle,speedNet,aot}, masked video modeling \cite{videomae, stmae}, and hybrid approaches combining multiple objectives \cite{taco,jenni2021time,seco}. 
We compare our results with all these approaches in the experiments but focus the following discussion on related works also relying on temporal self-supervision, which is most relevant to our approach.

\noindent\textbf{Temporal Pretext Tasks.}
These methods are based on recognizing different distortions of the natural temporal evolution of videos. 
Several works have explored \emph{temporal ordering tasks}, which mainly deal with verifying or predicting the temporal order of video frames or short clips.
For example, \cite{misra2016shuffle, seco, taco} posed frame order verification as a prediction task of whether a sequence of 3-frames is in order or not. 
Other works posed temporal ordering as a sorting task, \eg, of 3-frame sequences ~\cite{lee2017unsupervised} or a set of short video clips ~\cite{vcop,corp}.
Other temporal learning signals in videos can be found in the playback direction \cite{aot} (forward vs. backward playback) and the playback speed of the video. 
For example, several works proposed the classification of different artificial playback speeds applied to a video for learning \cite{speedNet,pace_pred}, or additionally recognizing non-uniformly warped videos \cite{simon}.

\emph{How do our task formulations for OFL and TSP differ from prior work?} 
Prior temporal pretext tasks are primarily formulated at the sequence level, \eg, identifying whether the whole sequence of frames is in correct order \cite{misra2016shuffle} or exhibits a normal playback speed \cite{speedNet}. 
In contrast, our formulation is posed as a frame-wise prediction task, where each frame in the sequence is assigned a different label indicating if the frame is in order (OFL) or at what skip rate is applied to it (TSP). 
As our experiments show (Table~\ref{table:cliplevel}), our more challenging per-frame formulations result in significantly better-performing features.
A visual comparison is provided in \supp{Supp. Sec.~\ref{sec:add_method}}. 
\\
\noindent\textbf{Shortcuts in Self-Supervised Learning.}
Many self-supervised learning tasks suffer from trivial solutions exploiting low-level cues in the data to solve the task (shortcuts). 
This hinders the learning of high-level generalizable features. 
For example, chromatic aberration cues were identified as shortcuts in image-based pretext tasks \cite{doersch2015unsupervised} and codec artifacts in video \cite{aot}.
Likewise, contrastive learning approaches rely on strong data augmentation to prevent trivial solutions \cite{simclr}.
\textit{To our best knowledge, we are the first to identify and prevent shortcuts in temporal-pretext tasks due to local patch appearances} (see Figure~\ref{fig:shortcut}). 
As we demonstrate in Table~\ref{table:shortcut} and~\ref{table:ablFramework}, our frame-wise augmentation strategy is effective at preventing this shortcut and considerably improves downstream feature performance.

\section{Method}
\label{sec:method}

Let $\{\mathbf{x_1}, \ldots, \mathbf{x_n} \}$ be set of unlabeled videos where each video consists of a sequence of frames, \ie, $\mathbf{x_i}=[x_{i}^1, \ldots, x_i^{n_i}]$ and $n_i$ defines the number of frames in video $\mathbf{x_i}$.
Let further $\mathbf{x_i}[\mathcal{I}]=[x_{i}^{j}]_{j\in \mathcal{I}}$ denote a sequence of video frames based on frame indices $\mathcal{I}\subset \{1,\ldots,n_i\}$.
In our model, we first extract frame feature vectors $f_i^j=F(x_i^j)\in \mathbb{R}^d$ with an image encoder $f$ for each frame of the sequence, \ie, $F\left(\mathbf{x_i}[\mathcal{I}]\right)=[f_{i}^{j}]_{j\in \mathcal{I}}$.
These feature vectors are then considered as frame-tokens and fed as input to a transformer network $E$ along with learnable position encodings. 
Concretely, the full model is given by $E\left(F\left(\mathbf{x_i}[\mathcal{I}]\right) + \operatorname{PE}_{|\mathcal{I}|}  \right)=[e_i^1, \ldots, e_i^{|\mathcal{I}|}]$, where $\operatorname{PE}_{|\mathcal{I}|}$ is a sequence of $|\mathcal{I}|$ learnable position encodings, and $e_i^j$ denote the set of output tokens after the transformer. 
We will now describe the various self-supervised learning objectives in our framework.

\noindent \textbf{Out-of-order Frame Localization.}
As a first temporal pretext task, we propose to train the frame token transformer $E$ to localize out-of-order frames.
In this task, a subset of the frames fed to the transformer are incorrectly placed, \ie, they have a random time shift applied to them. 
We then train the network $E$ to predict whether each frame in the sequence is correctly placed. 
Since most frames will have a correct placement, the transformer can leverage this global context to infer the correct movement patterns in the sequence and learn to detect frames that do not conform to them. 
We call this pretext task Out-of-order Frame Localization (OFL).

To build example sequences for training, we manipulate the frame sampling indices $\mathcal{I}$. 
A correct sequence during training is considered to be one where 1) all frames are in correct temporal order and 2) there is a constant offset between consecutive frames. 
Concretely, for a video $\mathbf{x_i}$ we sample correct frame indices as $\mathcal{I}=[t, t+\Delta, \ldots, t+(p-1)\Delta ]$, where $p$ denotes the number of frames, $\Delta$ is the fixed frame offset, and $t$ is the starting frame, which is randomly sampled from $t\sim \mathcal{U}(\{1, \ldots,  n_i - (p-1)\Delta\})$.
To build inputs and targets for OFL training, we then sample and replace a random subset of the frame indices with other indices.
Concretely, let $\mathcal{M}\in \{0, 1\}^p$ be a randomly sampled binary sequence indicating for each index in $\mathcal{I}$ whether it should be changed ($\mathcal{M}_i=1$) or kept ($\mathcal{M}_i=0$). 
The ratio of changed indices $\rho = \frac{1}{p}\sum \mathcal{M}_i$ is randomly sampled from the interval $[0, 0.5]$ during training. 
Finally, the transformed indices are constructed as $\hat{\mathcal{I}}=(1-\mathcal{M})\otimes \mathcal{I} + \mathcal{M}\otimes \mathcal{S}$, where $\otimes$ denotes element-wise multiplication and $\mathcal{S}\subset \{1, \ldots, n_i\}$ is a random sequence of other frame indices. 
In our best setting, we restrict the sampling of $\mathcal{S}$ to $\{\min(\mathcal{I})-\Delta, \ldots, \min(\mathcal{I}-1)\} \cup \{\max(\mathcal{I})+1, \ldots, \max(\mathcal{I})+\Delta\}$, \ie, sampling frames before or after $\mathcal{I}$ with a maximum distance of $\Delta$. 
Visual aid in \supp{Supp. Sec.~\ref{sec:add_method}}.

Finally, the OFL objective $\mathcal{L}_{OFL}$ is given by a standard binary cross-entropy loss, \ie,
\begin{equation}
\mathcal{L}_{OFL}=-\smashoperator{\sum_{e_i^j \in E\left(F\left(\mathbf{x_i}[\hat{\mathcal{I}}]\right) \right)}} \mathcal{M}_j \log \sigma(e_i^j) + (1 - \mathcal{M}_j)\log\left(1- \sigma(e_i^j)\right),
\end{equation}
where $\sigma$ indicates a linear layer followed by a sigmoid activation function. An example input sequence and out-of-order frame are shown in Figure~\ref{fig:dto}. We extensively study the design choices of this task in the Ablation section.

\noindent \textbf{Time-varying Skiprate Prediction.}
As a second temporal pretext task, we propose the recognition of time-varying playback speeds (skiprate). 
For this task, we modify the frame indices for sampling model inputs to $\mathcal{I}_s=[t, t+s^1, \ldots, t+\sum_{i=1}^p s^i ]$, where $s^i \sim \mathcal{U}(\{1, 4, 8\})$ are independently sampled skiprates between frames $i$ and $i+1$. 
The task is then to predict the sequence of time-varying skiprates $s^i$ from the input sequence $\mathbf{x_i}[\mathcal{I}_s]$.
Concretely, we model the probability of observing the three different skiprate classes as $\hat{s}^i=\psi(e^{i+1}-e^i) \in \mathbb{R}^3$, where $\psi$ is a linear layer followed by a softmax activation acting on temporal differences of token embeddings. 
The TSP loss $\mathcal{L}_{TSP}$ is then given by a standard 3-way classification loss between the predicted and ground-truth skiprate classes $\hat{s}^i$ and $s^i$.
Example inputs and target outputs are provided in Figure~\ref{fig:dsp}. 
TSP design choices are studied in the Ablation section.

\noindent \textbf{Contrastive Loss}
We utilize two types of frame contrastive losses in our model training. 
The first one is a cross-clip term, wherein positive pairs for learning are built from frames of the \textit{same clip} (or an augmented version of it), and negative pairs are built with frames belonging to \textit{different videos}. 
Concretely, given a training batch $\mathcal{B}$ of the frame features $f_i^j$, the cross-clip loss is given by
\begin{equation}
\mathcal{L}_{C1} = \smashoperator[l]{\sum_{F(\mathbf{x_i}[\mathcal{I}]) \in \mathcal{B}}}  \sum_{j,k \in \mathcal{I}} \log \left( \frac{d\left( f_i^j, \hat{f}_i^k \right)}{ \sum_{f_l^k  \in \mathcal{B}} \mathds{1}\{i \neq l\} d\left( f_i^j, \hat{f}_l^k \right)} \right),
\end{equation}
where $f_i^j$ and $\hat{f}_i^j$ denote features of two differently augmented views of $x_i^j$ and 
\begin{equation}
    d(u_{1}, u_{2}):= \exp \left(\frac{1}{\lambda}  \frac{ { \phi(u_{1})}^\intercal { \phi(u_{2})}}{\Vert {\phi(u_{1})} \Vert_2 \Vert {\phi(u_{2})} \Vert_2} \right),
    \label{eq:sim}
\end{equation}
is a measure of similarity between the feature vectors $u_{1}$ and $u_{2}$, and $\phi$ is a projection implemented via a multilayer perceptron (MLP).

As a second frame-wise loss, we consider within-clip terms, where positive pairs are built solely through different image augmentations of the \textit{same frame}, and negative pairs are constructed with two different frames from the \textit{same clip}.
Formally, this loss term is given by
\begin{equation}
    \mathcal{L}_{C2} = \smashoperator{\sum_{F(\mathbf{x_i}[\mathcal{I}]) \in \mathcal{B}}}  \log \left( \frac{d\left( f_i^j, \hat{f}_i^j \right)}{ \sum_{f_i^k  \in F(\mathbf{x_i}[\mathcal{I}])}  d\left( f_i^j, \hat{f}_i^k \right)} \right).
\end{equation}
\noindent \textbf{Combined Training Objective.}
Finally, we combine our frame-level temporal pretext tasks with the contrastive loss $\mathcal{L}_C$ ($\mathcal{L}_{C1}+ \mathcal{L}_{C2}$).
To summarize, we optimize 
\begin{equation}
    \mathcal{L}_{SSL} = \lambda_O \mathcal{L}_{OFL}+\lambda_T \mathcal{L}_{TSP}+\lambda_C \mathcal{L}_{C},
\end{equation}
where $\lambda_O$, $\lambda_T$, and $\lambda_C$ are loss weights. 
The model architecture and training objectives are illustrated in Figure~\ref{fig:framework}.

\begin{table}[t]
\small
\centering
\begingroup
\setlength{\tabcolsep}{2pt}
\begin{tabular}{lc!{\color[rgb]{0.753,0.753,0.753}\vrule}l!{\color[rgb]{0.753,0.753,0.753}\vrule}l} 
\arrayrulecolor{black}\hline

\hline

\hline\\[-3mm]
\multirow{2}{*}{\textbf{Scale}} & \multirow{2}{*}{\begin{tabular}[c]{@{}c@{}}\textbf{Framewise}\\\textbf{Augment?}\end{tabular}} & \multicolumn{2}{c}{\textbf{Pretext Task Accuracies}}                                           \\
                                &                                                                                                & \textbf{OFL (mAP)}                        & \textbf{TSP (Top-1)}                       \\ 
\hline
Full (224 x 224)                & \multirow{2}{*}{No}                                                                            & 87.6                                      & 57.9                                       \\
Patch (32 x 32)                 &                                                                                                & 78.2 \textcolor{red}{(-10.7\%)}          & 52.3 \textcolor{red}{(-9.6\%)}            \\ 
\hline
Full (224 x 224)                & \multirow{2}{*}{Yes}                                                                           & 84.1                                      & 56.8                                       \\
Patch (32 x 32)                 &                                                                                                & 26.1 \textcolor{red}{(\textbf{-68.9\%})} & 33.1 \textcolor{red}{(\textbf{-41.5\%})}  \\
\hline
\end{tabular}
\endgroup
\caption{\textbf{Evidence for Shortcuts in Temporal Pretext Tasks.} 
We show the pretext task performance of models trained on the full scene ($224 \times 224$) or on only local patches ($32\times32$),  with and without our frame-wise augmentation strategy.  
We observe that a local patch model can achieve high accuracy when consistent augmentations are used, indicating that global  features are not necessary. 
In contrast, only the model with the full scene context is able to achieve non-trivial performance when frame-wise augmentations are used. 
The relative drop in accuracy is shown in  \textcolor{red}{red}. }
\label{table:shortcut}
\end{table}

\noindent \textbf{Avoiding Shortcuts in Temporal Pretext Tasks.}
The recognition of wrongly placed frames in our OFL task should encourage the learning of motion features, \eg, temporal patterns of object deformations. 
Such patterns are crucial for video understanding tasks, \eg, for human action recognition. 
However, we observe that the OFL task can often be solved by comparing the appearance of spatially local image patches in neighboring frames (see Figure~\ref{fig:shortcut} and experimental evidence in Table~\ref{table:shortcut}).
Such low-level task solutions that only consider local spatio-temporal pixel statistics could hamper the learning of higher-level video features that capture the more important changes in object dynamics.
As a solution to this problem, we propose to process the training frame sequences $\mathbf{x_i}[\hat{\mathcal{I}}]$ with spatial jittering applied \textit{independently} to each frame in the sequence. 
Concretely, when augmenting a frame sequence, we first apply a standard augmentation strategy for contrastive learning $\tau_{c}$ (\eg, color jittering, horizontal flipping, etc.), which is applied consistently to all frames and contains only weak spatial cropping.
We then independently apply additional spatial augmentations $\tau_{i}$ (\eg, random resizing and cropping) to each frame. The proposed augmentation for an example $\mathbf{x_i}$ can be expressed as 
\begin{equation}
    \mathbf{\hat{x}_i} = [\tau_1 \circ \tau_c (x_i^1), \ldots, \tau_{n_i} \circ \tau_c (x_i^{n_i})] .
\end{equation}
Such frame-wise augmentations introduce large differences in the local frame patch appearances and force the model to consider more global spatio-temporal features (\eg, object dynamics) to solve the task.

\section{Experiments}

In this section, we first provide details about datasets, implementation, and experimental setup. We perform ablation studies and evaluate our SSL representation on various downstream tasks.

\noindent  \textbf{Datasets.}
We use the following set of established video benchmarks in our experiments:

\noindent\textbf{UCF101}~\cite{ucf101} is human action dataset containing 101 classes of indoor and outdoor actions. 

\noindent\textbf{HMDB51}~\cite{hmdb} is a relatively small-scale dataset of 51 action classes with high intra-class diversity.

\noindent\textbf{Kinetics400}~\cite{kinetics} is a large-scale action dataset containing 400 human activity classes collected from YouTube.

\noindent\textbf{Something-Something V2 (SSv2)}~\cite{goyal2017something} consists of over 174 action classes, providing a challenging and diverse set of actions and environmental contexts. %

\noindent\textbf{NTU60}~\cite{ntu60}  is a large-scale benchmark for human action recognition, containing over 56,000 action samples with 60 diverse action classes performed by multiple subjects in real-world indoor and outdoor environments.%

\noindent\textbf{Charades}~\cite{charades} is a multi-label action dataset containing 157 daily-life indoor actions in untrimmed videos. 

\noindent \textbf{Holistic Video Understanding (HVU)}~\cite{hvu} is a large-scale benchmark addressing multi-label and multi-task video understanding of multiple semantic aspects, including scenes, objects, actions, attributes, and concepts.

\noindent\textbf{DAVIS-2017}~\cite{davis} provides object-level pixel-wise annotations. The evaluation set contains 57 different objects. 

\noindent\textbf{CASIA-B}~\cite{casia} is a gait-recognition dataset for indoor walking videos of 124 subjects with 11 views from each. 

\noindent\textbf{JHMDB Pose}~\cite{jhmdb21} provides 31,838 annotated frames with 13 joints (shoulder, elbow, knee, etc.).
\\ 
More dataset details can be found in \supp{Supp. Sec.~\ref{sec:dataset}}.

\noindent \textbf{Implementation Details.}
Our framework is built on the Video Transformer Network (VTN)~\cite{vtn} architecture. 
In our default experimental setting, we utilize a Vision Transformer (ViT)~\cite{vit} network as our image encoder $F(\cdot)$. 
We perform our self-supervised pertaining on unlabelled videos of Kinetics400.
As inputs to our network, we feed 8 frames of resolution $224\times224$. 
During training, we use the common set of geometric augmentations (random crop, resize, flipping) and color jittering (random grayscale, color jittering, random erasing). 
More implementation details can be found in \supp{Supp. Sec.~\ref{sec:implementation}}.

\noindent \textbf{Experimental Setup.}
We focus our experiments on \emph{fixed-feature} evaluation, \ie, we keep the learned video representations fixed and evaluate features via retrieval and linear probing experiments. 
This focus is motivated by 1) the relevance to downstream video search applications, 2) the more direct probing of properties in the learned features, and 3) the better scalability of this approach to large-scale video processing.
While we acknowledge that superior performance can often be achieved through a full finetuning of the network, it is often infeasible to fine-tune networks and reprocess massive amounts of high-dimensional video data in practice.
In contrast, training and inference with a shallow model (\eg, a linear classifier) on pre-extracted video features are considerably more scalable.

\begin{table}
\centering
\small
\begingroup
\setlength{\tabcolsep}{3pt}
\begin{tabular}{cccccccc} 
\arrayrulecolor{black}\hline

\hline

\hline\\[-3mm]
\multicolumn{1}{l}{} & \textbf{OFL} & \textbf{TSP} & \multirow{2}{*}{\begin{tabular}[c]{@{}c@{}}\textbf{Framewise }\\\textbf{Augment}\end{tabular}} & \multicolumn{2}{c}{\textbf{UCF101}}           & \multicolumn{2}{c}{\textbf{HMDB51 }}           \\
\multicolumn{1}{l}{} & $\mathcal{L}_{OFL}$           & $\mathcal{L}_{TSP}$           &                                                                                                & \textbf{\textbf{R@1}} & \textbf{\textbf{Lin}} & \textbf{\textbf{R@1}} & \textbf{\textbf{Lin}}  \\ 
\hline
\texttt{init}                  & \textcolor[rgb]{0.7,0.7,0.7}{\xmark}           & \textcolor[rgb]{0.7,0.7,0.7}{\xmark}            &  \textcolor[rgb]{0.7,0.7,0.7}{\xmark}                                                                                               & 79.20                 & 85.11                 & 41.44                 & 50.20                  \\
\texttt{(a)}                  & \xmark            & \xmark            & \textcolor[rgb]{0.7,0.7,0.7}{\cmark}                                                                                              & 80.15                 & 85.31                 & 43.35                 & 52.42                  \\
\texttt{(b)}                  & \xmark            & \textcolor[rgb]{0.7,0.7,0.7}{\cmark}            & \textcolor[rgb]{0.7,0.7,0.7}{\cmark}                                                                                              & 82.73                 & 88.12                 & 46.14                 & 54.22                  \\
\texttt{(c)}                  & \textcolor[rgb]{0.7,0.7,0.7}{\cmark}            & \xmark            & \textcolor[rgb]{0.7,0.7,0.7}{\cmark}                                                                                              & 83.90                 & 89.29                 & 48.98                 & 58.28                  \\
\texttt{(d)}                  & \textcolor[rgb]{0.7,0.7,0.7}{\cmark}            & \textcolor[rgb]{0.7,0.7,0.7}{\cmark}            & \xmark                                                                                              & 82.50                 & 87.80                 & 46.51                 & 55.60                  \\ 
\hline
\texttt{(e)}                  & \textcolor[rgb]{0.7,0.7,0.7}{\cmark}            & \textcolor[rgb]{0.7,0.7,0.7}{\cmark}            & \textcolor[rgb]{0.7,0.7,0.7}{\cmark}                                                                                              & 84.68                 & 89.90                 & 50.20                 & 58.70                  \\
\hline
\end{tabular}
\endgroup
\caption{\textbf{SSL Objectives Ablation Experiments.} We performed experiments to investigate the effect of the various loss terms in our model (\texttt{a-d}) and the frame-wise augmentation strategy to remove shortcuts (\texttt{e}).}
\label{table:ablFramework}

\end{table}

\subsection{Ablations}
\label{sec:ablation}
We perform extensive ablations to verify our frame-wise pretext task formulation and illustrate the importance of shortcut removal for temporal self-supervision. 
All ablations are performed with UCF101 pretraining, and we report results with linear probing and nearest-neighbor retrieval. 
More ablation studies can be found in \supp{Supp. Sec.~\ref{sec:add_ablations}}.
\\

\noindent\textbf{Influence of No-Shortcuts Temporal SSL.} 
In Table~\ref{table:ablFramework}, we analyze the influence of the temporal pretext tasks $\mathcal{L}_{OFL}$ and $\mathcal{L}_{TSP}$ and illustrate the importance of avoiding shortcuts in temporal SSL through our frame-wise augmentation strategy.
We also report the performance when using the image SSL pre-trained backbone only (\texttt{init}) and when using contrastive learning on videos only \texttt{(a)}.
We observe only minor improvements from further contrastive training on videos in \texttt{(a)}.
In contrast, both our temporal pretext tasks ($\mathcal{L}_{OFL}$ and $\mathcal{L}_{TSP}$) \texttt{(b)-(c)} contribute significantly to downstream performance, increasing accuracy by \textbf{4-7\%} across downstream tasks over the contrastive baselines.
This highlights the importance of capturing temporal features for video representation learning. 
Finally, in \texttt{(d)}, we see that removing the framewise augmentation strategy reduces the performance by a significant \textbf{2-4\%}. 
This illustrates our key insight that temporal SSL methods have been plagued by shortcuts and did not achieve their full potential.

\begin{table}[h]
\centering
\small
\begingroup
\setlength{\tabcolsep}{1pt}
\arrayrulecolor[rgb]{0.502,0.502,0.502}
\arrayrulecolor[rgb]{0.753,0.753,0.753}
\begin{tabular}{l!{\color{black}\vrule}c|c!{\color{black}\vrule}c|c} 
\arrayrulecolor{black}\hline

\hline

\hline\\[-3mm]
\multirow{2}{*}{\textbf{Method}} & \multicolumn{2}{c!{\color{black}\vrule}}{\textbf{UCF101}} & \multicolumn{2}{c}{\textbf{HMDB51}}  \\
                                 & \textbf{R@1} & \textbf{Lin}         & \textbf{R@1} & \textbf{Lin}          \\ 
\hline
Clip-level            & 80.85        & 86.42                & 44.85        & 54.54                 \\
Frame-level (OFL)              & 83.90\increase{4\%}        & 89.29\increase{3\%}                & 48.98\increase{9\%}        & 58.28\increase{7\%}                 \\
\hline
Clip-level                    & 81.35        & 86.71                & 44.85        & 52.47                 \\
Frame-level (TSP)         & 82.73\increase{2\%}        & 88.12\increase{2\%}                & 46.14\increase{3\%}        & 54.22\increase{3\%}                 \\
\hline
\end{tabular}
\endgroup
\caption{Comparison with clip-level temporal tasks}
\label{table:cliplevel}
\end{table}

\noindent\textbf{Comparing Clip-Level vs. Frame-Level Pretext Tasks.}
In this study, we compare our frame-wise reformulation of temporal pretext tasks with the traditional global (\ie, clip-level) task formulations. 
A conceptual comparison to these clip-level tasks is given in \supp{Supp. Sec.~\ref{sec:conceptual_compare}}. 
In these experiments, we only incorporate either OFL or TSP in our training objective. 
The first two rows of Table~\ref{table:cliplevel} compare clip-level frame verification vs. our frame-level OFL task. 
We can see clear performance gains across all downstream protocols going from a clip-level task to a frame-level task. The last two rows of Table~\ref{table:cliplevel} also suggest similar conclusions for the clip-level vs. frame-level skip rate prediction task.
We also find that transitioning from clip-level to frame-level tasks significantly raises the difficulty of the pretext task, as indicated by the respective pretext accuracies of 99\% vs. 87\% for order-verification, and 96\% vs. 55\% for the skip-prediction (numbers not shown in Table~\ref{table:cliplevel}).

\noindent\textbf{Out-of-order Frame Localization.} 
OFL has two main design parameters to explore: (1) the percentage of out-of-order frames in the frame sequence, and (2) from where in the video to sample out-of-order frames (\ie, how far from the correct position). 
We also report pretext task performance in terms of mAP to indicate the difficulty of OFL. 
In Table~\ref{table:dto_amount}, we compare fixed outlier rates \texttt{(b)-(c)} to sampling outlier rates from a given interval at random \texttt{(d)-(e)}.
We find that randomizing the outlier rate provides clear benefits.
Note the negative correlation between SSL and downstream performance, which validates our aim to increase the difficulty of temporal SSL.
We use unrestricted sampling (\ie, Table~\ref{table:dto_distance} \texttt{(a)}) in this experiment. 
\begin{table}[h]
\centering
\small
\begingroup

\setlength{\tabcolsep}{4pt}
\begin{tabular}{llccccc} 
\arrayrulecolor{black}
\hline

\hline

\hline\\[-3mm]
&\multirow{2}{*}{\begin{tabular}[c]{@{}l@{}}\textbf{Probability of }\\\textbf{outlier token}\end{tabular}} & \multirow{2}{*}{\begin{tabular}[c]{@{}c@{}}\textbf{OFL}\\\textbf{Task}\end{tabular}} & \multicolumn{2}{c}{\textbf{UCF101}} & \multicolumn{2}{c}{\textbf{HMDB51}}  \\
                                                   &                                                            &                                                                                     & \textbf{R@1} & \textbf{Lin}         & \textbf{R@1} & \textbf{Lin}          \\ 
\hline
\texttt{(a)} & 0.0                                                                                                            & -                                                                                   & 82.73        & 88.12                & 46.14        & 54.22                 \\
\texttt{(b)} & 0.25                                                                                                          & 87.39                                                                               & 83.40        & 88.89                & 47.30        & 55.91                 \\
\texttt{(c)} & 0.50                                                                                                           & 79.88                                                                               & 83.10        & 88.40                & 46.80        & 55.10                 \\
\texttt{(d)} & $\mathcal{U}([0.0, 0.25])$                                                                                                      & 81.15                                                                               & 83.70        & 89.14                & 47.90        & 56.19                 \\
\texttt{(e)} & $\mathcal{U}([0.0, 0.50])$                                                                                                      & 76.76                                                                               & \textbf{84.01}        & \textbf{89.49}                & \textbf{48.76}        & \textbf{57.87}                 \\
\hline
\end{tabular}
\endgroup

\caption{Ablations of outlier token probability in OFL task.}

\label{table:dto_amount}
\end{table}

In Table~\ref{table:dto_distance}, we explore the sampling position of outlier frames.
We find that both a too-simple \texttt{(b)} and a too-difficult and potentially ambiguous OFL task \texttt{(c)} result in poor downstream performance. 
The best results are achieved by sampling the out-of-order frames within 64 frames of the in-order position.

\begin{table}[h]
\centering
\small
\begingroup
\setlength{\tabcolsep}{3pt}
\arrayrulecolor[rgb]{0.502,0.502,0.502}
\resizebox{\linewidth}{!}{%
\begin{tabular}{llccccc} 
\arrayrulecolor{black}\hline

\hline

\hline\\[-3mm]
&\multirow{2}{*}{\begin{tabular}[c]{@{}l@{}}\textbf{Out of Order frame}\\\textbf{Sampling Restriction}\end{tabular}} & \multirow{2}{*}{\begin{tabular}[c]{@{}c@{}}\textbf{OFL}\\\textbf{Task}\end{tabular}} & \multicolumn{2}{c}{\textbf{UCF101}} & \multicolumn{2}{c}{\textbf{HMDB51}}  \\
                                                                                                               & &                                                                                       & \textbf{R@1} & \textbf{Lin}         & \textbf{R@1} & \textbf{Lin}          \\ 
\hline
\texttt{(a)} & Unrestricted                                                                                                   & 76.76                                                                                 & 84.01        & 89.49                & 48.76        & 57.87                 \\
\texttt{(b)} & Min. Distance = 8                                                                                              & 81.26                                                                                 & 83.27        & 87.67                & 47.52        & 56.50                 \\
\texttt{(c)} & Max. Distance = 8                                                                                              & 61.21                                                                                 & 83.79        & 88.41                & 47.93        & 56.68                 \\
\texttt{(d)} & Max. Distance = 64                                                                                             & 74.37                                                                               & \textbf{84.68}        & \textbf{89.90}                & \textbf{50.20}        & \textbf{58.70}                 \\
\hline
\end{tabular}

}
\endgroup
\caption{Abl of replacement frame sampling in OFL task.}

\label{table:dto_distance}

\end{table}

\noindent\textbf{Time-varying Skiprate Prediction.}
We explore the design of TSP in Table~\ref{table:dsp}, by using different subsets of $\{1,2,4,8\}$ as skip rates for TSP. We observe that going from 2-way classification \texttt{(b)-(c)} to 3-way classification \texttt{(d)-(e)} consistently improves performance.
This again suggests that more challenging pretext task formulations help across downstream tasks. 
\begin{table}[h]
\centering
\small
\begingroup
\setlength{\tabcolsep}{4pt}
\arrayrulecolor[rgb]{0.502,0.502,0.502}
\begin{tabular}{llccccc} 
\arrayrulecolor{black}\hline

\hline

\hline\\[-3mm]
& \multirow{2}{*}{\begin{tabular}[l]{@{}l@{}}\textbf{Playback}\\\textbf{Set}\end{tabular}} & \multirow{2}{*}{\begin{tabular}[c]{@{}c@{}}\textbf{TSP}\\\textbf{Task Acc. }\end{tabular}} & \multicolumn{2}{c}{\textbf{UCF101}} & \multicolumn{2}{c}{\textbf{HMDB51}}  \\
                                    &  &                                                                                                             & \textbf{R@1}   & \textbf{Lin}       & \textbf{R@1}   & \textbf{Lin}        \\ 
\hline
\texttt{(a)} & $\Phi$                                 & -                                                                                                             & 83.09          & 88.05              & 47.58          & 56.70               \\
\texttt{(b)} &\{1,4\}                             & 72.10                                                                                                         & 83.62          & 88.64              & 48.30          & 57.43               \\
\texttt{(c)} &\{2,4\}                             & 71.55                                                                                                         & 83.53          & 88.76              & 48.30          & 57.55               \\
\texttt{(d)} & \textbf{\{1,4,8\}}                  & 55.14                                                                                                         & \textbf{84.01} & \textbf{89.49}     & 48.76          & 57.87               \\
\texttt{(e)} & \{2,4,8\}                           & 54.66                                                                                                         & 84.00          & 89.23              & \textbf{48.81} & \textbf{57.92}      \\
\hline
\end{tabular}
\endgroup

\caption{Ablations of various skip rates in TSP task.}
\label{table:dsp}

\end{table}

\noindent\textbf{Different Frame-wise Augmentations.}
We proposed using frame-wise augmentation to break the local patch similarity-based shortcuts in temporal pretext tasks. In this study, we compare the performance of various frame-wise augmentations. From Table~\ref{table:framewise_aug}, we can observe that 
adding color jittering to the random cropping decreases the performance by a small margin, however, adding the frame-wise horizontal flipping reduces the performance noticeably by 1-2\%. In our default setting, we only use frame-wise cropping. 
\begin{table}
\centering
\small

\begingroup
\setlength{\tabcolsep}{3pt}
\begin{tabular}{cccrrrr}
\arrayrulecolor{black}\hline

\hline

\hline\\[-3mm]
\multirow{2}{*}{\begin{tabular}[c]{@{}c@{}}\textbf{Spatial}\\\textbf{Cropping}\end{tabular}} & \multirow{2}{*}{\begin{tabular}[c]{@{}c@{}}\textbf{Color}\\\textbf{Jittering}\end{tabular}} & \multirow{2}{*}{\begin{tabular}[c]{@{}c@{}}\textbf{Horizontal}\\\textbf{Flipping}\end{tabular}} & \multicolumn{2}{c}{\textbf{UCF101}}                                 & \multicolumn{2}{c}{\textbf{HMDB51}}                                  \\
                                                                                             &                                                                                             &                                                                                                 & \multicolumn{1}{c}{\textbf{R@1}} & \multicolumn{1}{c}{\textbf{Lin}} & \multicolumn{1}{c}{\textbf{R@1}} & \multicolumn{1}{c}{\textbf{Lin}}  \\ 
\hline
                                                                                        \cmark                                                                                             & \textcolor[rgb]{0.7,0.7,0.7}{\xmark}                                                                                                     & \textcolor[rgb]{0.7,0.7,0.7}{\xmark}                                                        & \textbf{84.68}                            & \textbf{89.90}                            & \textbf{50.20}                            & \textbf{58.70}                             \\
\cmark                                                                                             & \cmark                                                                                                    & \textcolor[rgb]{0.7,0.7,0.7}{\xmark}                                                                             & \multicolumn{1}{l}{84.24}        & \multicolumn{1}{l}{89.30}        & \multicolumn{1}{l}{49.50}        & \multicolumn{1}{l}{57.81}         \\
                                                                                    \cmark                                                                                             & \textcolor[rgb]{0.7,0.7,0.7}{\xmark}                                                                                                     & \cmark                                                                                                & 84.04                            & 88.97                            & 48.08                            & 56.61                             \\
\hline
\end{tabular}
\endgroup
\caption{Ablation for frame-wise augmentations}
\label{table:framewise_aug}
\end{table}

\noindent\textbf{Static and Temporal Feature Analysis.}
Since our model learns and disentangles static and temporal features, we explore the influence of temporal features on our fused video representation in  Table~\ref{table:features} (Kinetics400 pretraining).
While static appearance features achieve a strong baseline performance, temporal features lead to consistent and significant improvements. 
\begin{table}[h]
\centering
\small
\begingroup
\setlength{\tabcolsep}{6pt}
\begin{tabular}{lcccc} 
\arrayrulecolor{black}\hline

\hline

\hline\\[-3mm]
\multirow{2}{*}{\textbf{Features}} & \multicolumn{2}{c}{\textbf{UCF101}} & \multicolumn{2}{c}{\textbf{HMDB51}}  \\
                                   & \textbf{R@1} & \textbf{Lin}         & \textbf{R@1} & \textbf{Lin}          \\ 
\hline
Static only                        & 84.53        & 89.11                & 50.91        & 55.90                 \\
Static + Temporal                     & \textbf{86.22}        & \textbf{91.50}                & \textbf{52.61}        & \textbf{62.50}                 \\
\hline
\end{tabular}

\endgroup
\caption{Influence of temporal features.}
\label{table:features}
\end{table}

\noindent\textbf{Different Backbones and Initializations.}
Following the prior work~\cite{svt, seco, mcl}, we use self-supervised weights learned on ImageNet-1k~\cite{deng2009imagenet} as initialization for the image encoder. We compare various image encoders and their initialization methods in Table~\ref{table:init}. \texttt{(a)} In our default setting, we report results with ViT-L initialized with MUGS~\cite{mugs2022SSL} pre-training on ImageNet-1k. \texttt{(b)} We get similar/slightly better results using iBOT~\cite{zhou2021ibot} pre-training. \texttt{(c)} Our results improve by 2-4\% by using the iBOT pre-training on ImageNet-21k, which shows the advantage of our method to leverage the improvement in the image self-supervised methods. \texttt{(d,e)} show our results with ViT-B backbone initialized by MUGS~\cite{mugs2022SSL} and DINO~\cite{dino} pre-training on Imagenet-1k. Lastly, in\texttt{(f)}, we show similar results to our default setting with computationally efficient SWIN-B backbone initialized with EsViT~\cite{li2021esvit} pre-training. 
\begin{table}[h]
\centering
\small
\begingroup
\setlength{\tabcolsep}{3pt}
\begin{tabular}{lllcccc} 
\arrayrulecolor{black}\hline

\hline

\hline\\[-3mm]
\multirow{2}{*}{} & \multirow{2}{*}{\textbf{Arch.}} & \multirow{2}{*}{\begin{tabular}[c]{@{}l@{}}\textbf{Pretraining}\\\textbf{Method}\end{tabular}} & \multicolumn{2}{c}{\textbf{UCF101}} & \multicolumn{2}{c}{\textbf{HMDB51}}  \\
                  &                                    &                                                                                                & \textbf{R@1} & \textbf{Lin}         & \textbf{R@1} & \textbf{Lin}          \\ 
\hline
\texttt{(a)}                  & ViT-L                              & MUGS                                                                                           & 84.68        & 89.90                & 50.20        & 58.70                 \\
\texttt{(b)}                  & ViT-L                              & iBOT                                                                                           & 84.52        & 89.81                & 51.21        & 59.45                 \\
\texttt{(c)}                  & ViT-L                              & iBOT (21k)                                                                                   & 86.40        & 91.60                & 54.37        & 63.30                 \\
                  \hline
\texttt{(d)}                  & ViT-B                              & MUGS                                                                                           & 84.08        & 89.33                & 48.43        & 58.20                 \\
\texttt{(e)}                  & ViT-B                              & DINO                                                                                           & 84.90        & 90.71                & 45.81        & 57.97                 \\
                  \hline
\texttt{(f)}                  & SWIN-B                             & EsViT                                                                                          & 84.25        & 89.64                & 46.73        & 58.11                 \\
\hline
\end{tabular}

\endgroup
\caption{Different initialization and backbone}
\label{table:init}
\end{table}

\newcolumntype{"}{@{\hskip\tabcolsep\vrule width 0.6pt\hskip\tabcolsep}}

\begin{table*}[t]
\small
\centering

\begingroup
\setlength{\tabcolsep}{3.5pt}
\arrayrulecolor[rgb]{0.753,0.753,0.753}

\begin{tabular}{l!{\color{black}\vrule}c|c|c|c|c|c!{\color{black}\vrule}c|c!{\color{black}\vrule}c|c|c|c} 
\arrayrulecolor{black}\hline

\hline

\hline\\[-3mm]
\multirow{3}{*}{\textbf{Method}}      & \multicolumn{6}{c!{\color{black}\vrule}}{\textbf{Action Linear Classification}}                               & \multicolumn{2}{c!{\color{black}\vrule}}{\textbf{Action Finetuning}} & \multicolumn{4}{c}{\textbf{Video Retrieval}}                                \\ 
\arrayrulecolor[rgb]{0.753,0.753,0.753}\cline{2-13}
                                      & \textbf{U101} & \textbf{H51} & \textbf{K400} & \textbf{SSv2} & \textbf{NTU60} & \textbf{Charades} & \textbf{UCF101} & \textbf{HMDB51}                                    & \multicolumn{2}{c|}{\textbf{UCF101}} & \multicolumn{2}{c}{\textbf{HMDB51}}  \\
                                      & Top-1        & Top-1        & Top-1             & Top-1      & Top-1       & mAP (\%)          & Top-1        & Top-1                                           & R@1  & R@5                           & R@1  & R@5                           \\ 
\arrayrulecolor{black}\hline
MoCo~\venue{(CVPR-20)~\cite{moco}}                   & 65.4            & -               & 34.5                 & 7.4           & 16.0             & 8.1               & 83.5            & -                                                  & -    & -                             & -    & -                             \\
{VCOP}~\venue{(CVPR-20)~\cite{vcop}}              & -               & -               & -                    & -             & -              & -                 & 64.9              & 29.5                                               & 14.1 & 30.3                          & - & -                          \\
Pace Pred~\venue{(ECCV-20)~\cite{pace_pred}}              & -               & -               & -                    & -             & -              & -                 & 68.0              & 36.6                                               & 25.6 & 42.7                          & 12.9 & 31.6                          \\
SeLaVi~\venue{(NeurIPS-20)~\cite{selavi}}              & 51.2            & -               & 24.1                 & 4.5           & 15.7           & 8.2               & 84.9            & -                                                  & 52.0   & 68.6                          & 24.8 & 47.6                          \\
CoCLR-RGB~\venue{(NeurIPS-20)~\cite{coclr}}           & 74.5            & 46.1            & -                    & -             & -              & -                 & 87.9            & 54.6                                               & 53.3 & 69.4                          & 23.2 & 43.2                          \\
{DSM}~\venue{(AAAI-21)~\cite{dsm}}                   & -            & -           & -                 & -             & -              & -                 & 70.3            & 40.5                                               & 16.8    & 33.4                             & 8.2    & 25.9                             \\
SeCO~\venue{(AAAI-21)~\cite{seco}}                   & 79.6            & 42.2            & 61.9                 & -             & -              & -                 & 88.3            & 55.6                                               & -    & -                             & -    & -                             \\
{SimSiam}~\venue{(CVPR-21)~\cite{simsiam}}                   & -            & -            & -                 & -             & -              & -                 & -            & -                                               & 39.0    & 53.1                             & 17.1    & 37.3                             \\

CVRL~\venue{(CVPR-21)~\cite{cvrl}}                   & 89.8            & 58.3            & 66.1                 & -             & -              & -                 & 92.9            & 67.9                                               & -    & -                             & -    & -                             \\
$\rho$BYOL~\venue{(CVPR-21)~\cite{Feichtenhofer_2021_CVPR}}& 90.1            & \secondbest{61.1}            & \bestresult{68.3}                 & \secondbest{24.5}          & 51.2           & 18.1              & 94.2            & \secondbest{72.1}                                               & 76.8 & 87.1                          & 39.6 & 64.1                          \\
VidMoCo~\venue{(CVPR-21)~\cite{videomoco}}                & 66.3            & -               & 31.0                   & 19.5          & \secondbest{51.6}           & 10.5              & 78.7            & 49.2                                               & -    & -                             & -    & -                             \\
CORP~\venue{(ICCV-21)~\cite{corp}}                    & 90.2            & 58.7            & 66.6                    & -             & -              & -                 & 93.5            & -                                               & - & -                          & - & -                          \\
MFO~\venue{(ICCV-21)~\cite{mfo}}                    & 63.2            & 33.4            & -                    & -             & -              & -                 & 79.1            & 47.6                                               & 39.6 & 57.6                          & 18.8 & 39.2                          \\
TECon~\venue{(ICCV-21)~\cite{jenni2021time}}                  & 74.1            & 47.5            & -                    & -             & -              & -                 & 83.7            & 60.8                                               & 64.3 & 80.9                          & 29.5 & 55.8                          \\
Vi2CLR~\venue{(ICCV-21)~\cite{vi2clr}}                 & 75.4            & 47.3            & 63.4                 & -             & -              & -                 & 89.1            & 55.7                                               & 55.4 & 70.9                          & 24.6 & 45.1                          \\
MCL~\venue{(ICCV-21)~\cite{mcl}}                    & 79.9            & -               & -                    & -             & -              & -                 & 90.5            & 63.5                                               & 67.0 & 80.8                          & 26.7 & 52.5                          \\
CtP~\venue{(ICCV-21)~\cite{ctp}}                    & 37.9            & -               & 7.6                  & 12.2          & 22.6           & 9.6               & 88.4            & 61.7                                               & 29.0 & 47.3                          & 11.8 & 30.1                          \\
\textcolor[rgb]{0.502,0.502,0.502}{GDT~\venue{(ICCV-21)~\cite{gdt}}+ A} & \textcolor[rgb]{0.502,0.502,0.502}{75.7} & \textcolor[rgb]{0.502,0.502,0.502}{-} & \textcolor[rgb]{0.502,0.502,0.502}{38.6} & \textcolor[rgb]{0.502,0.502,0.502}{11.9} & \textcolor[rgb]{0.502,0.502,0.502}{38.2} & \textcolor[rgb]{0.502,0.502,0.502}{8.5} & \textcolor[rgb]{0.502,0.502,0.502}{89.3} & \textcolor[rgb]{0.502,0.502,0.502}{60.0} & \textcolor[rgb]{0.502,0.502,0.502}{62.8} & \textcolor[rgb]{0.502,0.502,0.502}{79.0} & \textcolor[rgb]{0.502,0.502,0.502}{26.1} & \textcolor[rgb]{0.502,0.502,0.502}{51.7} \\
CPNet~\venue{(AAAI-22)~\cite{cpnet}}                  & -               & -               & -                    & -             & -              & -                 & 83.8            & 57.1                                               & 35.3 & 49.9                          & 14.0   & 32.8                          \\
TCLR~\venue{(CVIU-22)~\cite{tclr}}                   & 69.9            & -               & 19.9                 & 10.9          & 33.5           & 11.1              & 84.1            & 53.6                                               & 56.9 & 72.2                          & 24.1 & 45.8                          \\
\textcolor[rgb]{0.502,0.502,0.502}{TransRank~\venue{(CVPR-22)~\cite{transrank}}+ D} & \textcolor[rgb]{0.502,0.502,0.502}{} & \textcolor[rgb]{0.502,0.502,0.502}{-} & \textcolor[rgb]{0.502,0.502,0.502}{-} & \textcolor[rgb]{0.502,0.502,0.502}{-} & \textcolor[rgb]{0.502,0.502,0.502}{-} & \textcolor[rgb]{0.502,0.502,0.502}{-} & \textcolor[rgb]{0.502,0.502,0.502}{89.6} & \textcolor[rgb]{0.502,0.502,0.502}{63.5} & \textcolor[rgb]{0.502,0.502,0.502}{54.0} & \textcolor[rgb]{0.502,0.502,0.502}{71.8} & \textcolor[rgb]{0.502,0.502,0.502}{25.5} & \textcolor[rgb]{0.502,0.502,0.502}{52.3} \\
SLIC~\venue{(CVPR-22)~\cite{slic}}                   & 72.3            & 41.8            & -                    & -             & -              & -                 & 83.2            & 52.2                                               & 66.7 & 77.3                          & 25.3 & 49.8                          \\
SVT~\venue{(CVPR-22)~\cite{svt}}                    & \secondbest{90.8}            & 57.8            & \secondbest{68.1}                 & 18.3          & 50.8           & \secondbest{18.8}              & 93.7            & 67.2                                               & \secondbest{82.9} & \secondbest{88.0}                          & \secondbest{44.4} & \secondbest{67.4}                          \\
\textcolor[rgb]{0.502,0.502,0.502}{MSCL~\venue{(ECCV-22)~\cite{mscl}}+ F} & \textcolor[rgb]{0.502,0.502,0.502}{88.7} & \textcolor[rgb]{0.502,0.502,0.502}{56.5} & \textcolor[rgb]{0.502,0.502,0.502}{-} & \textcolor[rgb]{0.502,0.502,0.502}{-} & \textcolor[rgb]{0.502,0.502,0.502}{-} & \textcolor[rgb]{0.502,0.502,0.502}{-} & \textcolor[rgb]{0.502,0.502,0.502}{91.5} & \textcolor[rgb]{0.502,0.502,0.502}{62.8} & \textcolor[rgb]{0.502,0.502,0.502}{65.6} & \textcolor[rgb]{0.502,0.502,0.502}{80.3} & \textcolor[rgb]{0.502,0.502,0.502}{28.9} & \textcolor[rgb]{0.502,0.502,0.502}{56.2} \\
\textcolor[rgb]{0.502,0.502,0.502}{MaCLR~\venue{(ECCV-22)~\cite{maclr}}+ F} & \textcolor[rgb]{0.502,0.502,0.502}{91.5} & \textcolor[rgb]{0.502,0.502,0.502}{63.0} & \textcolor[rgb]{0.502,0.502,0.502}{-} & \textcolor[rgb]{0.502,0.502,0.502}{-} & \textcolor[rgb]{0.502,0.502,0.502}{-} & \textcolor[rgb]{0.502,0.502,0.502}{-} & \textcolor[rgb]{0.502,0.502,0.502}{94.0} & \textcolor[rgb]{0.502,0.502,0.502}{67.4} & \textcolor[rgb]{0.502,0.502,0.502}{73.4} & \textcolor[rgb]{0.502,0.502,0.502}{} & \textcolor[rgb]{0.502,0.502,0.502}{-} & \textcolor[rgb]{0.502,0.502,0.502}{-}
    \\
    
VideoMAE~\venue{(NeuRIPS-22)~\cite{videomae}}            & 84.6            & 60.5            & 61.2                 & 23.1          & 51.2           & 15.6              & \bestresult{96.1}            & \bestresult{73.3}                                               & 64.0 & 81.0                          & 32.5 & 58.9                          \\
\textcolor[rgb]{0.502,0.502,0.502}{AVCL~\venue{(AAAI-23)~\cite{jenni2023audio}}+ A} & \textcolor[rgb]{0.502,0.502,0.502}{88.0} & \textcolor[rgb]{0.502,0.502,0.502}{58.2} & \textcolor[rgb]{0.502,0.502,0.502}{-} & \textcolor[rgb]{0.502,0.502,0.502}{-} & \textcolor[rgb]{0.502,0.502,0.502}{-} & \textcolor[rgb]{0.502,0.502,0.502}{-} & \textcolor[rgb]{0.502,0.502,0.502}{91.8} & \textcolor[rgb]{0.502,0.502,0.502}{71.2} & \textcolor[rgb]{0.502,0.502,0.502}{70.7} & \textcolor[rgb]{0.502,0.502,0.502}{-} & \textcolor[rgb]{0.502,0.502,0.502}{40.5} & \textcolor[rgb]{0.502,0.502,0.502}{-}\\
TubeletCon~\venue{(ICCV-23)~\cite{thoker2023tubelet}}             & -               & -               & -                    & -             & -              & 10.3              & 91.0            & 64.1                                               & -    & -                             & -    & -                             \\ 
\hline
Ours (ViT-B)                          & 91.0            & 60.8            & 68.2                 & 27.0          & 52.3           & 20.1              & 94.2            & 64.1                                               & 85.1 & 93.1                          & 49.4 & 74.0                          \\
Ours (ViT-L)                          & \bestresult{91.5}            & \bestresult{62.5}            & \bestresult{68.3}                 & \bestresult{27.7}          & \bestresult{53.1}           & \bestresult{20.4}              & \secondbest{94.3}            & 64.3                                               & \bestresult{86.2} & \bestresult{93.4}                          & \bestresult{52.6} & \bestresult{75.1}                          \\
\hline
\end{tabular}
\endgroup
\caption{\textbf{Comparison with state-of-the-art methods on Action-related tasks.} 
We report results for linear probing, full fine-tuning, and video retrieval. Methods are sorted chronologically. Methods using additional modality over RGB videos are shown in \textcolor[rgb]{0.6,0.6,0.6}{Grey} (A = audio, D = frame differences, F = optical flow). R@1 and R@5 indicate video retrieval accuracy in Top-1 and Top-5 nearest neighbors, respectively. Best results are shown in \bestresult{Red}, and second-best in \secondbest{Blue}.}
\label{table:big}
\end{table*}

\subsection{Downstream Tasks}
\label{sec:downstream}

We compare our self-supervised representations learned on Kinetics400 to prior video SSL approaches on numerous video understanding benchmarks.
Note that, while prior methods differ widely in terms of network architecture (among other factors), our results with ViT-B are directly comparable with the prior state-of-the-art approaches SVT~\cite{svt}, and VideoMAE~\cite{videomae}.

\noindent \textbf{Video Retrieval on UCF101 and HMDB51.} 
We perform action retrieval experiments to demonstrate the suitability of our features for semantic video similarity search. 
Following prior works \cite{memdpc, tclr, vi2clr}, the test set of each dataset is used as a \textit{query-set}, and the training set is considered as a \textit{search-set}. 
We report Top-1 and Top-5 retrieval accuracy in Table~\ref{table:big}. 
Our method outperforms all prior works and achieves \textbf{3.3\%} and \textbf{8.2\%} absolute improvement of Top-1 accuracy on UCF101 and HMDB51.

\noindent\textbf{Action Recognition on UCF, HMDB, and Kinetics.} 
We report  top-1 accuracies of linear probes and finetuning in Table~\ref{table:big}.
The results demonstrate that our method is highly competitive and outperforms most previous works on these standard benchmarks. 
This highlights its potential to achieve excellent results on videos found on the \textit{web}.

\noindent\textbf{Action Recognition on SSv2 and NTU60.} Since these datasets are captured in controlled and shared settings, they exhibit less scene bias than datasets such as UCF, HMDB, and Kinetics and require a stronger temporal understanding to accurately classify actions.
Our method outperforms the best previous methods in linear probing by an absolute margin of \textbf{3.2\%} and \textbf{1.5\%} on SSv2 and NTU60, respectively, demonstrating its suitability for action datasets captured in \textit{controlled, real-world settings}.

\noindent\textbf{Multi-Label Action Recognition on Charades.} 
We follow the protocol of~\cite{severe}, where the video-level multi-label prediction task is considered. 
We report linear multi-label classification performance in terms of mean average precision (mAP) in Table~\ref{table:big}.
 Our method achieves an absolute improvement of \textbf{1.3\%} over the previous state-of-the-art method, demonstrating its effectiveness in real-world \textit{multi-label and untrimmed videos}.

\noindent\textbf{Holistic Video Downstream on HVU.} We perform linear classification on various semantic categories, including scenes, objects, events, attributes, and concepts, along with actions. 
As all semantics are in multi-label format, we report the performance in terms of mean average precision (mAP), as shown in Table~\ref{table:hvu}. 
Our method consistently outperforms the prior state-of-the-art methods and achieves the best overall score for \textit{holistic video understanding}. 

\begin{table}
\centering
\small
\begingroup
\setlength{\tabcolsep}{1.2pt}
\arrayrulecolor[rgb]{0.753,0.753,0.753}
\begin{tabular}{l|c|c|c|c|c|c!{\color{black}\vrule}c} 
\arrayrulecolor{black}\hline

\hline

\hline\\[-3mm]
\textbf{Method} & \textbf{Action} & \textbf{Obj.} & \textbf{Scene} & \textbf{Event} & \textbf{Attr.} & \textbf{Concept} & \textbf{Overall}  \\ 
\hline
SVT             & 38.48           & 30.35           & 30.97          & 37.87          & 28.20              & 35.64            & \secondbest{33.58}             \\
$\rho$BYOL         & 33.20           & 25.82           & 28.40          & 35.50          & 24.16              & 33.21            & 30.05             \\
VideoMAE        & 27.49           & 23.36           & 24.56          & 29.78          & 21.04              & 28.75            & 25.83             \\
Ours\footnotesize{(ViT-B)}            & 38.65           & 33.46           & 34.24          & 40.23          & 30.99              & 38.38            & \bestresult{35.99}             \\
\hline
\end{tabular}
\endgroup
\caption{Downstream on \textbf{HVU dataset}~\cite{hvu}}
\label{table:hvu}
\end{table}

\begin{figure}[h]
    \centering
    \includegraphics[width=\columnwidth]{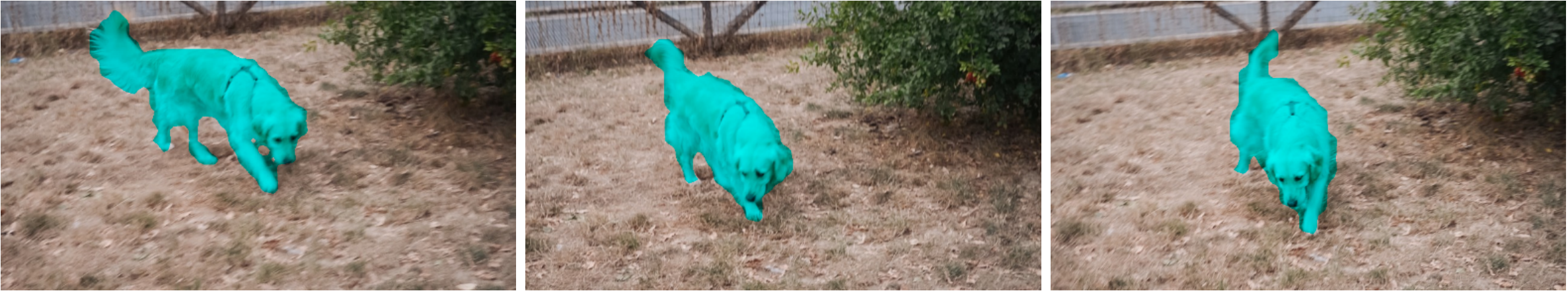}
    \includegraphics[width=\columnwidth]{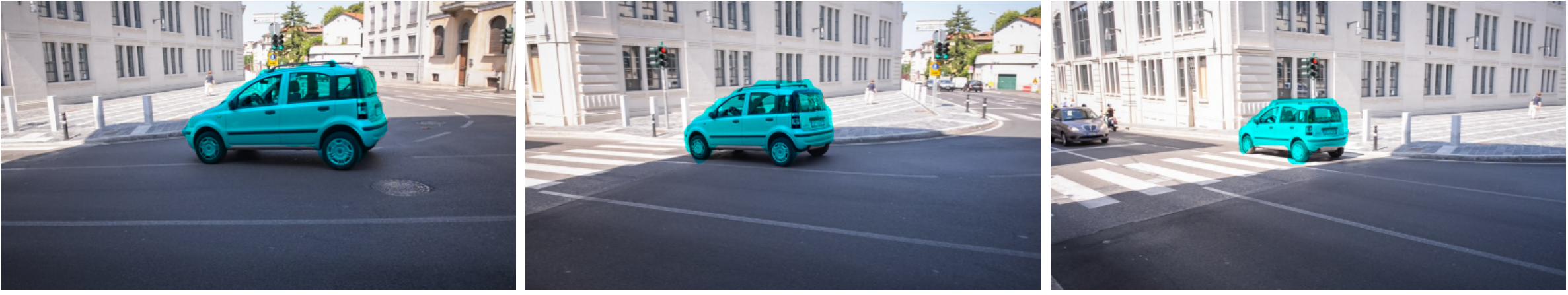}
    \caption{\textbf{Qualitative Results on Video Object Segmentation (DAVIS-17)}: Uniformly sampled frames from sequences (top to bottom) \texttt{dog} and \texttt{car-shadow}.
    }
    \label{fig:vos}
\end{figure}

\begin{table}[h]
\small
\centering
\begingroup
\setlength{\tabcolsep}{3pt}
\begin{tabular}{lccc}
\arrayrulecolor{black}\hline

\hline

\hline\\[-3mm]

\textbf{Pretraining} & \textbf{J\&F-Mean} & \textbf{J-Mean} & \textbf{F-Mean}  \\ 
\hline

ST-MAE~\venue{\cite{stmae}}           & 53.5               & 52.6            & 54.4             \\
VideoMAE~\venue{\cite{videomae}}            & 53.8               & 53.2            & 54.4             \\
MotionMAE~\venue{\cite{motionmae}}            & \secondbest{56.8}               & \secondbest{55.8}            & \secondbest{57.8}             \\
SVT~\venue{\cite{svt}}                  & 48.5                 & 46.8              & 50.1               \\
Ours (ViT-B)                & \bestresult{62.1}                 & \bestresult{60.5}              & \bestresult{63.6}               \\

\hline

\end{tabular}
\endgroup
\caption{Video Object Segmentation on \textbf{DAVIS-2017}. 
}
\label{table:vos}
\end{table}
\noindent\textbf{Video Object Segmentation (VOS) on DAVIS.} 
We follow the semi-supervised protocol of DAVIS-2017~\cite{davis}, where the object masks of the first frame of a video are given, and the task is to predict the masks in the rest of the frames.
Table~\ref{table:vos} shows a comparison with the prior works in the same protocol. 
All video SSL methods use a ViT-B architecture and are pre-trained on Kinetics400. 
Our method outperforms other video SSL methods. 
Some qualitative results are shown in Figure~\ref{fig:vos}. 
More visualizations in \supp{Supp. Sec.~\ref{sec:qualitative}} and \supp{Project Webpage}.

\begin{figure}[h]
    \centering
    \includegraphics[width=0.9\columnwidth]{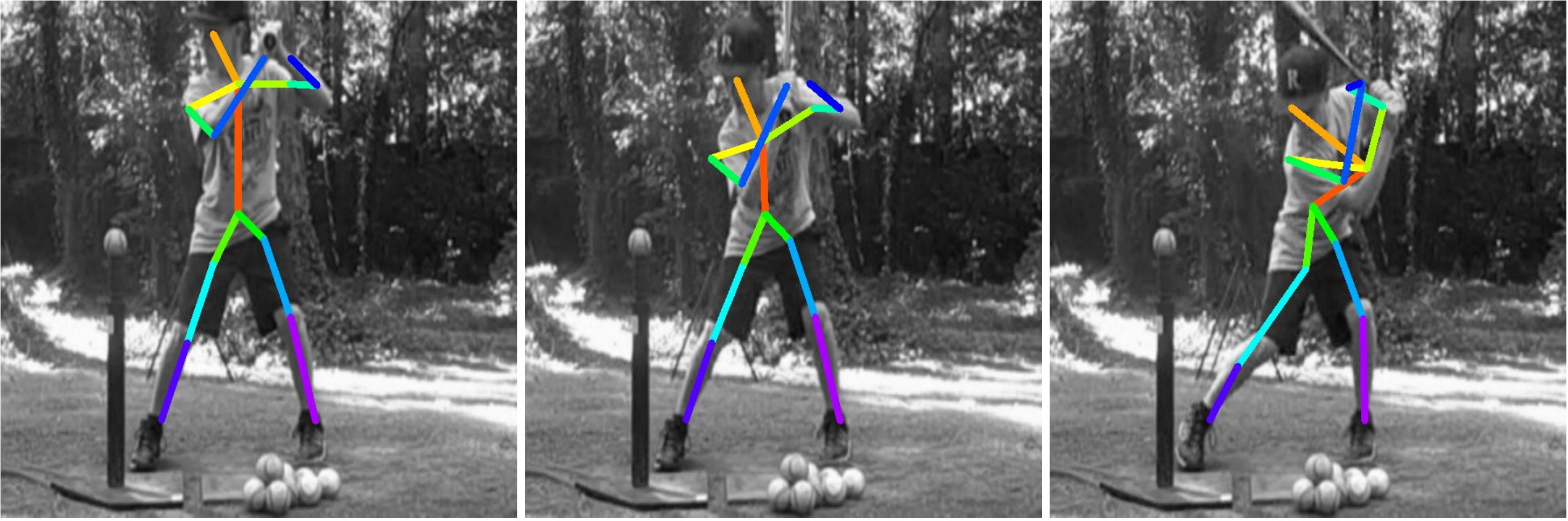}
    \caption{Qualitative results on JHMDB Pose Propagation.}
    \label{fig:pose}
\end{figure}

\noindent\textbf{Human Pose Propagation}
We use validation videos of JHMDB and follow the evaluation protocol of~\cite{uvc}. 
In this protocol, the key points of the human pose are given for the first frame, and the task is to predict the location of those key points in the subsequent frames. 
We employ our Kinetics400 pre-trained video SSL model without any further tuning. 
The performance is measured as the percentage of correct key points (PCK@X), where X is a distance threshold from the ground-truth joint position. 
The results in Table~\ref{table:pose} show the superior performance of our method compared to prior video SSL methods. 
Qualitative results are shown in Figure~\ref{fig:pose} and \supp{Supp. Sec.~\ref{sec:qualitative}}. 

\begin{table}[h]
\centering
\small
\begingroup
\setlength{\tabcolsep}{2pt}
\begin{tabular}{lccccc} 
\arrayrulecolor{black}
\hline

\hline

\hline\\[-3mm]
\textbf{Pretraining} & \multicolumn{1}{l}{\textbf{$PCK_{0.1}$}} & \multicolumn{1}{l}{\textbf{$PCK_{0.2}$}} & \multicolumn{1}{l}{\textbf{$PCK_{0.3}$}} & \multicolumn{1}{l}{\textbf{$PCK_{0.4}$}} & \multicolumn{1}{l}{\textbf{$PCK_{0.5}$}}  \\ 
\hline
SVT                  & 35.3                                 & \secondbest{62.66}                                & \secondbest{77.6}                                 & 87.26                                & \secondbest{91.94}                                 \\
VideoMAE             & \secondbest{36.5 }                                & 62.1                                 & 76.7                                 & \secondbest{88.1}                                 & 91.5                                  \\
Ours \footnotesize{(ViT-B)}                & \bestresult{43.1}                        & \bestresult{69.7}                        & \bestresult{81.6}                        & \bestresult{88.3}                        & \bestresult{92.7}                         \\
\hline

\end{tabular}
\endgroup
\caption{Pose Propagation on \textbf{JHMDB} dataset.}
\label{table:pose}
\end{table}
\noindent\textbf{Human Gait Recognition.}
We evaluate our model on the CASIA-B gait recognition dataset with the standard split. The test set includes 50 subjects each with 10 sequences. The test set is divided into two splits: \textit{gallery} and \textit{probe} set, and the goal is to retrieve the probe set. 
Ours and prior video SSL methods
are pre-trained on Kinetics400, and no further learning is performed for gait recognition. We report rank-1 accuracy for each split in Table~\ref{table:gait}. Our method outperforms prior works by a considerable margin. 
\begin{table}[H]
\centering
\small

\begin{tabular}{lccc|c}
\arrayrulecolor{black}\hline

\hline

\hline\\[-3mm]
\textbf{Method} & \textbf{NM} & \textbf{BG} & \textbf{CL} & \textbf{Mean}                       \\ 
\hline
$\rho$BYOL            & \secondbest{90.65}          & \secondbest{80.51}          & \secondbest{28.59}          & \secondbest{66.58}                                  \\
VideoMAE       & 65.30          & 57.21          & 21.40          & 47.97                                  \\
Ours            & \bestresult{98.60}       & \bestresult{92.57}       & \bestresult{28.66}       & \multicolumn{1}{l}{\bestresult{73.28}}  \\
\arrayrulecolor{black}\hline

\end{tabular}
\caption{Gait Recognition on \textbf{CASIA-B} dataset.}
\label{table:gait}
\end{table}

\noindent\textbf{Robustness to input perturbations} Following~\cite{schiappa2023large}, we adopt the robustness protocol for the video retrieval task on HMDB51, where, query-set and search-set videos are corrupted using random frame-independent perturbations like translation, Gaussian noise, or random JPEG compression. 
We report Top-1 retrieval accuracy for clean and perturbed videos for recent methods in Table~\ref{table:perturbation}, where our method achieves the smallest drop in performance across various perturbations. 
This superior robustness can be attributed to our model's capacity to learn and maintain the temporal correspondence between frames, even when they are independently perturbed. These qualities make our method highly suitable for \textit{robust video retrieval} scenarios, where noise and perturbations are common challenges.
\begin{table}[h]
\centering
\small
\begingroup
\setlength{\tabcolsep}{3pt}
\arrayrulecolor[rgb]{0.502,0.502,0.502}
\begin{tabular}{lllll} 
\arrayrulecolor{black}
\hline

\hline
\textbf{Method} & \textbf{Clean} & \textbf{Translation} & \textbf{Gaussian} & \textbf{JPEG}  \\ 
\hline
SVT             & 44.40          & 43.21\decrease{2.7}          & 41.80\decrease{5.9}       & 41.52\decrease{6.5}   \\
$\rho$BYOL         & 39.60          & 35.84\decrease{9.6}          & 33.31\decrease{15.9}     & 36.23\decrease{8.5}   \\
VideoMAE        & 32.50          & 26.72\decrease{17.8}         & 26.22\decrease{19.3}     & 26.61\decrease{18.1}  \\
Ours (ViT-B)    & 49.40          & \textbf{48.43}\decrease{\textbf{2}}           & \textbf{47.10}\decrease{\textbf{4.7}}       & 47.21\decrease{4.4}   \\
Ours (ViT-L)           & 52.60          & 51.50\decrease{{2.1}}          & 50.06\decrease{{4.8}}      & \textbf{50.46}\decrease{\textbf{4.1}}   \\
\hline
\end{tabular}
\endgroup
\caption{Action Retrieval with perturbation. \decrease{n} shows relative drop in \% compare to clean video retrieval.}
\label{table:perturbation}
\end{table}

\section{Conclusion}
We have introduced a self-supervised approach for video representation learning.
Our model extends a representation of static video frames with a transformer, which we train through self-supervision to capture temporal features.
Importantly, we identified and addressed shortcuts in learning through temporal self-supervision and reformulated time-related learning tasks as more challenging frame-wise prediction tasks.
We demonstrated the effectiveness of our approach on a wide variety of video understanding tasks for both generalization and robustness of the learned representations.
We believe that our advancements in temporal self-supervision could inspire future work in other temporal data modalities (\ie, time-series data) or multi-modal video understanding, \eg, in combination with audio or language.

{
    \small
    \bibliographystyle{ieeenat_fullname}
    \bibliography{main}
}

\clearpage
\appendix
\section{Overview}
\begin{itemize}
    \item Section~\ref{sec:dataset}: Details on datasets used in the paper.    
    \item Section~\ref{sec:implementation}: Implementation details about network architectures, training setup, and data augmentations. 
    \item Section~\ref{sec:add_ablations}: Ablation study for our framework.
    \item Section~\ref{sec:add_method}: Supportive diagrams and explanation for our method.    
    \item Section~\ref{sec:qualitative}: Qualitative results

    \item Section~\ref{sec:conceptual_compare}: Conceptual comparison with prior works based on temporal pretext tasks.

\end{itemize}

\section{Datasets}
\label{sec:dataset}
\noindent\textbf{UCF101}~\cite{ucf101} contains a total number of 13,320 videos of 101 human actions, which are collected from internet videos. We use split-1 for our experiments, where there are 9,537 training videos and 3,783 test videos. 

\noindent\textbf{HMDB51}~\cite{hmdb} is collected from movie videos, which contain a total number of 6,849 videos over 51 action classes. We report results on split-1, which has 3,570 training videos and 1,530 test videos. 

\noindent\textbf{Kinetics400}~\cite{kinetics} is a large-scale action recognition dataset collected from YouTube videos. It contains 400 different human actions and a total number of 240k training and 20k validation videos. 

\noindent\textbf{Charades}~\cite{charades} is a multi-label action dataset of daily-life actions in indoor setups. It contains 157 total classes and has a standard split of 9.8k training and 1.8k validation videos. Typically, Charades videos are 30 seconds, and their action may occur over a longer time compared to typical 10 seconds videos of Kinetics400.  The multi-label actions could be co-occurring (\eg \texttt{walking} and \texttt{talking} at the same time) or temporally dependent on each other (\eg \texttt{standing-up} followed by \texttt{walking}).  

\noindent\textbf{Something-SomethingV2}~\cite{goyal2017something} contains a total of 220,847 videos, out of which 168,913 videos are used for training, 24,777 for validation, and 27,157 for testing.
The dataset includes diverse fine-grained and coarse-grained actions, challenging scenarios such as occlusion and motion blur, and has become a popular benchmark for evaluating the performance for action recognition

\noindent\textbf{NTU60}~\cite{ntu60} dataset is a large-scale dataset for human action recognition, consisting of over 56,000 action samples performed by 40 different subjects. It contains 60 different action classes. The training set contains 40,320 samples and the testing set contains 16,560 samples, with a split of 20 samples per subject for testing and the rest for training.

\noindent\textbf{Holistic Video Understanding}~\cite{hvu} encompasses the full spectrum, with 248 categories for scenes, 1678 for objects, 739 for actions, 69 for events, 117 for attributes and 291 for concepts, covering a wide range of visual concepts that occur in real-world problems. This approach naturally captures the long tail distribution of visual concepts in such problems. We utilize the miniHVU split, which includes 129,628 training and 10,527 validation videos. Each video semantic is multilabel in nature.

\noindent\textbf{DAVIS2017}~\cite{davis} contains pixel-level annotations for 59 objects over 30 validation videos. DAVIS2017 videos are challenging due to the simultaneous presence of multiple objects at a time, which may undergo deformations, scale changes, and occlusions.  

\noindent\textbf{JHMDB Pose}~\cite{jhmdb21} is a subset of the HMDB51 dataset with 21 action categories and only contains a single actor per video. It provides annotations of 13 joints for each frame, resulting in 31,838 annotated frames in total. 

\noindent\textbf{CASIA-B}~\cite{casia} contains walking sequences in different indoor conditions. The test set includes 50 subjects, each with 10 walking videos under 3 walking conditions: normal walking (NM), walking with a bag (BG), and walking with different clothes (CL). \textit{Gallery-set} of the test-split contains walking sequences NM\#01-04, BG\#01-02, CL\#01-02 and \textit{Probe-set} contains NM\#05-\#06, BG\#01-02, and CL\#01-02.

\section{Implementation Details}
\label{sec:implementation}
\subsection{Architectural Details}
\subsubsection{Video Transformer Network}
As the image encoder $F(\cdot)$ in our Video Transformer Network model, we utilize a ViT implementation from the \texttt{timm} package\footnote{\url{https://github.com/rwightman/pytorch-image-models}}.
For the temporal encoder $E(\cdot)$, we utilize the official PyTorch implementation \texttt{torch.nn.TransformerEncoder}. 

We utilize the CLS output token from $F(\cdot)$ as frame-embedding $f$, which has a dimension of 1024. In order to reduce computation, we first reduce the dimension of $f$ from 1024 to 512 using a fully-connected layer followed by GeLU activation. In our default setting, we feed 512-dimensional input tokens to $E(\cdot)$, which consists of 6 transformer layers with 8 attention heads each. 

For the positional embedding, we utilize the relative frame positions (0 to 7) as input to the learnable positional embedding module (implemented via PyTorch's \texttt{nn.Embedding} module).

\subsubsection{Non-Linear Projection Heads}
For the contrastive losses $\mathcal{L}_C$, we first project the embedding into a lower dimensional representation space using non-linear projection heads. We utilize representation spaces of 128 dimensions following the most commonly used contrastive learning setup~\cite{simclr}. For non-linear projection, we utilize a multi-layer perceptron (MLP) of 2-layers which has the following design, 
where EmbeddingSize=1024.
\begin{verbatim}
LinearLayer of EmbeddingSize x 512
BatchNorm + ReLU
LinearLayer of 512 x 128
L2 normalization
\end{verbatim}

\subsubsection{Temporal Pretext Tasks Prediction Heads}
\noindent \textbf{Out-of-order Frame Localization (OFL)} We use a linear layer followed by a sigmoid over the output tokens $e^{j}$ and perform binary classification over each timestamp $j$. 

\noindent \textbf{Time-varying Skiprate Prediction (TSP)} We first compute the difference of consecutive tokens, which is followed by a linear layer of output M, where M is the number of available skiprates. In default setting M=3 classes.

\subsection{Training Setup}
In order to update our network parameters, we utilize the Adam optimizer~\cite{adam} with $\beta_1=0.9$, $\beta_2=0.999$, a base learning rate of 1e-4, and batch-size $\mathcal{B}=128$. We perform self-supervised pre-training for 80 epochs on Kinetics400. A linear warmup is utilized for the first 5 epochs, and a patience-based learning rate scheduler is implemented, which drops its value to half on the plateau. 

\subsection{Input and Data Augmentations}
We utilize a sequence of 8 frames of $224\times224$ resolution. We utilize separate clips for the OFL and TSP tasks. OFL clips have a fixed skiprate of 4 for in-order frames, and TSP clips have a time-varying skiprate from $\{1,4,8\}$.

As data augmentation, we first apply standard augmentations like color-jittering, horizontal flipping, and random scaling consistently on all frames. Then, we apply frame-wise spatial cropping.

\subsection{Training Losses}

\noindent \textbf{TSP loss} ($\mathcal{L}_{TSP}$) is implemented using standard Cross Entropy loss. 

\noindent \textbf{OFL loss} ($\mathcal{L}_{OFL}$) is implemented using standard Binary Cross Entropy loss. 

\noindent \textbf{Contrastive Loss} ($\mathcal{L}_C$)
is computed from two differently augmented versions of OFL clips.
We utilize a default temperature of 0.1.

\noindent  In default setting, $\lambda_O$ = $\lambda_T$ = $\lambda_C$ = 1.

\section{Additional Ablations}
\label{sec:add_ablations}
We perform additional ablation studies with ViT-L as the backbone and self-supervised pre-training on UCF101 for 250 epochs.

\subsection{Number of Layers in Temporal Encoder}
In Table~\ref{table:encoder_layers}, we observe that increasing the number of layers in temporal encoder $E(\cdot)$ from 1 to 3 leads to an increase in performance noticeably. However, further increasing the layers from 3 to 6 leads to smaller improvement. These improvements are more visible on the HMDB51 dataset compared to UCF101, which suggests that HMDB51 gets more benefit from learning stronger temporal dynamics compared to UCF101. In our default setting, we use 6 layers.

\begin{table}[h]
\centering
\begingroup
\setlength{\tabcolsep}{4pt}
\begin{tabular}{ccccc} 
\arrayrulecolor{black}\hline

\hline

\hline\\[-3mm]
\multicolumn{1}{l}{\multirow{2}{*}{\textbf{\#layer in $E(\cdot)$}}} & \multicolumn{2}{c}{\textbf{UCF101}} & \multicolumn{2}{c}{\textbf{HMDB51}}  \\
\multicolumn{1}{l}{}                                         & \textbf{R@1} & \textbf{Lin}         & \textbf{R@1} & \textbf{Lin}          \\ 
\hline
6                                                            & 84.68        & 89.90                & 50.20        & 58.70                 \\
3                                                            & 84.28                    & 89.50            & 48.90   & 58.10                  \\
1                                                            & 83.55            & 89.10                    & 47.10            & 57.22                     \\
\hline
\end{tabular}

\endgroup
\vspace{-1mm}
\caption{Number of layers in temporal encoder $E(\cdot)$}
\label{table:encoder_layers}
\end{table}

\subsection{Attention heads in Temporal Encoder}
In Table~\ref{table:atthead}, we observe that decreasing the number of attention heads leads to a small improvement in results. However, it increases the computation cost. In our default setting, we use 8 attention heads.
\begin{table}[h]
\centering
\begingroup
\setlength{\tabcolsep}{4pt}
\begin{tabular}{lcccc} 
\arrayrulecolor{black}\hline

\hline

\hline\\[-3mm]
\multirow{2}{*}{\textbf{\# Attention Heads}} & \multicolumn{2}{c}{\textbf{UCF101}}         & \multicolumn{2}{c}{\textbf{HMDB51}}  \\
                                             & \textbf{R@1}         & \textbf{Lin}         & \textbf{R@1} & \textbf{Lin}          \\ 
\hline
\multicolumn{1}{c}{8}                        & 84.68                & 89.90                & 50.20        & 58.70                 \\
\multicolumn{1}{c}{4}                        & 84.90 & 89.90 & 50.50 & 58.81                \\
\hline
\end{tabular}

\endgroup
\vspace{-1mm}
\caption{Number of attention heads in temporal encoder $E(\cdot)$}
\vspace{-3mm}
\label{table:atthead}
\end{table}

\begin{figure*}
    \centering
    \includegraphics[width=0.9\textwidth]{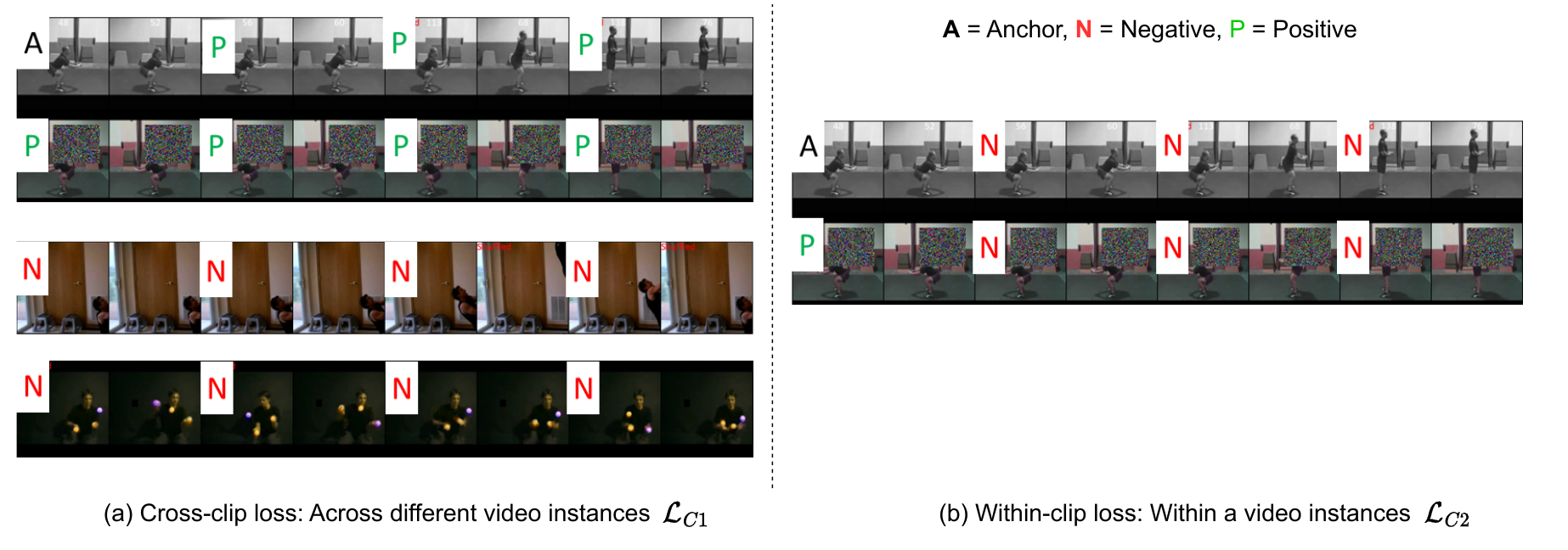}
    \caption{
    \textbf{Illustration of Frame Contrastive Losses.}
    We illustrate how positive and negative pairs for contrastive learning are built in the case of cross-clip contrastive learning (a); and within-clip contrastive learning (b). 
    Anchor frames are marked with A, and corresponding positive frames are marked with P, while negatives are indicated with N.    }
    \label{fig:imagelevel}
\end{figure*}
\begin{figure*}
    \centering
    \includegraphics[width=0.6\textwidth]{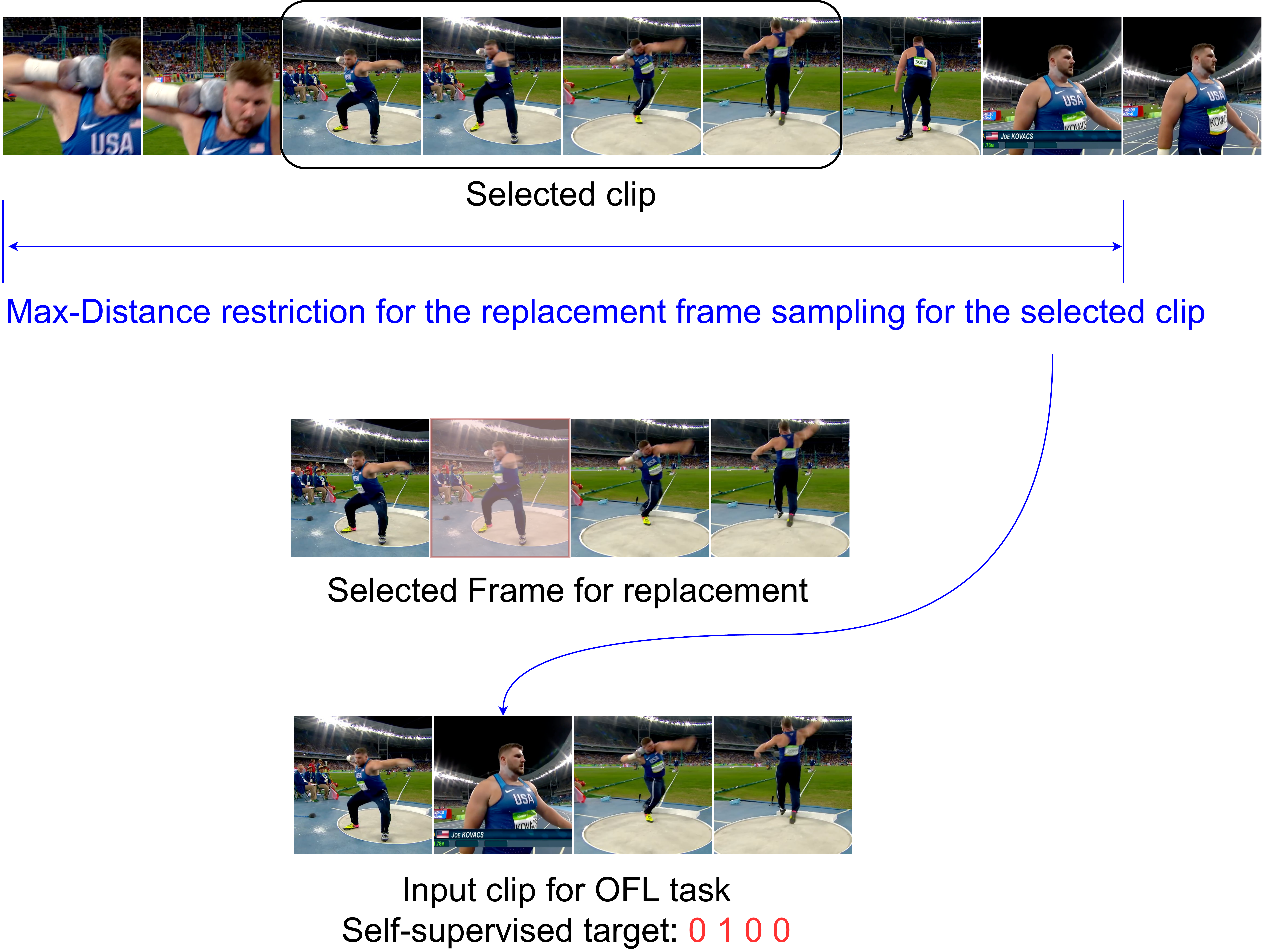}
    \caption{
    \textbf{Illustration for building clip and self-supervised target for Out-of-order Frame Localization}}
    \label{fig:ofl_clipbuilding}
\end{figure*}

\vspace{15mm}
\section{Method}
\label{sec:add_method}
\subsection{Out-of-order Frame Localization (OFL)}
An illustration for building input sequence and self-supervised target for OFL clip is shown in Fig.~\ref{fig:ofl_clipbuilding}.
\subsection{Contrastive Loss}

We illustrate the computation of frame-contrastive losses in Fig.~\ref{fig:imagelevel}.

\subsection{Pseudo-Code}
We provide pseudo-code for our method in Algorithm \ref{algo1:psuedo_code}.

\begin{algorithm*}[h]
\caption{Input Preprocessing for OFL and TSP and Proposed Augmentation strategy}
\label{algo1:psuedo_code}

\begin{algorithmic}[1]
\State \var{num\_frames} $\gets$ 8 
\Comment{number of frames to our VTN model}
\State \var{input\_video} $\gets$ \func{array}(all frames of video, length = \var{frame\_count})
\Comment{Read all frames of the video}

\sectionline %
\Comm{OFL input preprocessing}
\State \var{fix\_skip} $\gets$ 4 
\Comment{use a fix skip rate for in-order frames of OFL clip}
\State \var{md} $\gets$ 64 
\Comment{maximum distance to sample the shuffled frame indices}
\State \var{start\_ofl} $\gets$ \func{random}(0, \var{frame\_count} - \var{fix\_skip} $\times$ \var{num\_frames})
\State \var{replacement\_fraction} $\sim \mathcal{U}([0.0, 0.50])$ 
\Comment{sample the fraction of token replacement}
\State \var{replacement\_tokens} $\gets$ int(\var{replacement\_fraction} $\times$ \var{num\_frames})

\State \var{clip\_ofl\_indices} $\gets [\,]$
\For{\var{i} in \var{num\_frames}}
    \State \var{clip\_ofl\_indices}.\func{append}(\var{start\_ofl} + \var{i} $\times$ \var{fix\_skip})
\EndFor
\State \var{sampled\_shuffled\_frames} $\gets$ \func{random\_array}(\var{num\_frames}, [min(\var{clip\_ofl\_indices})-md, max(\var{clip\_ofl\_indices})+md])
\State \var{shuffled\_indices} $\gets$ \func{random\_array}(\var{replacement\_tokens}, [0, \var{num\_frames} - 1])
\State \var{clip\_ofl\_indices[shuffled\_indices]} $\gets$ \var{sampled\_shuffled\_frames[shuffled\_indices]} 
\Comment{Replace shuffled indices}
\State \var{clip\_ofl} $\gets$ \var{input\_video[clip\_ofl\_indices]} 
\Comment{reading the indices from the input\_video}
\State \var{clip\_ofl} $\gets$ \func{proposed\_augmentation\_strategy}(\var{clip\_ofl})

\sectionline %
\Comm{TSP input preprocessing}
\State \var{skip\_set} $\gets \mathcal{U}(\{1, 4, 8\})$ 
\Comment{Universal set of the skip rates}
\State \var{start\_tsp} $\gets$ \func{random}(0, \var{frame\_count} - max(\var{skip\_set}) $\times$ \var{num\_frames})
\State \var{clip\_tsp\_indices} $\gets [\,]$
\State \var{temp} $\gets$ \var{start\_tsp}
\For{\var{i} in \var{num\_frames}}
    \State \var{skip\_i} $\sim$ \var{skip\_set} 
    \Comment{Randomly select skip rate from skip\_set}
    \State \var{temp} $\gets$ \var{temp} + \var{skip\_i}
    \State \var{clip\_tsp\_indices}.\func{append}(\var{temp})
\EndFor
\State \var{clip\_tsp} $\gets$ \var{input\_video[clip\_tsp\_indices]} 
\Comment{clip building for TSP from the obtained indices}
\State \var{clip\_tsp} $\gets$ \func{proposed\_augmentation\_strategy}(\var{clip\_tsp})

\sectionline %
\Function{proposed\_augmentation\_strategy}{$input\_clip$}
    \Comm{First we apply \textbf{consistent} augmentation across the temporal dimensions}
    \State \func{resize\_crop}(Avg. 70\% of frame area)
    \State \func{color\_jittering}()
    \State \func{horizontal\_flip}()
    \Comm{Once we have consistently augmented clip, we do \textbf{frame-wise} spatial jittering}
    \State \func{resize\_crop}(Avg. 80\% of the consistently augmented frame area) 
    \State \Return \var{output\_clip}
\EndFunction
\end{algorithmic}
\end{algorithm*}

\section{Qualitative Results}
\label{sec:qualitative}

\subsection{Video Object Segmentation on DAVIS2017}
\begin{figure*}[h]
    \begin{subfigure}{\textwidth}
        \begin{center}
            \includegraphics[width=0.19\textwidth]{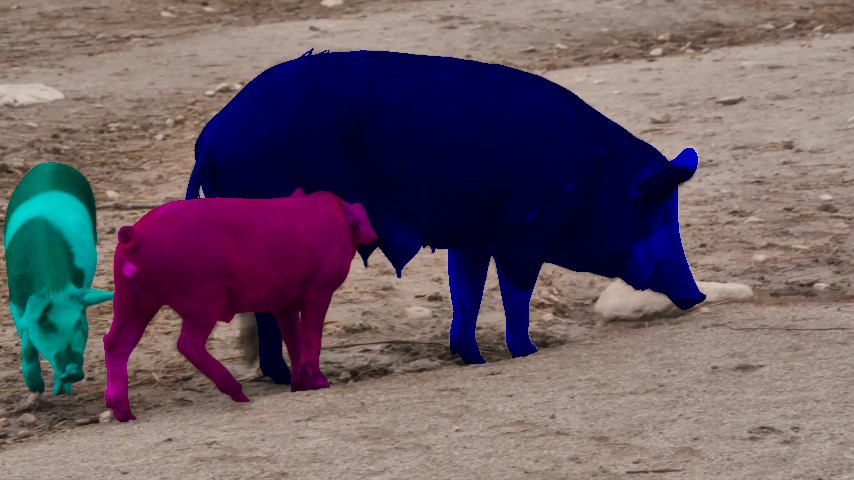}
            \includegraphics[width=0.19\textwidth]{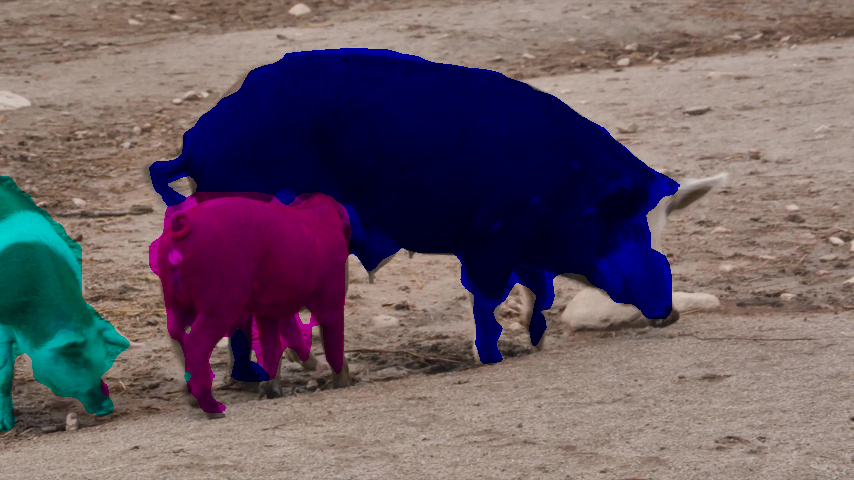}
            \includegraphics[width=0.19\textwidth]{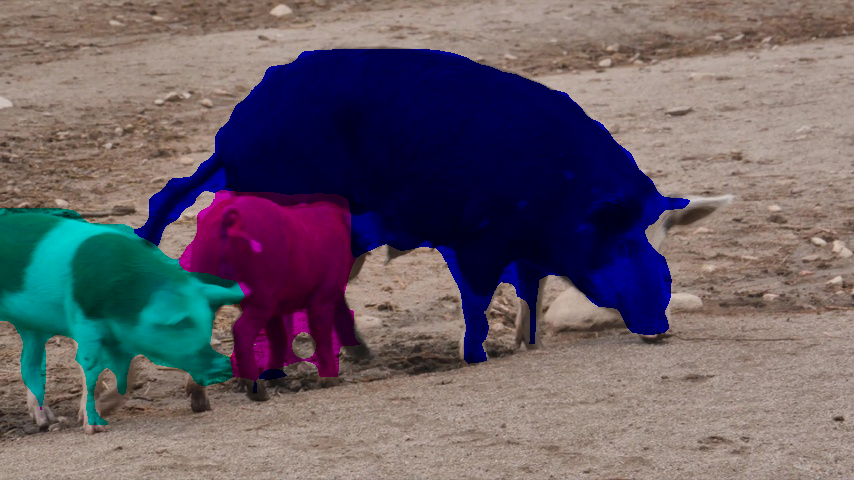}
            \includegraphics[width=0.19\textwidth]{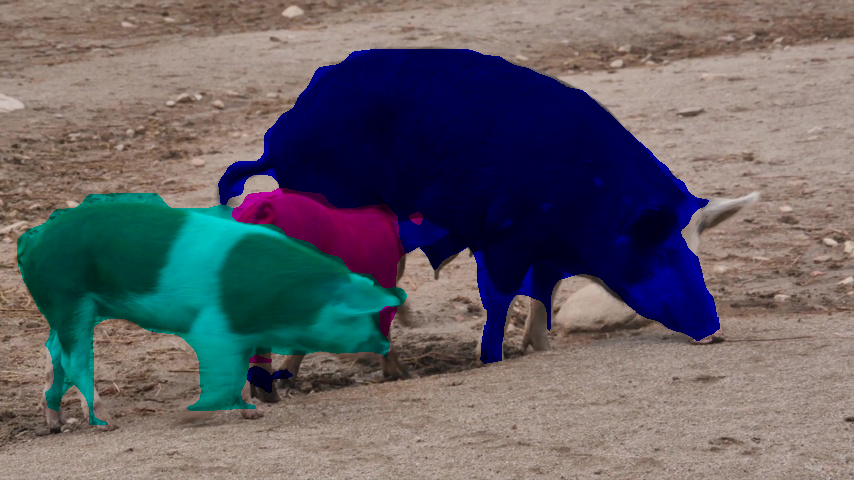}
            \includegraphics[width=0.19\textwidth]{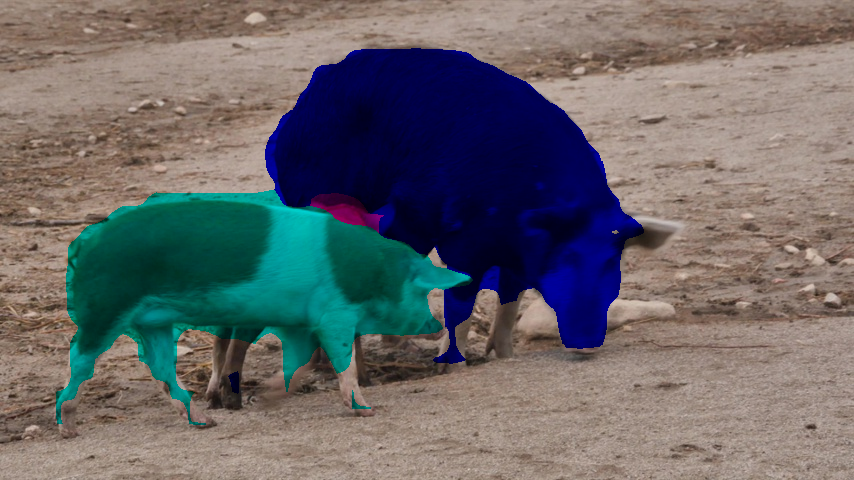}
        \end{center}
        \vspace{-4mm}
        \caption{Video name: \texttt{pig}}
        \vspace{2mm}
    \end{subfigure}
    
    \begin{subfigure}{\textwidth}
        \begin{center}
            \includegraphics[width=0.19\textwidth]{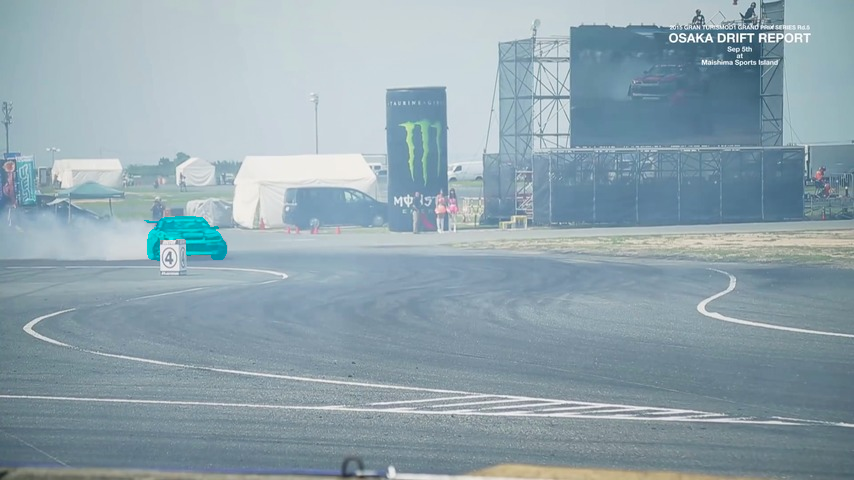}
            \includegraphics[width=0.19\textwidth]{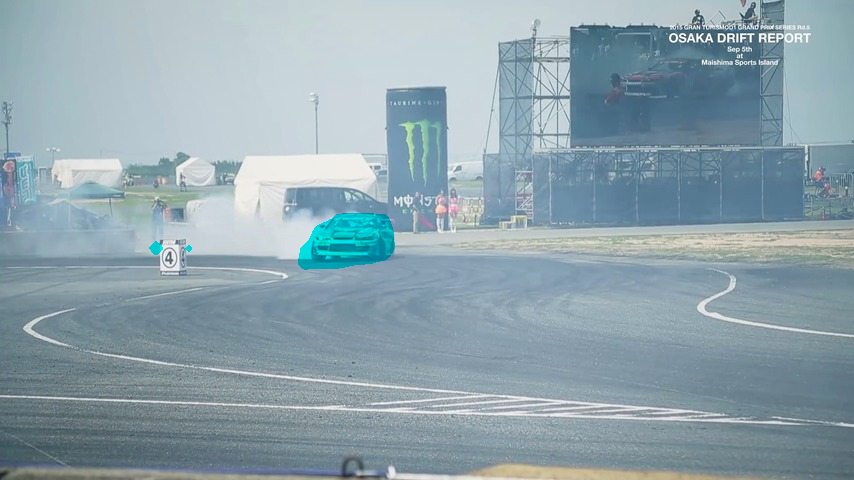}
            \includegraphics[width=0.19\textwidth]{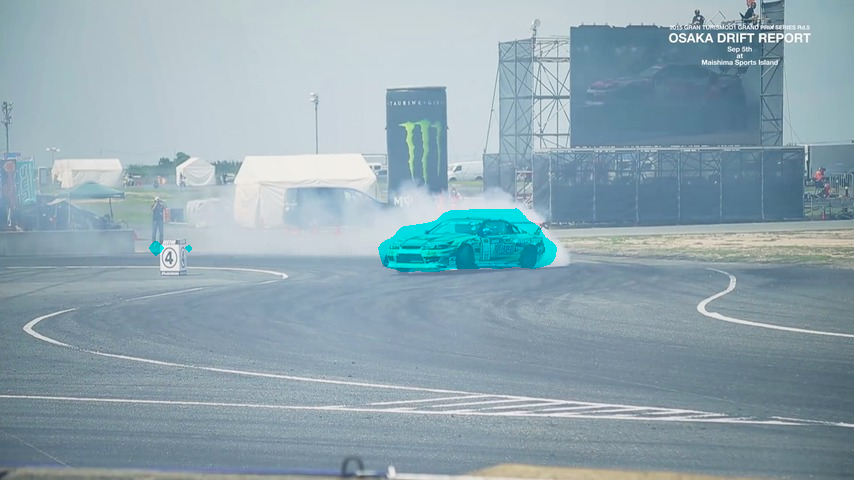}
            \includegraphics[width=0.19\textwidth]{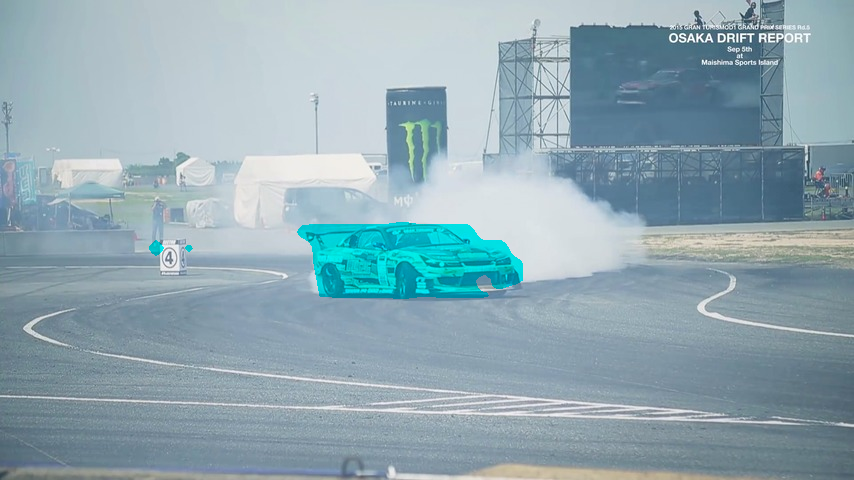}
            \includegraphics[width=0.19\textwidth]{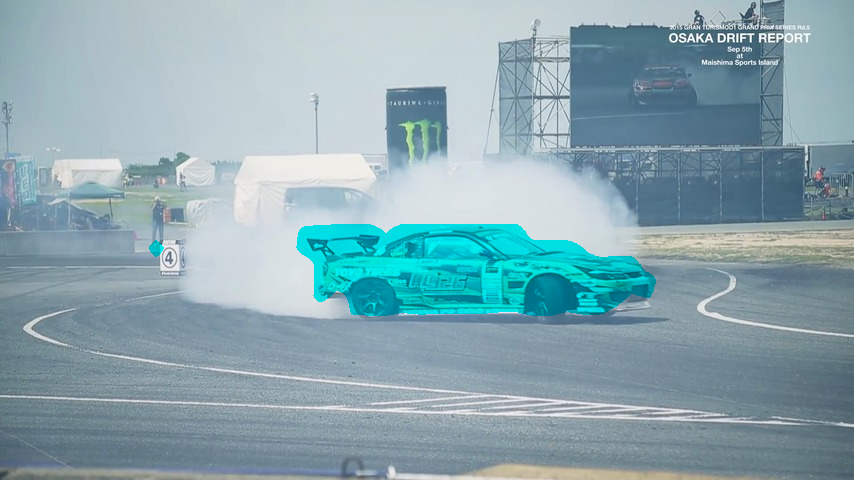}
        \end{center}
        \vspace{-4mm}
        \caption{Video name: \texttt{drift-chicane}}
        \vspace{2mm}
    \end{subfigure}
    \begin{subfigure}{\textwidth}
        \begin{center}
            \includegraphics[width=0.19\textwidth]{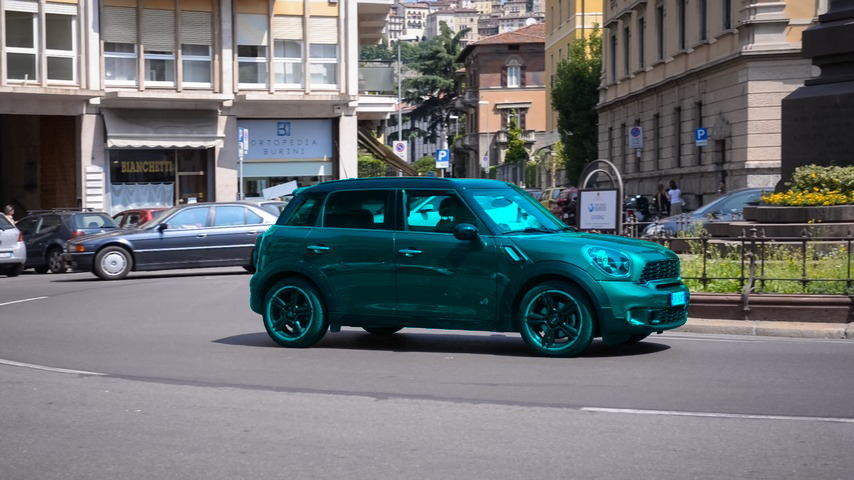}
            \includegraphics[width=0.19\textwidth]{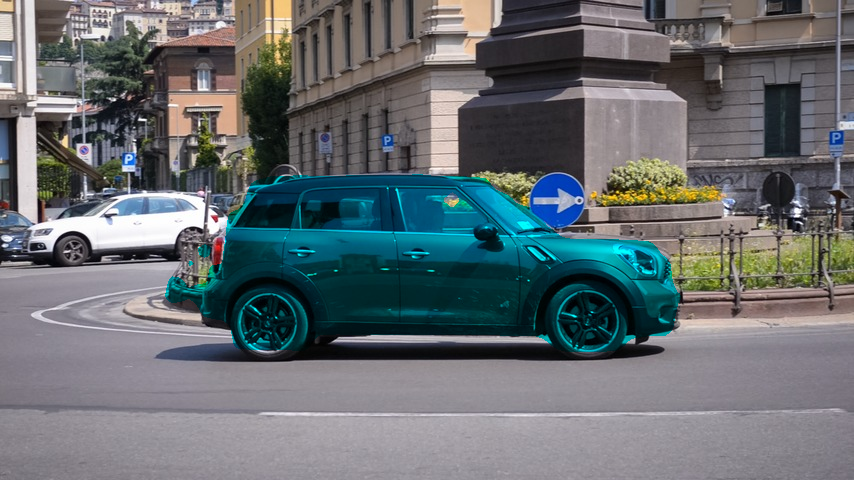}
            \includegraphics[width=0.19\textwidth]{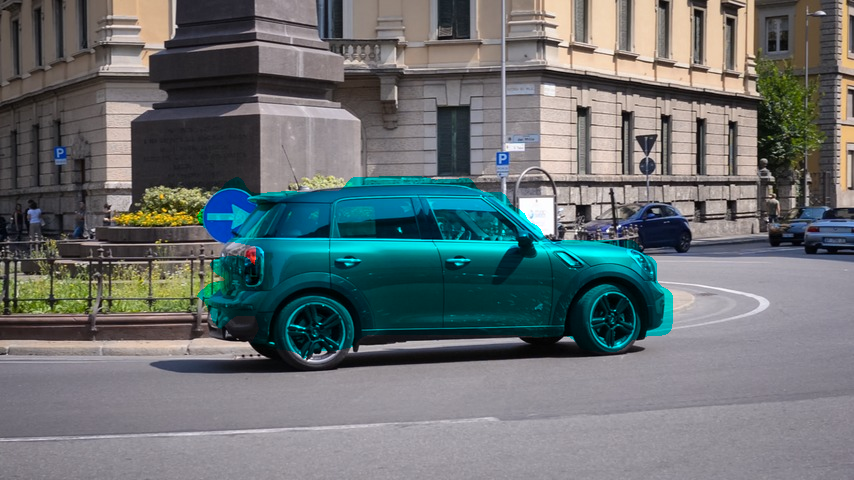}
            \includegraphics[width=0.19\textwidth]{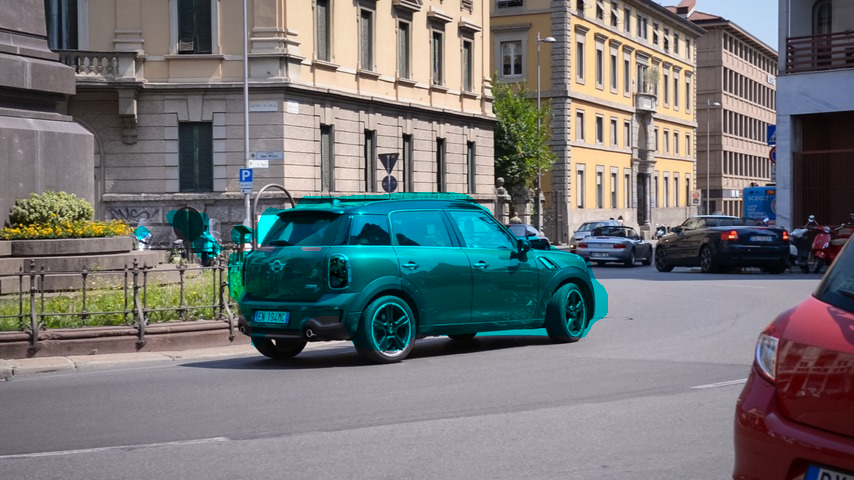}
            \includegraphics[width=0.19\textwidth]{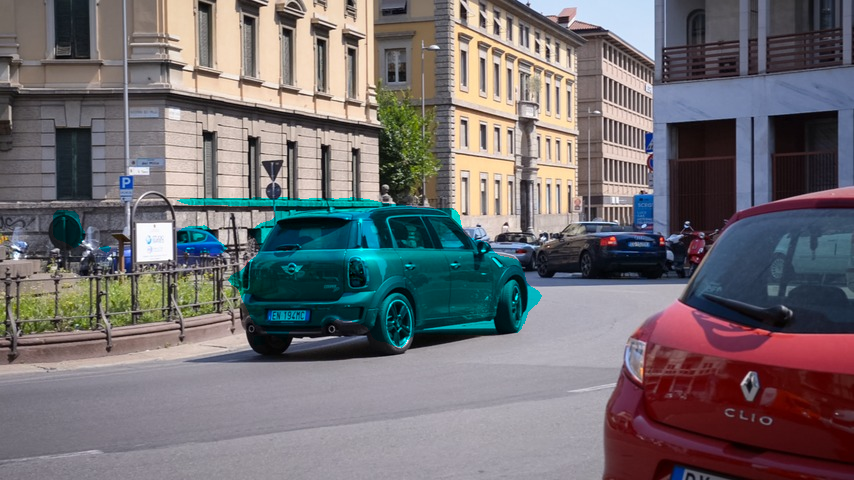}
        \end{center}
        \vspace{-4mm}
        \caption{Video name: \texttt{car-roundabout}}
        \vspace{2mm}
    \end{subfigure}
    \begin{subfigure}{\textwidth}
        \begin{center}
            \includegraphics[width=0.19\textwidth]{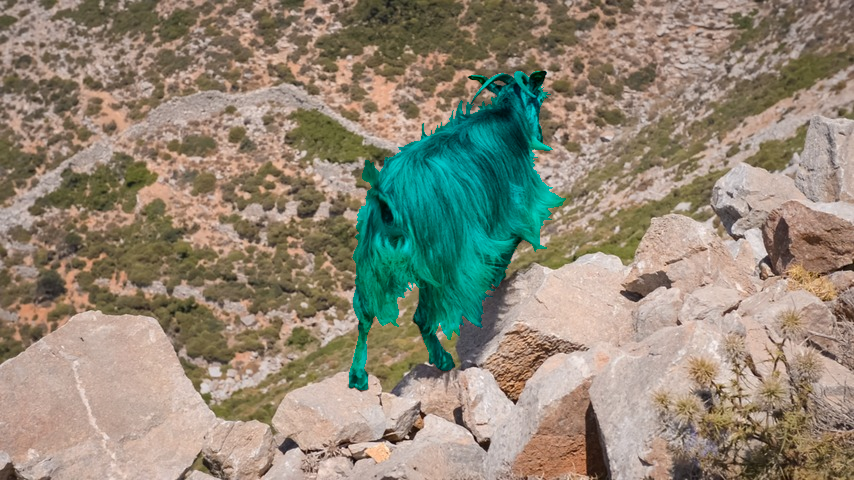}
            \includegraphics[width=0.19\textwidth]{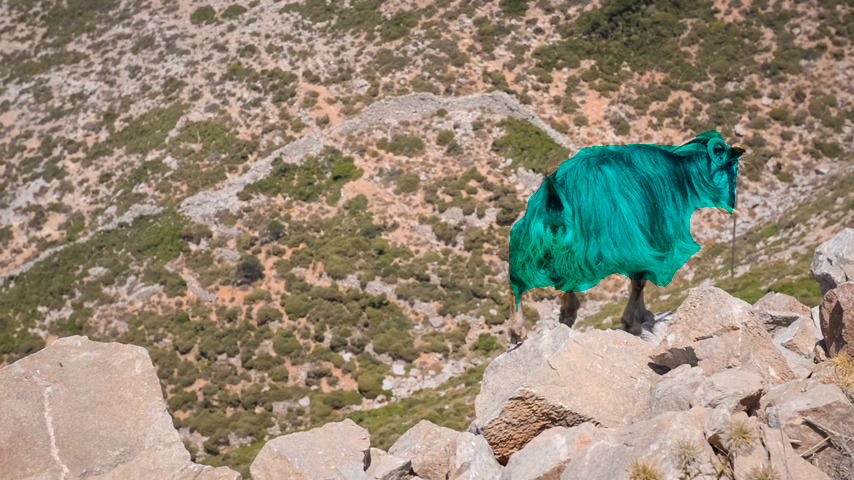}
            \includegraphics[width=0.19\textwidth]{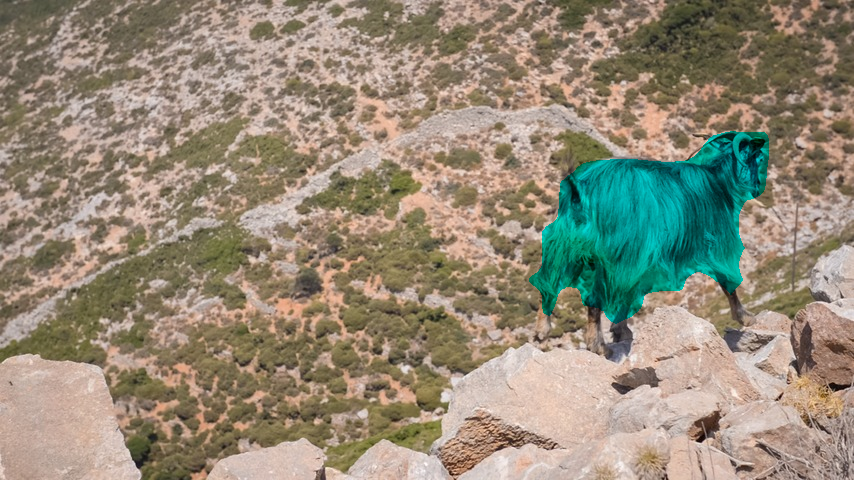}
            \includegraphics[width=0.19\textwidth]{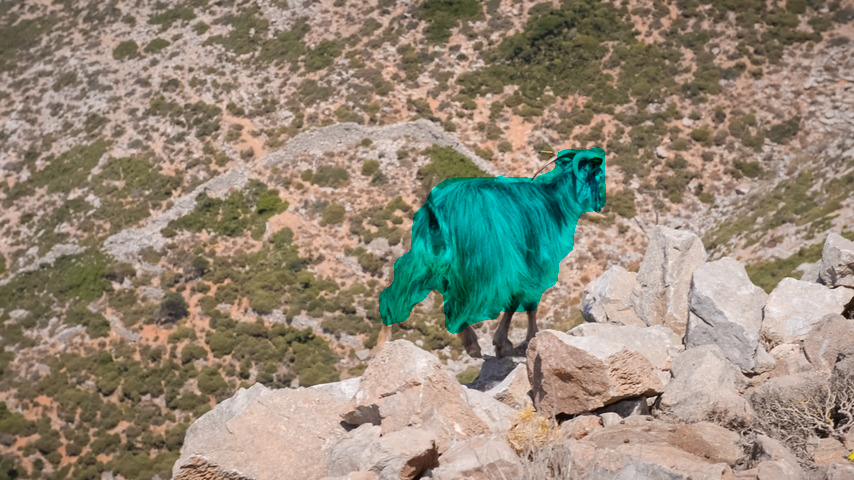}
            \includegraphics[width=0.19\textwidth]{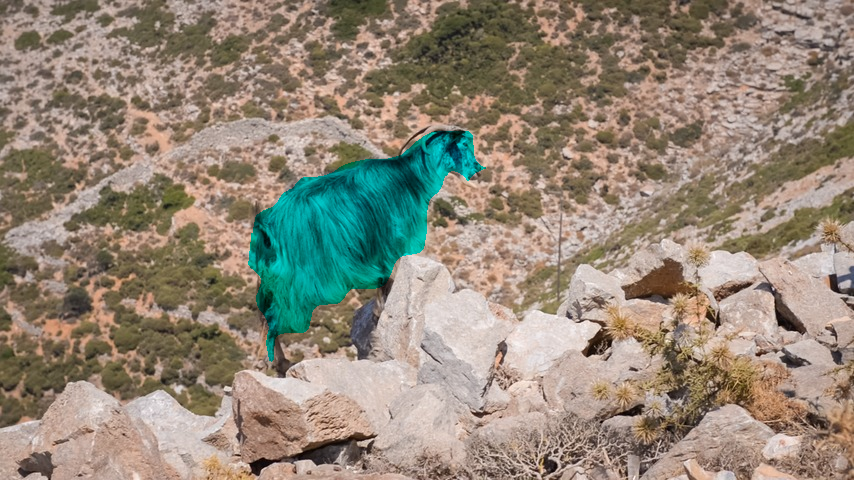}
        \end{center}
        \vspace{-4mm}
        \caption{Video name: \texttt{goat}}
        \vspace{2mm}
    \end{subfigure}
    \begin{subfigure}{\textwidth}
        \begin{center}
            \includegraphics[width=0.19\textwidth]{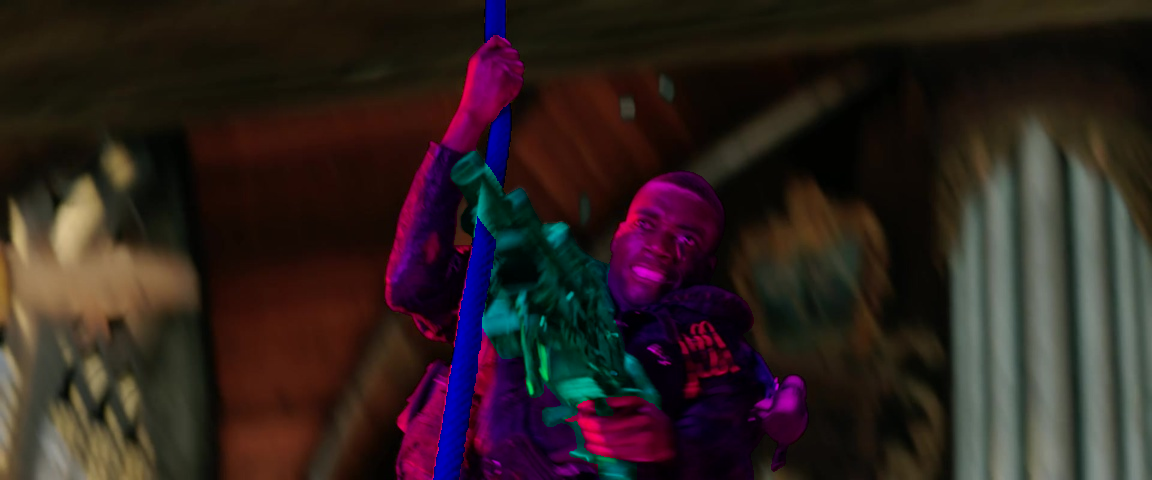}
            \includegraphics[width=0.19\textwidth]{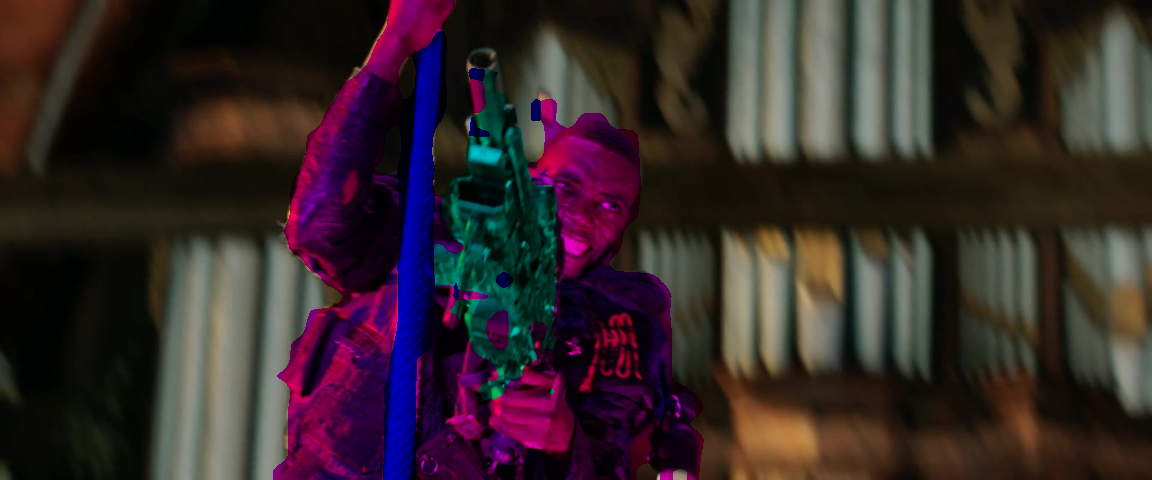}
            \includegraphics[width=0.19\textwidth]{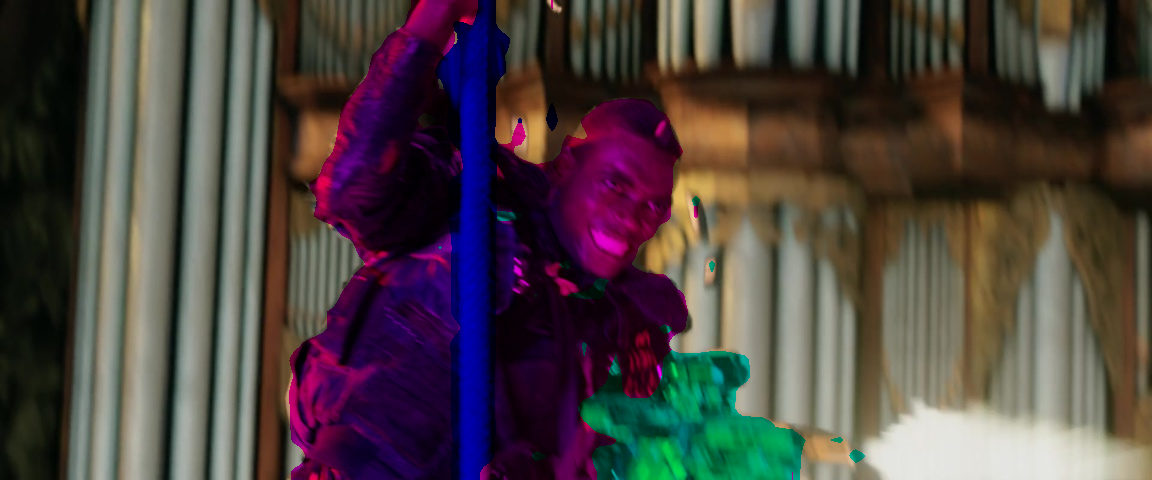}
            \includegraphics[width=0.19\textwidth]{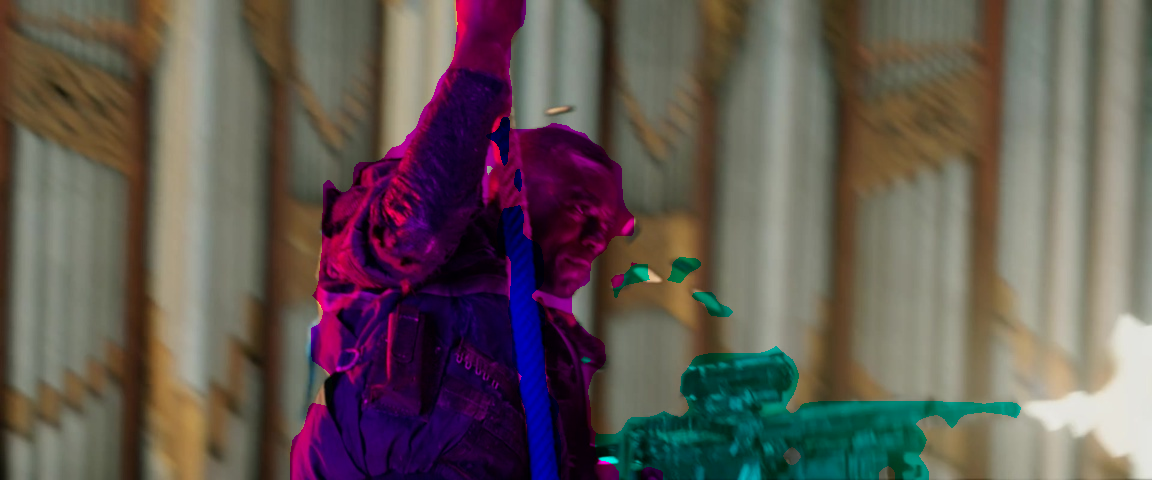}
            \includegraphics[width=0.19\textwidth]{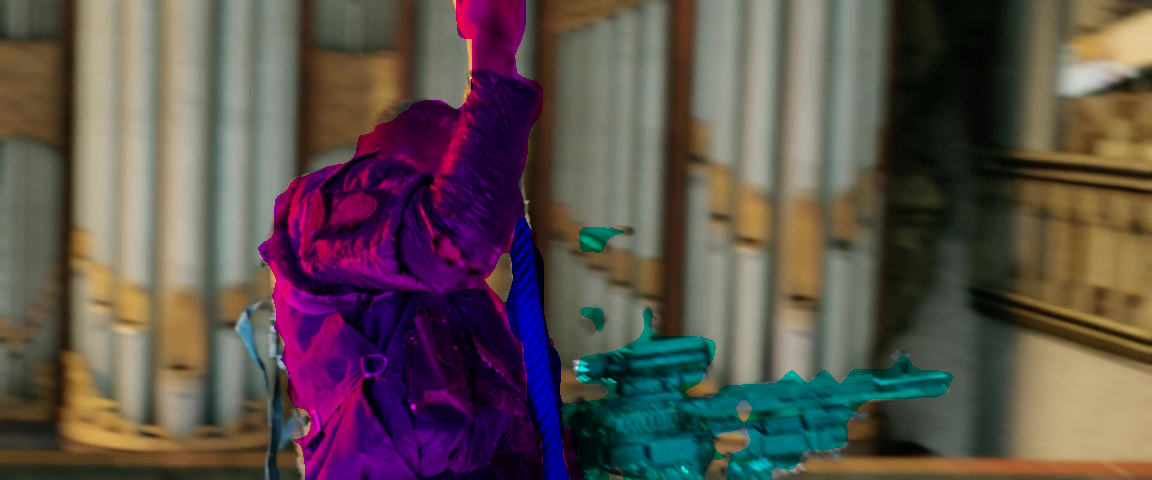}
        \end{center}
        \vspace{-4mm}
        \caption{Video name: \texttt{shooting}}
        \vspace{2mm}
    \end{subfigure}
    \begin{subfigure}{\textwidth}
        \begin{center}
            \includegraphics[width=0.19\textwidth]{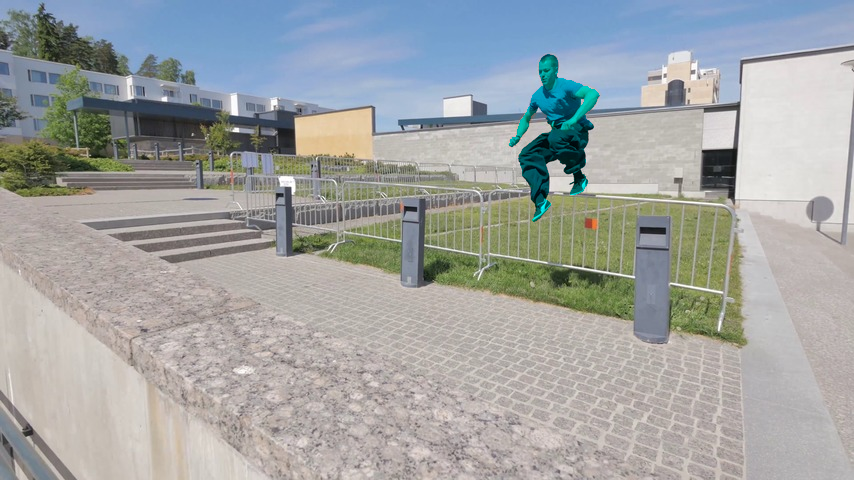}
            \includegraphics[width=0.19\textwidth]{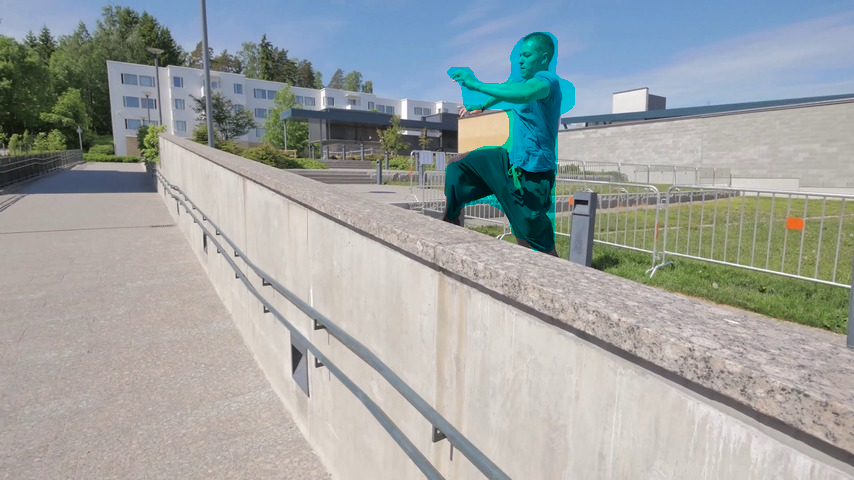}
            \includegraphics[width=0.19\textwidth]{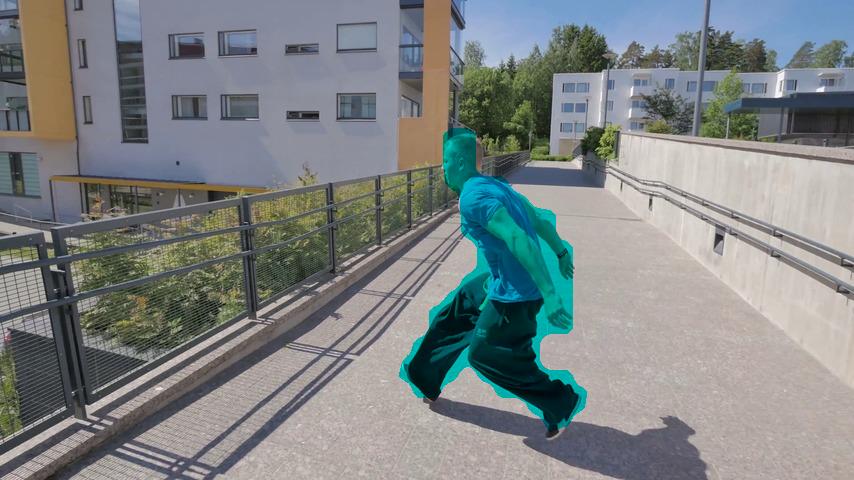}
            \includegraphics[width=0.19\textwidth]{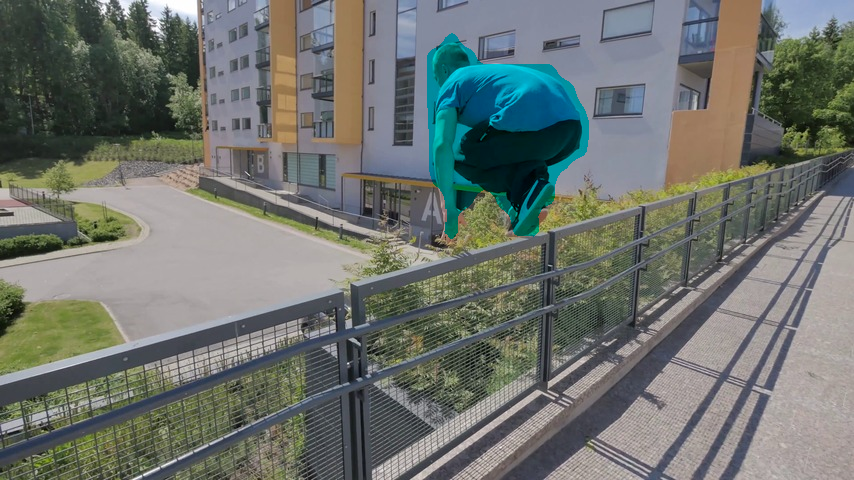}
            \includegraphics[width=0.19\textwidth]{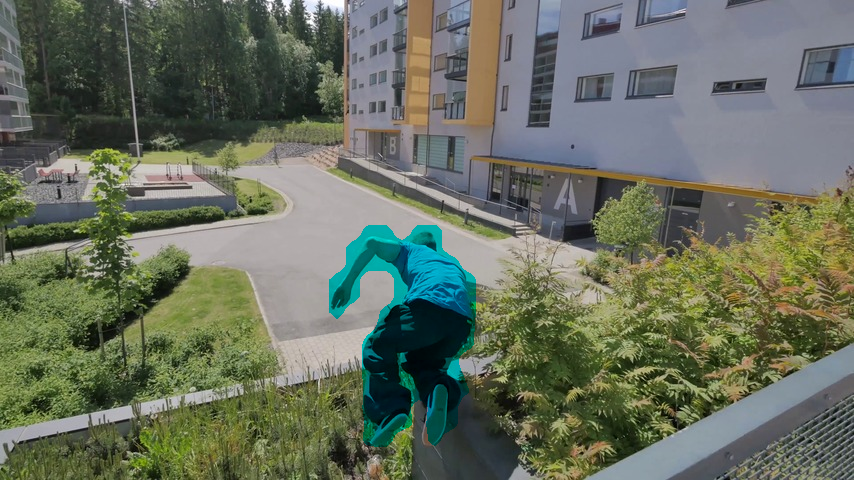}
        \end{center}
        \vspace{-4mm}
        \caption{Video name: \texttt{parkour}}
        \vspace{2mm}
    \end{subfigure}
    \begin{subfigure}{\textwidth}
        \begin{center}
            \includegraphics[width=0.19\textwidth]{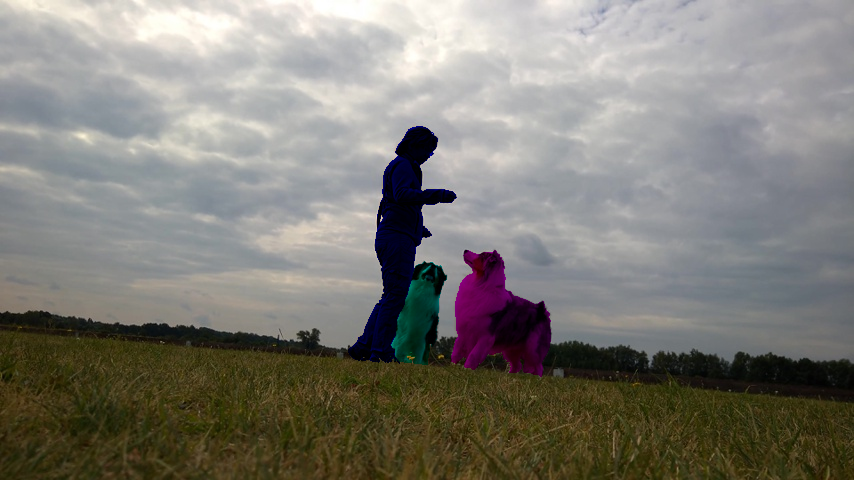}
            \includegraphics[width=0.19\textwidth]{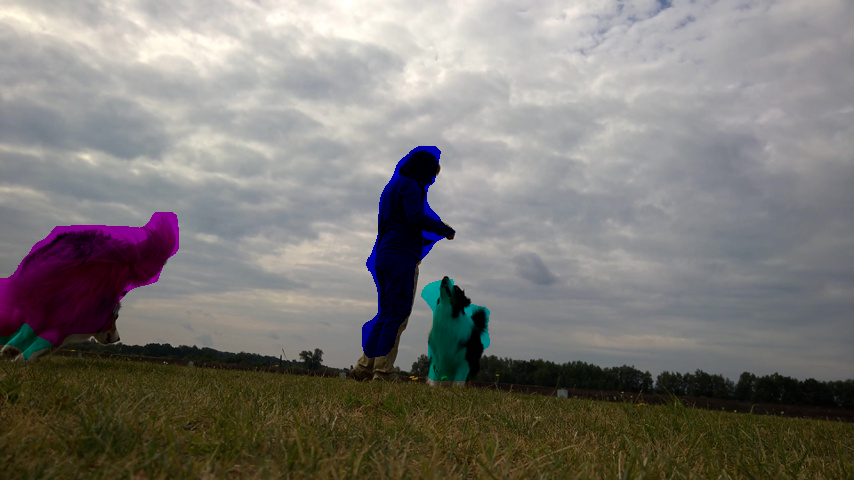}
            \includegraphics[width=0.19\textwidth]{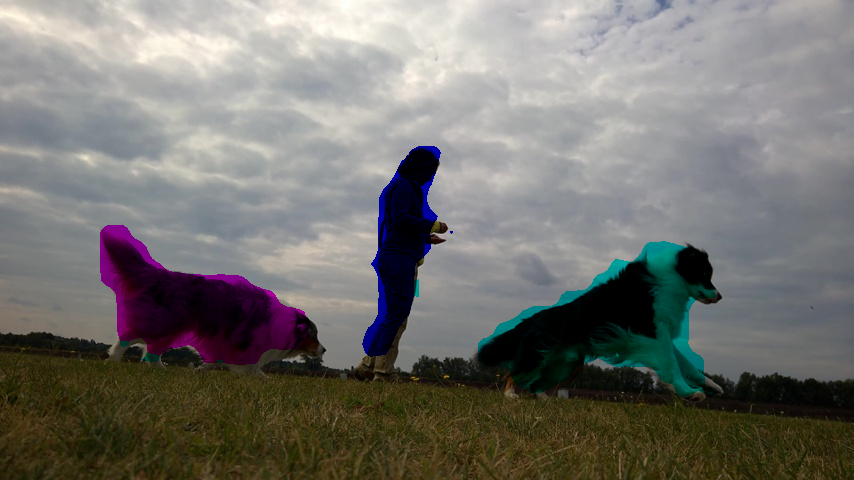}
            \includegraphics[width=0.19\textwidth]{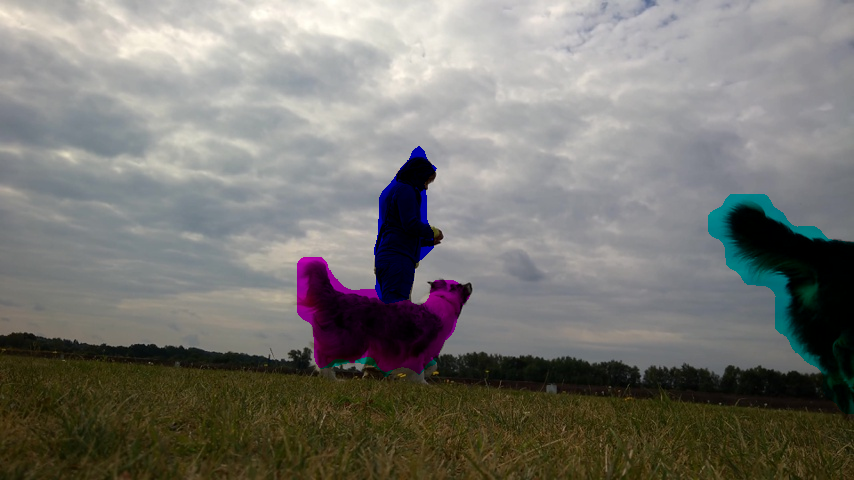}
            \includegraphics[width=0.19\textwidth]{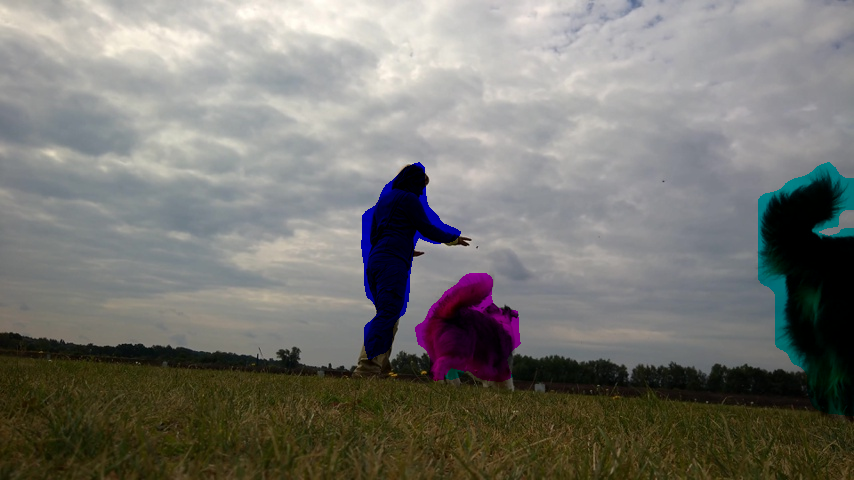}
        \end{center}
        \vspace{-4mm}
        \caption{Video name: \texttt{dogs-jump}}
        \vspace{2mm}
    \end{subfigure}
    \begin{subfigure}{\textwidth}
        \begin{center}
            \includegraphics[width=0.19\textwidth]{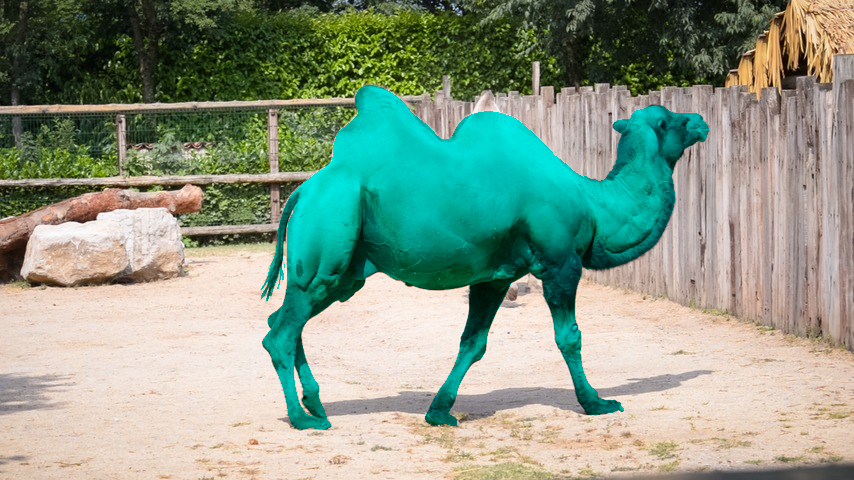}
            \includegraphics[width=0.19\textwidth]{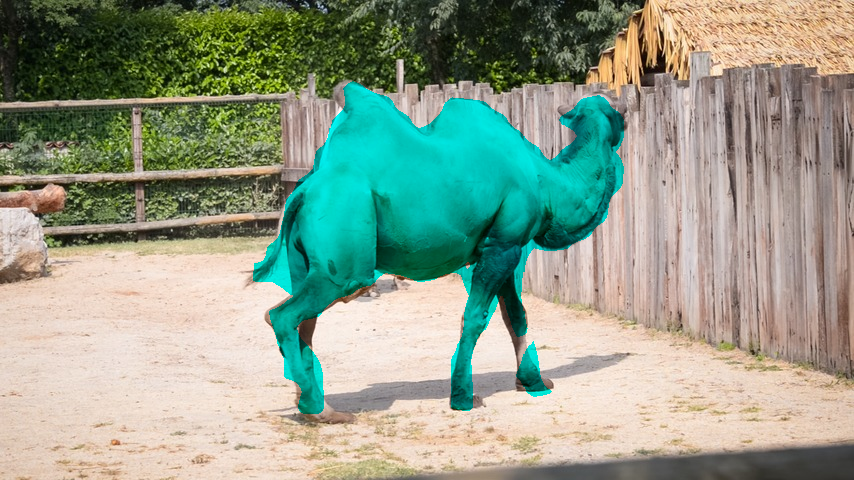}
            \includegraphics[width=0.19\textwidth]{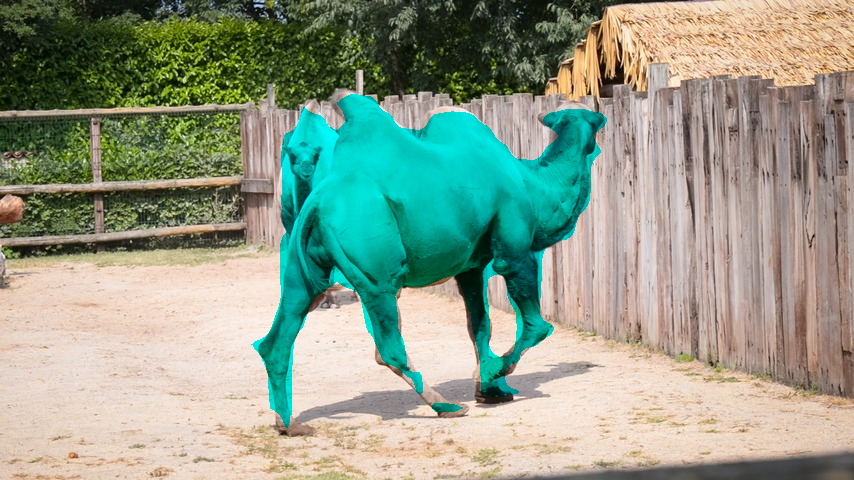}
            \includegraphics[width=0.19\textwidth]{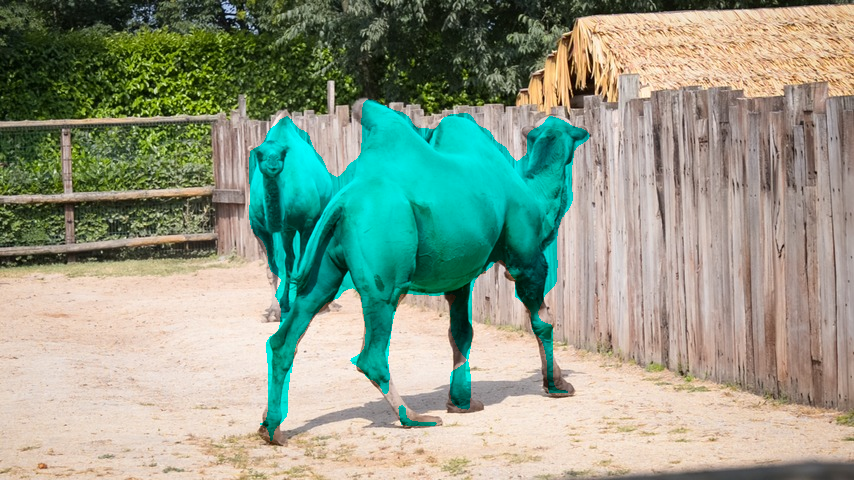}
            \includegraphics[width=0.19\textwidth]{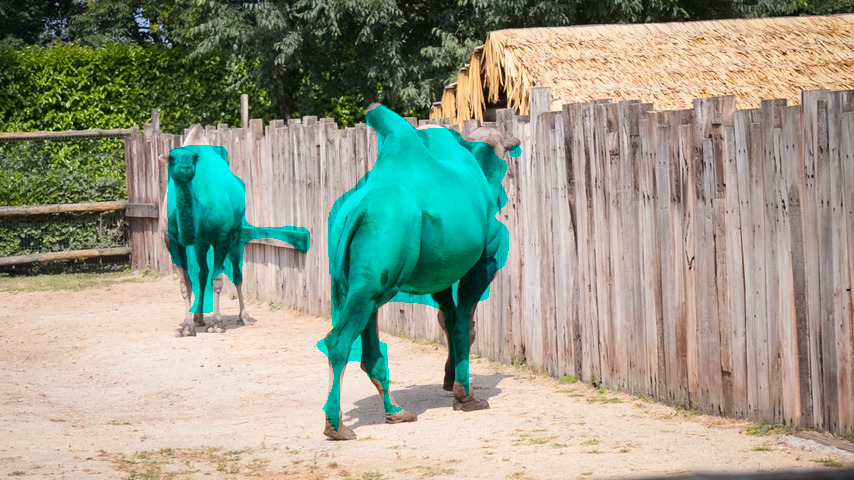}
        \end{center}
        \vspace{-4mm}
        \caption{Video name: \texttt{camel}}
    \end{subfigure}
    \caption{\textbf{Results on Video Object Segmentation (DAVIS-17~\cite{davis}).} The first frame in each video shows the ground-truth annotations masks which are propagated in the remaining frames. The remaining output frames are sampled uniformly over the length of the video.
    }
    \label{fig:vos_vis}
\end{figure*}

Qualitative results on video object segmentation task on DAVIS2017 are shown in Fig.~\ref{fig:vos_vis}.

\subsection{Pose Propagation on JHMDB}
\begin{figure*}[h]
    
    \begin{subfigure}{\textwidth}
        \begin{center}
            \includegraphics[width=0.19\textwidth]{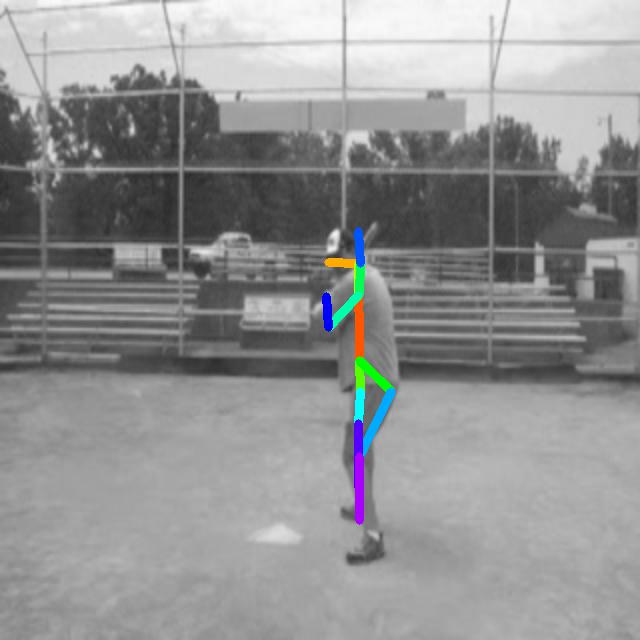}
            \includegraphics[width=0.19\textwidth]{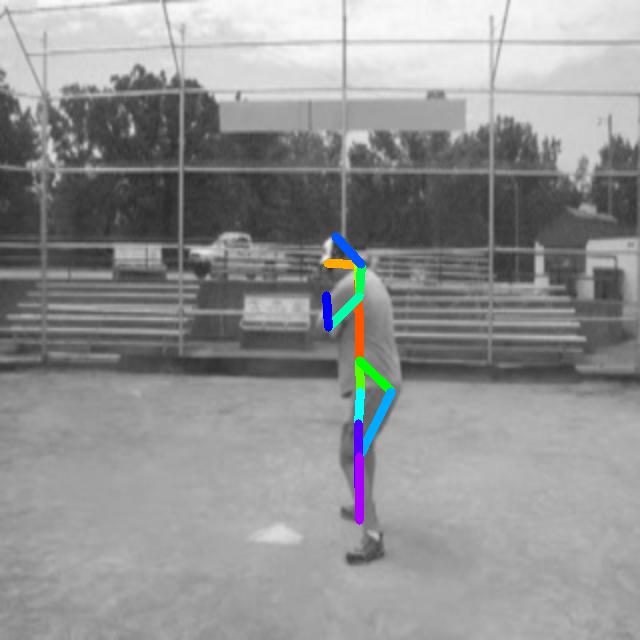}
            \includegraphics[width=0.19\textwidth]{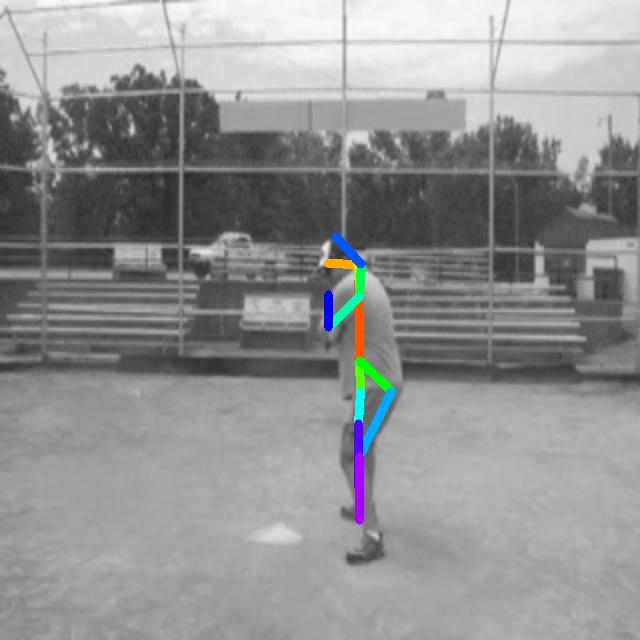}
            \includegraphics[width=0.19\textwidth]{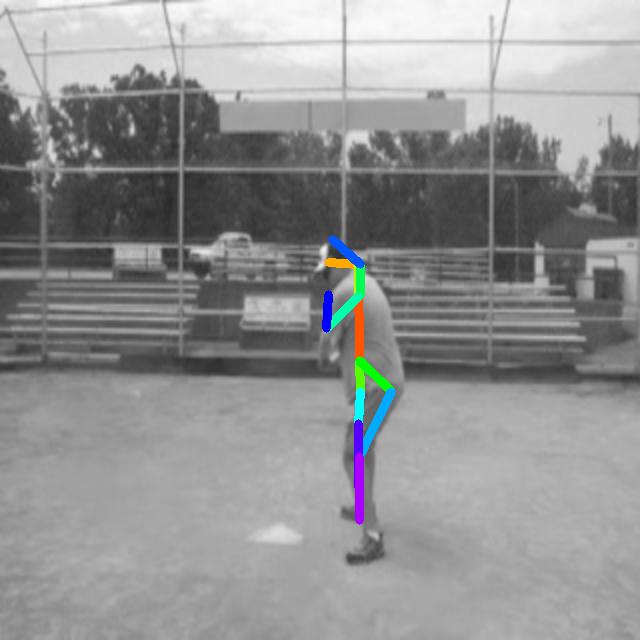}
            \includegraphics[width=0.19\textwidth]{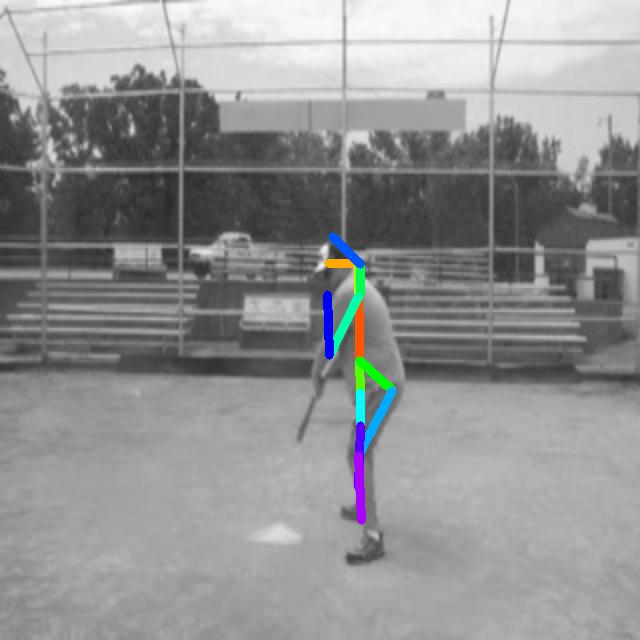}

        \end{center}
    \end{subfigure}
    \begin{subfigure}{\textwidth}
        \begin{center}
            \includegraphics[width=0.19\textwidth]{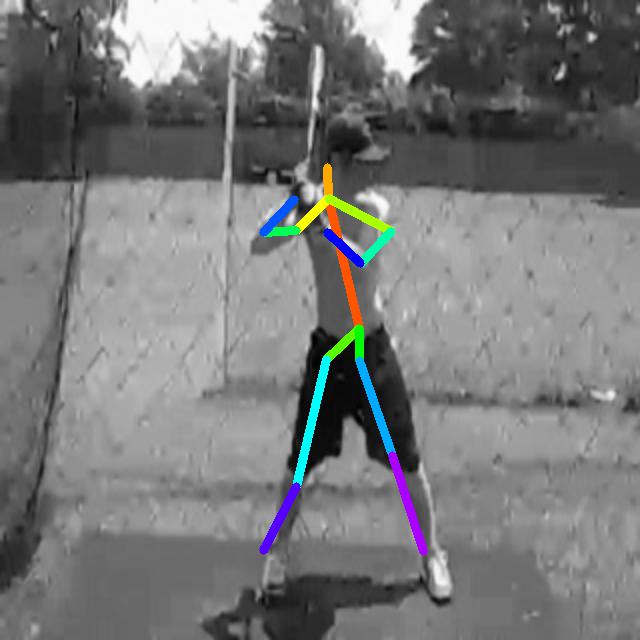}
            \includegraphics[width=0.19\textwidth]{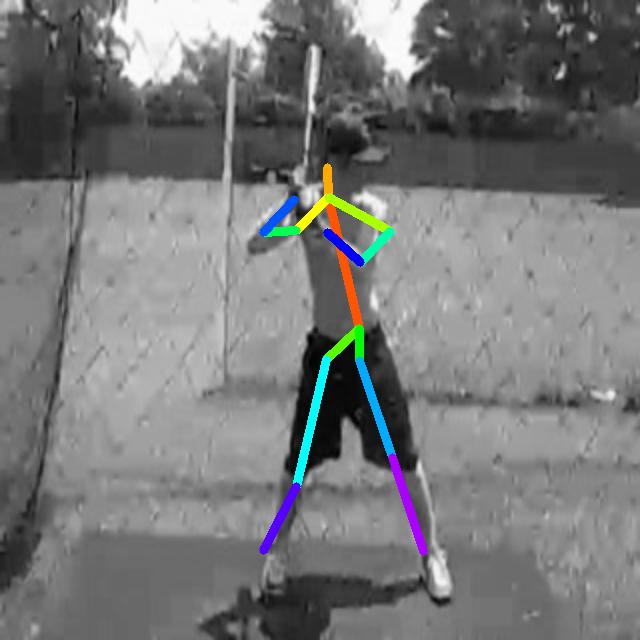}
            \includegraphics[width=0.19\textwidth]{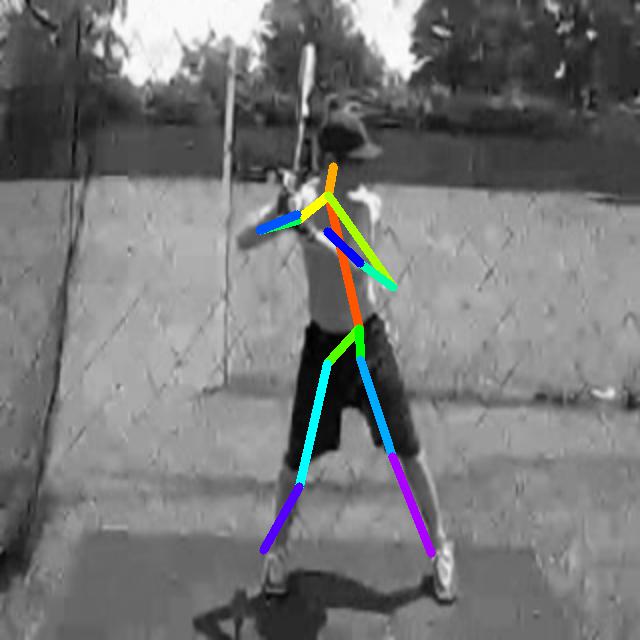}
            \includegraphics[width=0.19\textwidth]{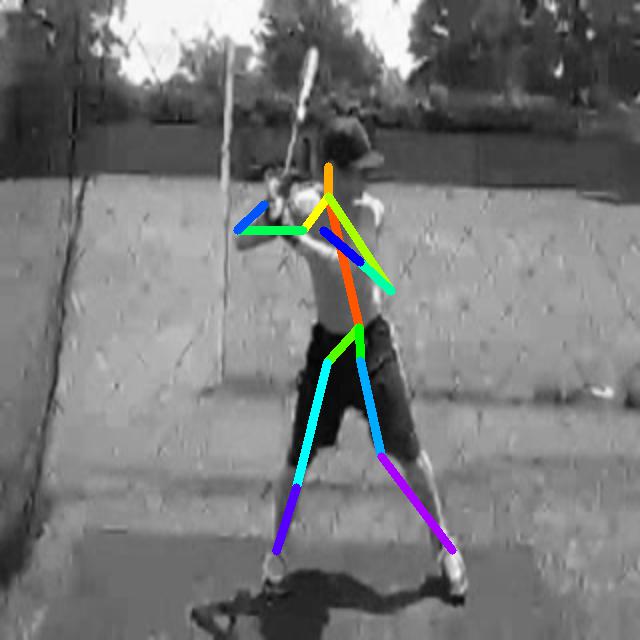}
            \includegraphics[width=0.19\textwidth]{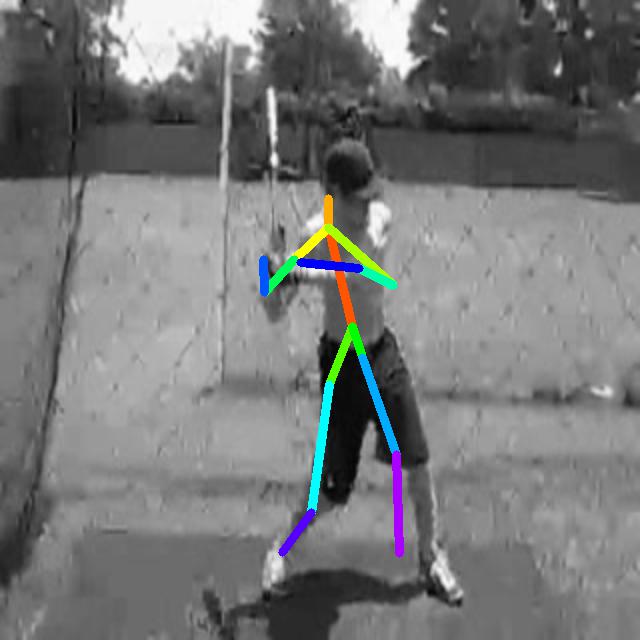}

        \end{center}
    \end{subfigure}
   \begin{subfigure}{\textwidth}
        \begin{center}
            \includegraphics[width=0.19\textwidth]{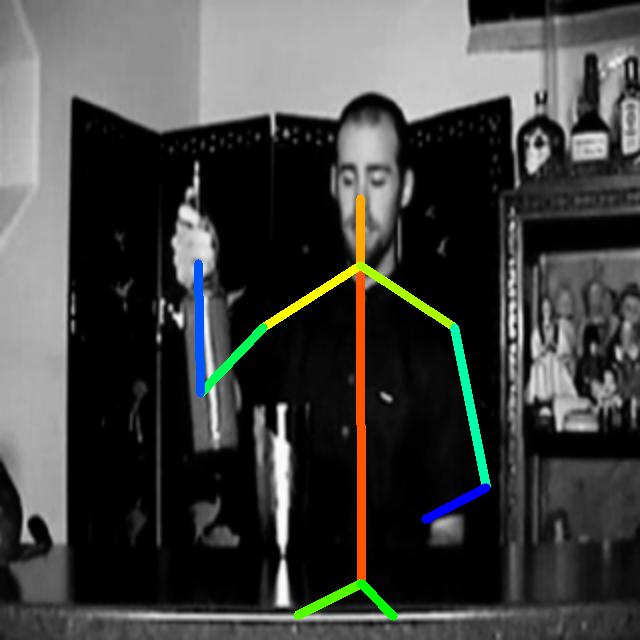}
            \includegraphics[width=0.19\textwidth]{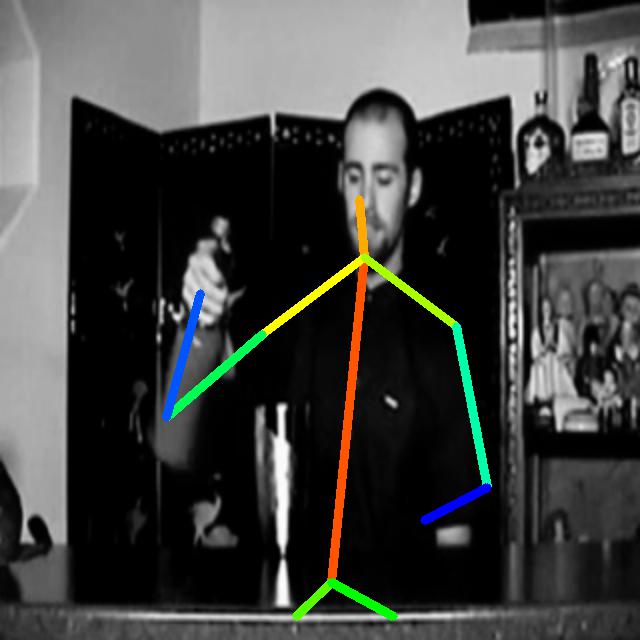}
            \includegraphics[width=0.19\textwidth]{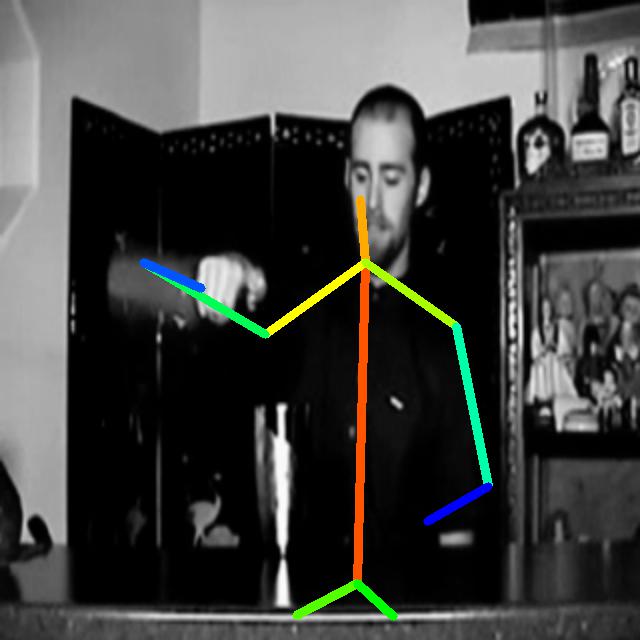}
            \includegraphics[width=0.19\textwidth]{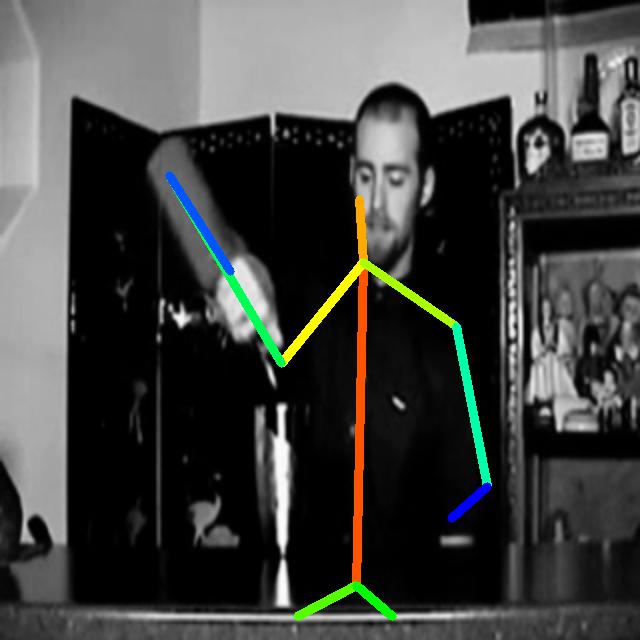}
            \includegraphics[width=0.19\textwidth]{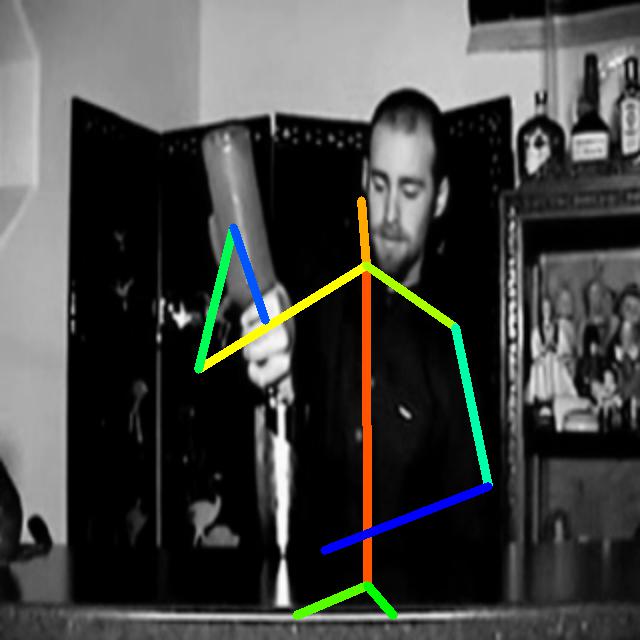}

        \end{center}
    \end{subfigure}
    \begin{subfigure}{\textwidth}
        \begin{center}
            \includegraphics[width=0.19\textwidth]{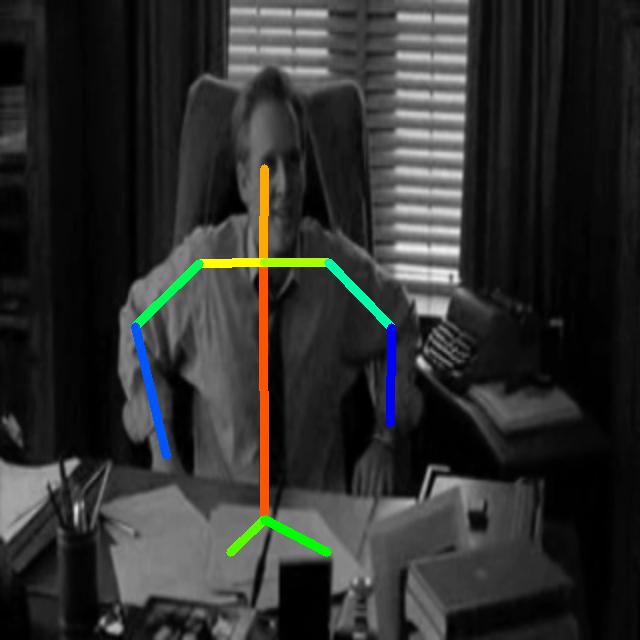}
            \includegraphics[width=0.19\textwidth]{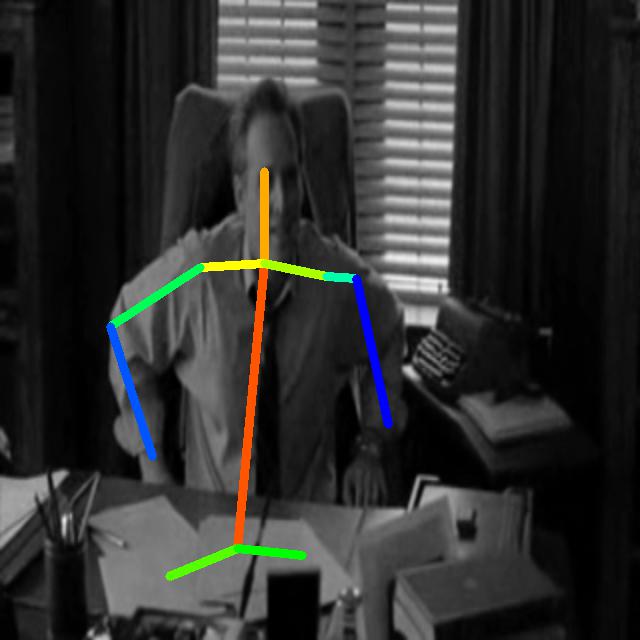}
            \includegraphics[width=0.19\textwidth]{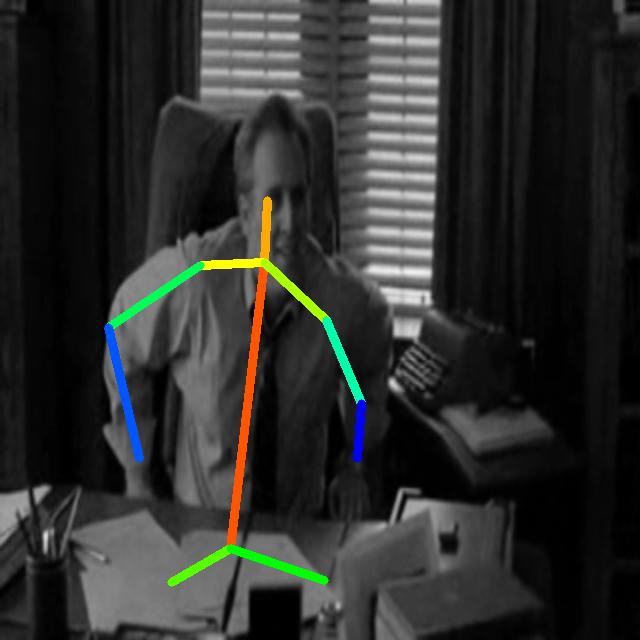}
            \includegraphics[width=0.19\textwidth]{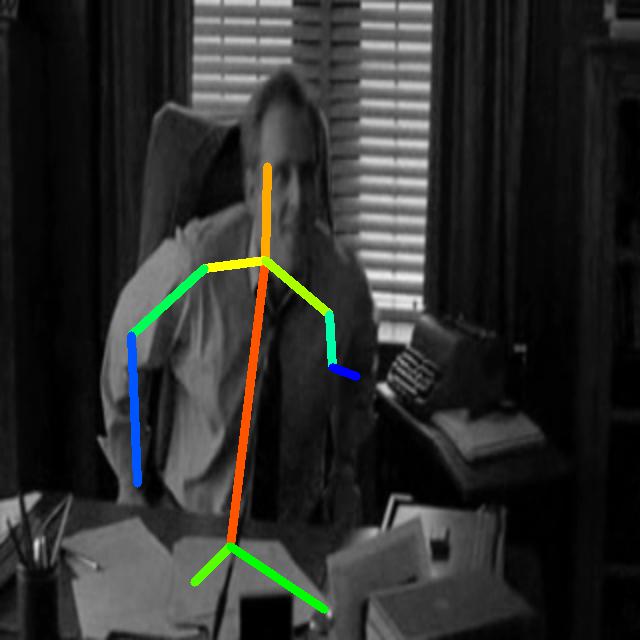}
            \includegraphics[width=0.19\textwidth]{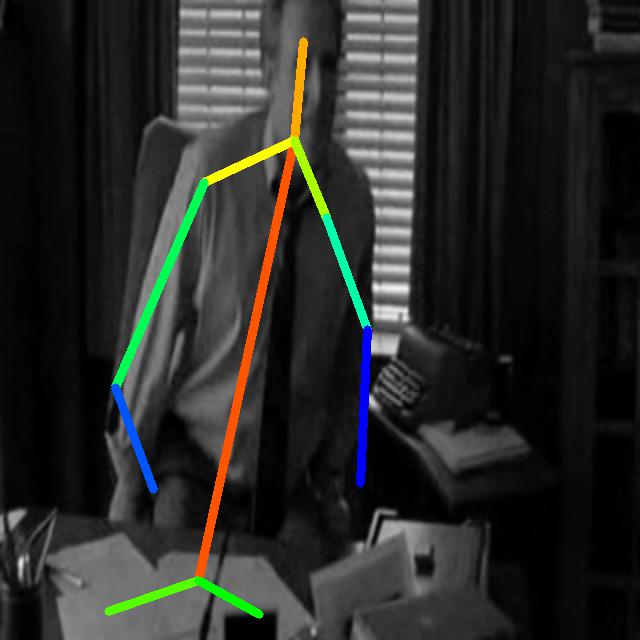}

        \end{center}
    \end{subfigure}
    \begin{subfigure}{\textwidth}
        \begin{center}
            \includegraphics[width=0.19\textwidth]{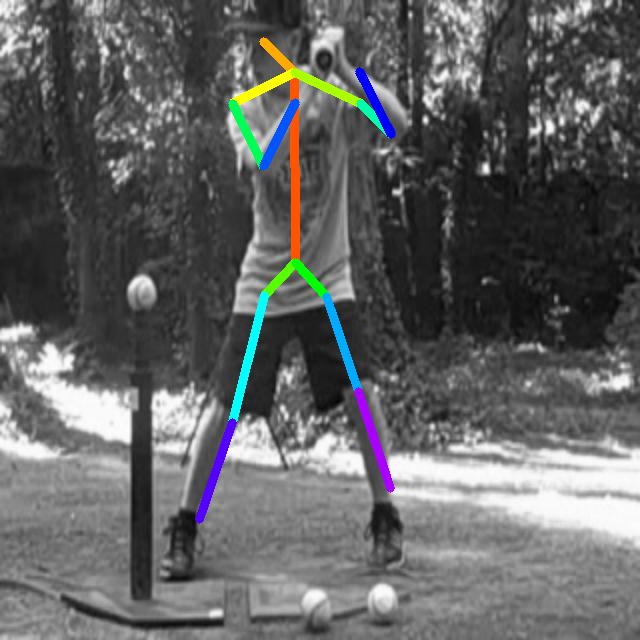}
            \includegraphics[width=0.19\textwidth]{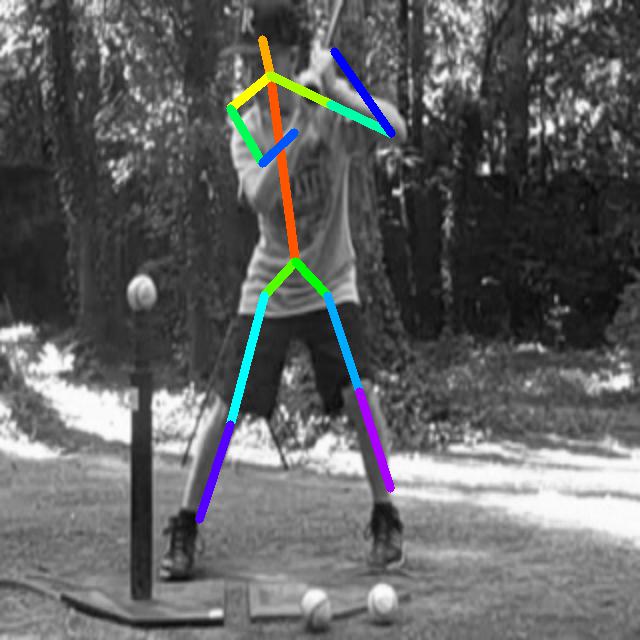}
            \includegraphics[width=0.19\textwidth]{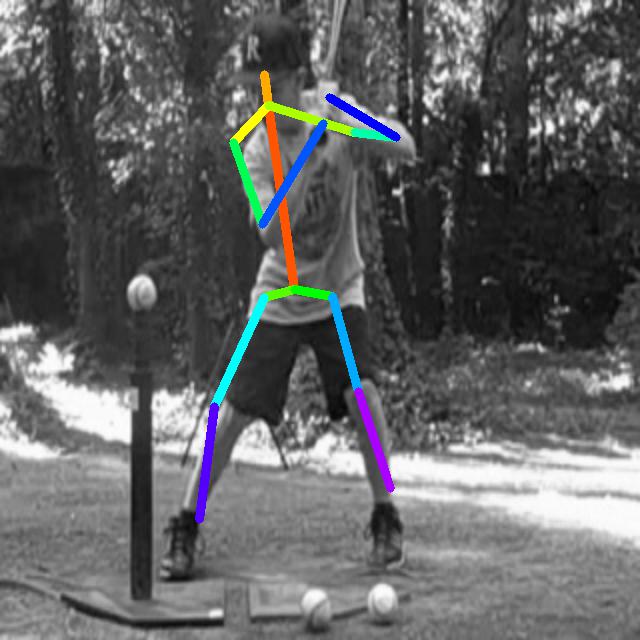}
            \includegraphics[width=0.19\textwidth]{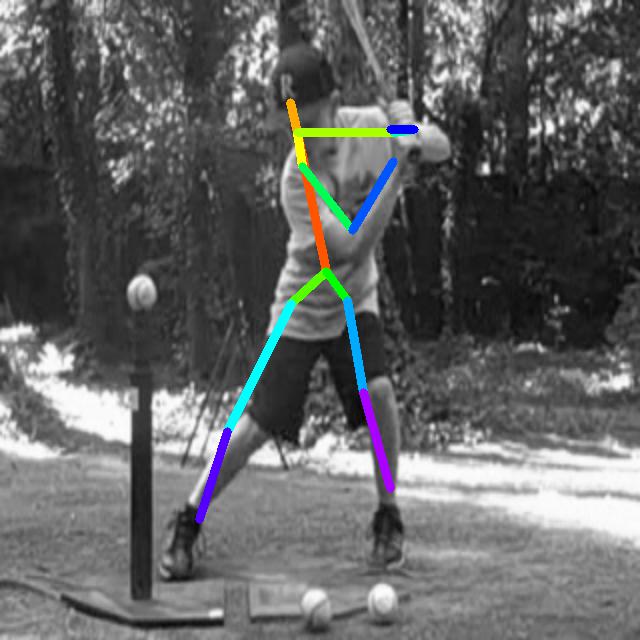}
            \includegraphics[width=0.19\textwidth]{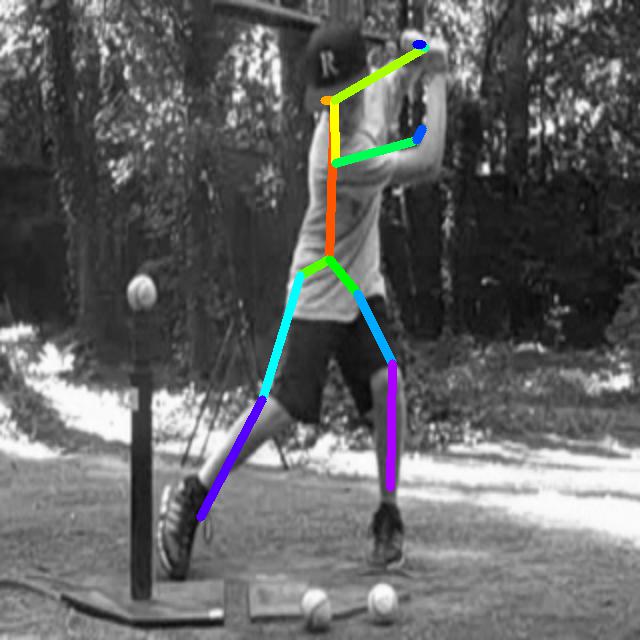}

        \end{center}
    \end{subfigure}
    \begin{subfigure}{\textwidth}
        \begin{center}
            \includegraphics[width=0.19\textwidth]{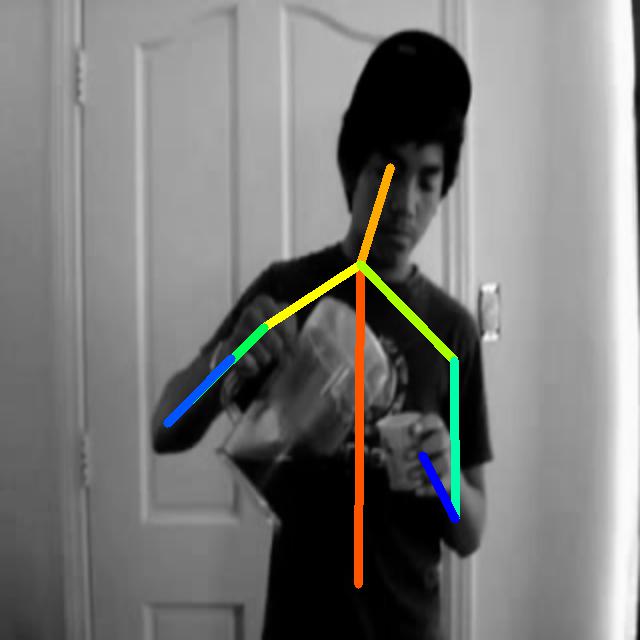}
            \includegraphics[width=0.19\textwidth]{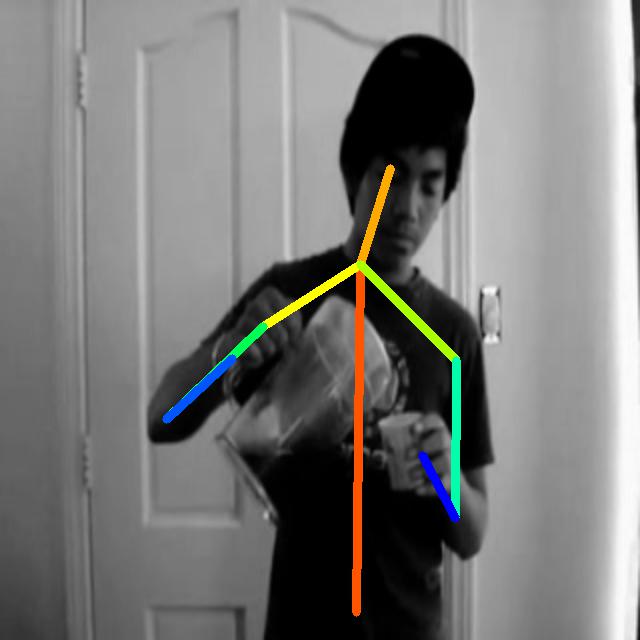}
            \includegraphics[width=0.19\textwidth]{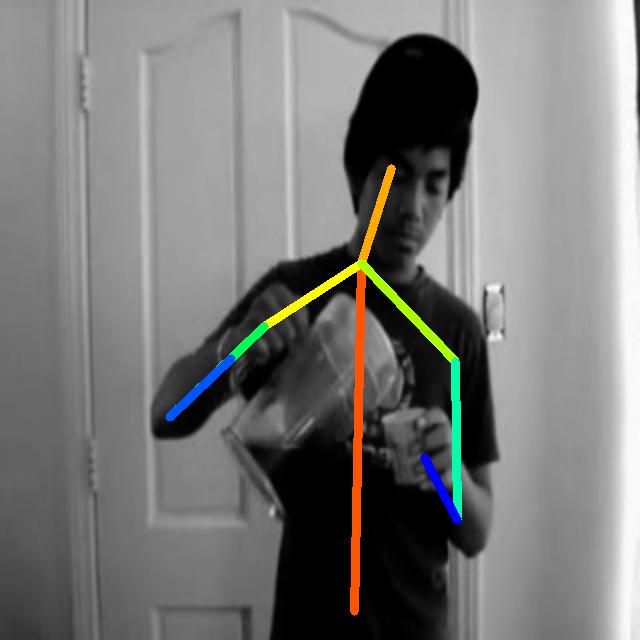}
            \includegraphics[width=0.19\textwidth]{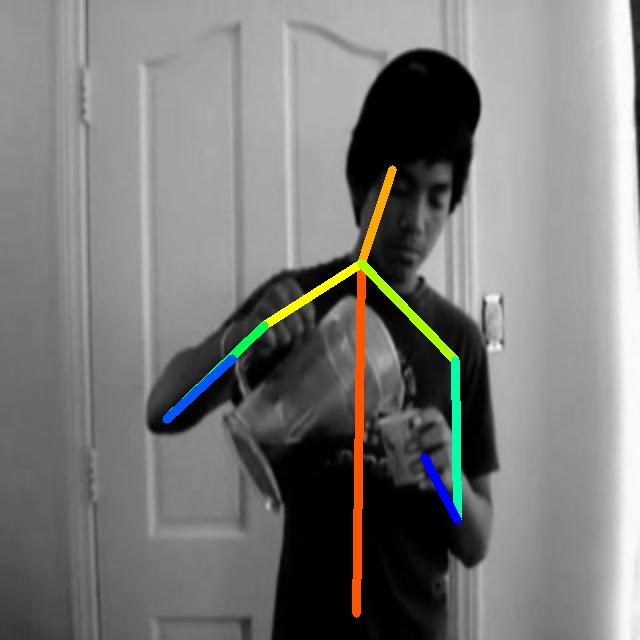}
            \includegraphics[width=0.19\textwidth]{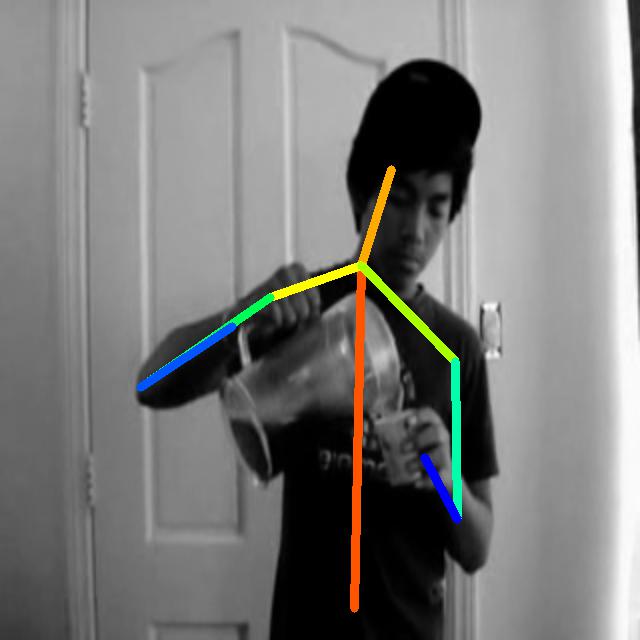}

        \end{center}
    \end{subfigure}
    \caption{\textbf{Qualitative Results on Pose Propagation (JHDMB~\cite{jhmdb21}).} Output frames are sampled uniformly over the length of the video.
    }
    \vspace{8mm}
    \label{fig:pose_vis}
\end{figure*}

Qualitative results on Pose Propagation task on JHMDB are shown in Fig.~\ref{fig:pose_vis}.

\clearpage

\section{Comparison with Prior Temporal Pretext Tasks}
\label{sec:conceptual_compare}
\noindent \textbf{Out-of-order Frame Localization} We illustrate the difference between our OFL and traditional sequence-level frame-order verification task in Fig.~\ref{fig:ofl_trad}.
\begin{center}
\includegraphics[width=0.9\textwidth]{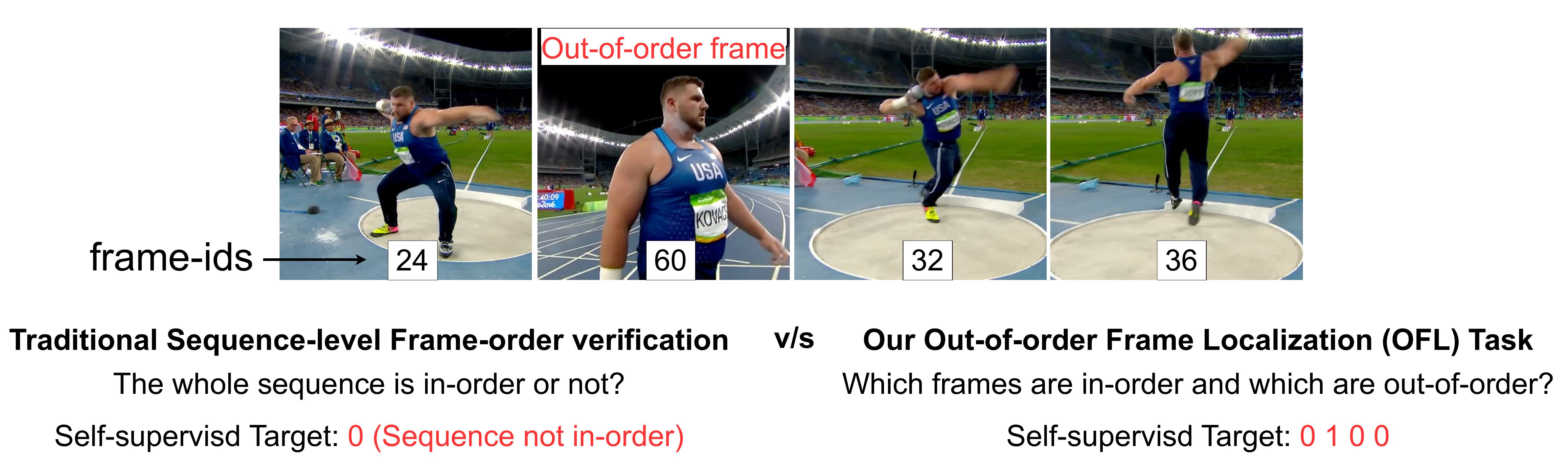}

\begin{minipage}{\textwidth} %
\captionof{figure}{Illustration for Comparison of Traditional Sequence-level Frame Verification vs. our Out-of-order Frame Localization.}
\label{fig:ofl_trad}
\end{minipage}
\end{center}

\end{document}